\documentclass{article}

\PassOptionsToPackage{numbers, compress}{natbib}

\usepackage[preprint]{neurips_2024}

\usepackage[utf8]{inputenc} 
\usepackage[T1]{fontenc}    
\usepackage{url}            
\usepackage{booktabs}       
\usepackage{amsfonts}       
\usepackage{nicefrac}       
\usepackage{microtype}      

\usepackage{times}
\usepackage{epsfig}
\usepackage{amsmath}
\usepackage{makecell} 

\usepackage{xspace}
\usepackage{graphicx}
\usepackage{subcaption}
\usepackage{multirow} 
\usepackage{colortbl}
\usepackage{pifont}
\usepackage{wrapfig}
\usepackage{tcolorbox}
\usepackage{float}
\usepackage[pagebackref, breaklinks=true, colorlinks, citecolor=citecolor, linkcolor=linkcolor, bookmarks=false]{hyperref}
\definecolor{citecolor}{HTML}{0071BC}
\definecolor{linkcolor}{HTML}{ED1C24}

\newcommand{\ie}{{\emph{i.e.}}}
\newcommand{\eg}{{\emph{e.g.}}}
\newcommand{\etc}{\emph{etc}}

\newcommand{\vs}{\emph{vs.}}

\newcommand{\yes}{\ding{51}}
\newcommand{\no}{\ding{55}}
\newcommand{\graybox}[1]{\colorbox{gray!15}{#1}}

\def\modelname{VisionLLM v2\xspace}

\makeatletter
\def\blfootnote{\xdef\@thefnmark{}\@footnotetext}
\makeatother

\title{\modelname: An End-to-End Generalist Multimodal Large Language Model for Hundreds of Vision-Language Tasks}

\author{%
  \hspace{-0.25cm}\textbf{
  Jiannan Wu$^{*2,1}$,
  Muyan Zhong$^{*3}$,
  Sen Xing$^{*3}$,
  Zeqiang Lai$^{*4}$,
  Zhaoyang Liu$^{*5,1}$, 
  Zhe Chen$^{*6,1}$,
  } \\
  \hspace{-0.25cm}\textbf{
  Wenhai Wang$^{*7,1}$,
  Xizhou Zhu$^{3,8,1}$,
  Lewei Lu$^{8,1}$, 
  Tong Lu$^{6}$,
  Ping Luo$^{2}$,
  Yu Qiao$^{1}$,
  Jifeng Dai$^{\dagger 3,1}$
  } \\
  \hspace{-0.25cm}
  $^1$OpenGVLab, Shanghai AI Laboratory \quad 
  $^2$The University of Hong Kong \quad 
  $^3$Tsinghua University \\
  \hspace{-0.25cm}
  $^4$Beijing Institute of Technology \quad 
  $^5$The Hong Kong University of Science and Technology \\
  \hspace{-0.25cm}
  $^6$Nanjing University \quad 
  $^7$The Chinese University of Hong Kong \quad 
  $^8$SenseTime Research \\\\
  \url{https://github.com/OpenGVLab/VisionLLM}
}

\begin{document}

\maketitle

\thispagestyle{empty}

\blfootnote{\noindent $^{*}$Equal contribution.  
$^{\dagger}$ Corresponding to Jifeng Dai <daijifeng@tsinghua.edu.cn>.}

\begin{abstract}

We present VisionLLM v2, an end-to-end generalist multimodal large model (MLLM) that unifies visual perception, understanding, and generation within a single framework. Unlike traditional MLLMs limited to text output, VisionLLM v2 significantly broadens its application scope. It excels not only in conventional visual question answering (VQA) but also in open-ended, cross-domain vision tasks such as object localization, pose estimation, and image generation and editing.
To this end, we propose a new information transmission mechanism termed ``super link'', as a medium to connect MLLM with task-specific decoders. 
It not only allows flexible transmission of task information and gradient feedback between the MLLM and multiple downstream decoders but also effectively resolves training conflicts in multi-tasking scenarios. In addition, to support the diverse range of tasks, we carefully collected and combed training data from hundreds of public vision and vision-language tasks. In this way, our model can be joint-trained end-to-end on hundreds of vision language tasks and generalize to these tasks using a set of shared parameters through different user prompts, achieving performance comparable to task-specific models. 
We believe VisionLLM v2 will offer a new perspective on the generalization of MLLMs.

\end{abstract}


\vspace{-5px}
\section{Introduction}
\vspace{-5px}

Multimodal large language models (MLLMs)~\cite{alayrac2022flamingo, li2023blip2, zhu2023minigpt4, liu2023llava, liu2023llava1.5, peng2023kosmos2, bai2023qwenvl, team2023gemini, chen2023internvl, chen2024internvl_1_5} have recently made significant progress, demonstrating outstanding performance across various vision-language tasks, even in scenarios requiring complex understanding and reasoning. 
However, \emph{a notable limitation is that current MLLM outputs are in text form, which significantly constrains their capacity to represent structured or visual information.} 
Some researchers~\cite{peng2023kosmos2, wang2023visionllm, wang2023allseeing, wang2024allseeingv2} have expanded the text-based output formats of MLLMs to better align with downstream tasks. While these efforts have shown promise, they have not fully addressed practical needs such as dense object detection, pose estimation, and image generation.

\begin{figure}[t]
    \centering
    \includegraphics[width=1.0\textwidth]{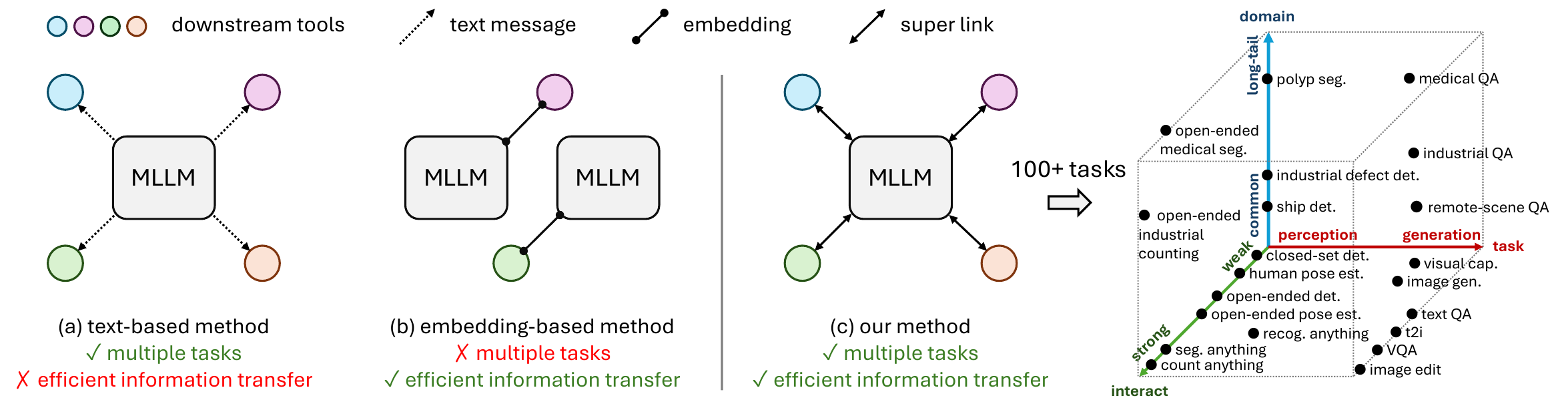}
    \vspace{-15pt}
    \caption{\textbf{Illustration of three information transmission methods.} (a) Text-based method shows MLLM connected to various downstream tools via text messages, capable of handling multiple tasks but suffering from inefficient information transfer. (b) The embedding-based method displays a connection using learnable embeddings, which facilitates efficient information transfer but lacks support for multitasking. (c) Our method employs a ``super link'' technique, where a unified MLLM interfaces with multiple task decoders through super links, supporting over 100 diverse tasks.
    } 
    \label{fig:moti}
    \vspace{-20pt}
\end{figure}

To overcome this limitation, a line of research~\cite{2023interngpt, wu2023visual-chatgpt, suris2023vipergpt, liu2023llava-plus, liu2023controlllm, fei2024vitron} enhances the capabilities of MLLMs by transmitting task information to tools via text messages, as illustrated in Figure \ref{fig:moti}(a). 
Despite these advances, these text-based methods are restricted by the information that text can convey. They are not end-to-end, and the feedback gradient from the tools cannot be relayed back to the MLLM. 
This limitation has spurred another research direction~\cite{lai2023lisa, rasheed2023glamm, xu2023pixelllm, dong2023dreamllm, koh2024gill, sun2023emu} that employs learnable embeddings as intermediaries to connect MLLM with one specific task decoder (see Figure \ref{fig:moti}(b)).
However, the naive embedding connection is difficult to scale to multi-task scenarios.
A routing mechanism is needed to ensure the correct selection of tools, and the issue of task conflicts~\cite{zhu2022uni} arising from joint multi-task training is also a problem that needs to be considered.
Therefore, \emph{developing an end-to-end MLLM generalist for various vision and vision-language tasks beyond text output remains a significant challenge.}

Given these challenges, developing an end-to-end generalist MLLM requires a more effective information transmission method than conventional text messages and naive embeddings. 
This method should ensure that task information and feedback gradients are accurately and flexibly communicated between the central MLLM and multi-task decoders while preventing task conflicts across various visual domains and input/output formats.
In addition, multi-task datasets for generalist MLLMs need to be well-prepared. Despite the abundance of annotations in the community, the diverse and inconsistent formats of these annotations across different tasks make it challenging to develop a unified dataset that effectively supports multi-task learning.

In this work, we introduce \modelname, an end-to-end generalist MLLM designed for a wide array of vision and vision-language tasks. This model not only performs typical visual question answering but also extends to image generation, image editing, and open-ended object detection/instance segmentation/pose estimation across diverse image domains.
To facilitate information transmission between the MLLM and multiple downstream task decoders,
we introduce the \textbf{super link} technique, 
which consists of two components: (1) \emph{Routing Token}: 
special tokens (\eg, \texttt{[DET]}, \texttt{[POSE]}, and \texttt{[GEN]}) added to the MLLM's vocabulary. Whenever the MLLM predicts a specific routing token, it triggers the selection of the appropriate decoder.
(2) \emph{Super-Link Queries} randomly initialized learnable weights bound to the routing tokens. These queries are appended after the routing tokens and processed by the MLLM to extract task-specific information, which is then sent to the target decoder. This method enables flexible task information transmission, allows decoder gradients to backpropagate to the MLLM, and avoids task conflicts by ensuring the queries are bound to routing tokens and not shared across tasks.

Furthermore, we carefully collected and curated training data from hundreds of public vision and vision-language tasks to support various tasks. The data includes high-quality examples of visual question answering, visual perception, recognition, and understanding tasks from various sources such as natural scenes, remote sensing images, medical images, and industrial images. To ensure effective training with these extensive datasets, we also implemented a multi-stage joint training strategy, integrating new abilities and reaching a performance comparable to the expert models while maintaining the MLLM's foundational VQA capabilities.

These designs endow \modelname with three distinct characteristics: (1) \emph{Generality}. With one suit of parameters, our model can be generalized to different tasks using different text and visual prompts. To our knowledge, it is the first end-to-end model to support hundreds of vision-language tasks while achieving performance comparable to expert models. (2) \emph{Openness}. By employing open-ended decoders, our model allows users to freely define tasks through multimodal prompts, breaking away from the constraints of closed-set models limited to predefined tasks or categories. Furthermore, users can flexibly combine various tasks into more complex ones through multi-round dialogue. (3) \emph{Multimodal In-Context Ability}. With multimodal inputs and outputs, our model demonstrates extensive versatility 
and exhibits superiority over the previous in-context models with single-modal outputs~\cite{wang2023seggpt, alayrac2022flamingo}. 
These features distinguish our model from previous 
approaches,
and establish a leading foundational MLLM for various vision and vision-language applications.

In summary, our main contributions are listed as follows:

(1) We propose \modelname, the first end-to-end generalist MLLM model to accomplish hundreds of vision and vision-language tasks\footnote{We consider tasks such that those with differing input and output formats, or those involving data from different domains as distinct tasks.}, covering visual perception, understanding, and generation.
It not only addresses the limitation of LLMs being confined to text outputs but also supports using textual, visual, and in-context instructions to flexibly combine tasks for real-world applications.

(2) We introduce the super-link technique, which integrates the MLLM with task-specific decoders. This integration facilitates end-to-end optimization across both linguistic and visual tasks. Additionally, we meticulously collect and re-organize data from a broad range of domains and develop an in-context learning dataset. These efforts lay a solid foundation for our progressive joint training process and enable the model to benefit from individual tasks.

(3) We comprehensively evaluate the proposed model on a wide range of vision and vision-language tasks, from visual perception to visual understanding, from weak interaction (\eg, closed-set) to strong interaction (\eg, visual prompt + language prompt), from common-seen domains to long-tailed domains (\eg, medical, remote-sensing, industry), as shown in the rightmost subfigure of Figure \ref{fig:moti}. In addition, with a generalist model, our method achieves comparable performance with the task-specialized models in various standard benchmarks.


\vspace{-5px}
\section{Related Work}
\vspace{-5px}

\subsection{Multimodal Large Language Model}
\vspace{-5px}

\noindent
\textbf{Conventional MLLMs}. With the advancement of large language models (LLMs)~\cite{radford2018improving,radford2019language,brown2020gpt3,zhang2022opt,touvron2023llama,vicuna2023,touvron2023llama2,qwen,2023internlm,falcon40b,lian2023mistralorca1,meta2024llama3,Gunasekar2023phi,deepseek-llm,cai2024internlm2}, multimodal large language models (MLLMs) have also gained significant momentum recently.
Notable commercial models include GPT-4V~\cite{gpt4v}, Gemini series~\cite{team2023gemini,reid2024gemini1_5}, Claude-3~\cite{claude3series2024}, and Qwen-VL-Max~\cite{bai2023qwenvl}, known for their outstanding performance. Early open-source MLLMs like InstructBLIP~\cite{instructblip}, LLaVA~\cite{liu2023llava} and MiniGPT-4~\cite{zhu2023minigpt4} fine-tune on instruction-following datasets. InternVL ~\cite{chen2023internvl, chen2024internvl_1_5} series models align a large-scale vision encoder with LLMs and perform comparably to commercial models. 
Efficient MLLMs~\cite{Li2024MiniGeminiMT,Zhu2024llavaPhi,Chu2024MobileVLM} have also studied.
However, these models only can output text, restricting their applications.

\textbf{Extension of MLLMs' Text Output}. To extend MLLMs to downstream tasks, models like Kosmo-2~\cite{peng2023kosmos2}, Shikra~\cite{chen2023shikra}, VisionLLM~\cite{wang2023visionllm}, Ferret~\cite{you2023ferret, zhang2024ferretv2}, and All-Seeing V2~\cite{wang2024allseeingv2} achieve this using specially-designed tokens or encoding coordinates as text tokens. Despite these advancements, using LLMs solely as visual decoders falls short of resolving the fine-grained visual context needed for precise detection and segmentation. 
The other line of works focus on broadening the modality scope. AnyGPT~\cite{zhan2024anygpt} builds a multimodal text-centric dataset for any-to-any multimodal generation (text, image, speech, music) with sequence modeling. Chameleon~\cite{team2024chameleon} uses fully token-based representations for both texts and images, capable of understanding and generating interleaved image-text sequences. CM3leon~\cite{aghajanyan2022cm3, yu2023cm3leon} are autoregressive models for text-to-image and image-to-text tasks. All these works could unify image understanding and generation in one network. Our model can support more vision and vision-language tasks. 

\textbf{MLLMs w/ Downstream Tools}. Recent works~\cite{2023interngpt, wu2023visual-chatgpt, suris2023vipergpt, liu2023llava-plus, liu2023controlllm, fei2024vitron, betker2023improving, xia2023llmga, huang2024dialoggen, hao2024vitron} have integrated external tools for vision-centric tasks, transmitting task information to these tools via text messages. However, such text-based communication between LLMs and tools hinders end-to-end optimization.
Another category of approaches~\cite{lai2023lisa,rasheed2023glamm,zhang2024psalm,koh2024gill,sun2023emu,sun2023emu2,ge2023planting,ge2023making,pan2023kosmos,dong2023dreamllm,fu2023guiding,huang2023smartedit} feeds the output embeddings of LLMs into a special decoder and trains them end-to-end to enhance information communication. However, they only support semantic segmentation or image generation tasks.
In this work, we target to develop an end-to-end MLLM generalist for diverse vision and vision-language tasks beyond text output.

\vspace{-5px}
\subsection{Vision Generalist Model}
\vspace{-5px}

\noindent
\textbf{Unified Vision Model.}
The unified model approach integrates multiple visual tasks into a single framework, enhancing efficiency and reducing the complexity of deploying separate models for each task. 
Works such as Pix2Seq-D~\cite{chen2023generalist}, SEEM~\cite{zou2024segment}, and Semantic-SAM~\cite{li2023semantic} focus on unifying the segmentation interface, achieving promising results. 
Grounding-DINO~\cite{liu2023groundingdino} and VisionLLM~\cite{wang2023visionllm} explore open-set detection grounded by language, while UniPose~\cite{yang2023unipose} excels in pose estimation. 
Additionally, pioneering works~~\cite{zhu2022uniperceiver, zhu2022uni, li2023uniperceiverv2, lu2022unified-io, zou2023xdecoder, wu2023omni} aim to design a unified model capable of solving multiple tasks,  including detection, segmentation, captioning, \etc. 
Their results demonstrate the feasibility of a single model performing diverse tasks. 

\noindent
\textbf{Visual Prompting.}
Visual prompting has emerged as a novel paradigm by providing visual marks in the input instruction. It requires the model to pay attention to the specific region on the image when answering the question. 
Techniques like red circle~\cite{shtedritski2023does}, SoM~\cite{yang2023set}, AutoVP~\cite{tsao2023autovp}, ILM-VP~\cite{chen2023understanding}, and PIVOT~\cite{nasiriany2024pivot} significantly reduce the need for textual prompt engineering, assisting models in focusing on relevant visual content. 
Similar to in-context learning in LLMs, Painter~\cite{wang2023images}, DINO v2~\cite{li2023visual}, and SegGPT~\cite{wang2023seggpt} leverage visual context to improve learning efficiency and adaptability, enabling models to adapt to new tasks with minimal input.

\noindent
\textbf{Diffusion Model as Interface.}
Diffusion models are a flexible interface between users and visual tasks, facilitating a more intuitive interaction paradigm. InstructCV~\cite{gan2023instructcv} and InstructDiffusion~\cite{geng2023instructdiffusion} exemplify using of natural language instructions to guide visual generation and manipulation.
Pix2Seq v2~\cite{chen2022pix2seqv2} showcases the potential of diffusion models in generating sequences of visual tokens, bridging the gap between vision and language.

Different from these works, our VisionLLM v2 integrating LLMs extends vision generalist to support a broader range of vision-language tasks and explore various visual prompting paradigms, thereby significantly broadening the scope of application.


\vspace{-5px}
\section{\modelname}
\vspace{-5px}

\subsection{Model Design} 
\vspace{-5px}
\label{subsec:model_design}

\begin{figure}[t]
    \centering
    \includegraphics[width=1.0\textwidth]{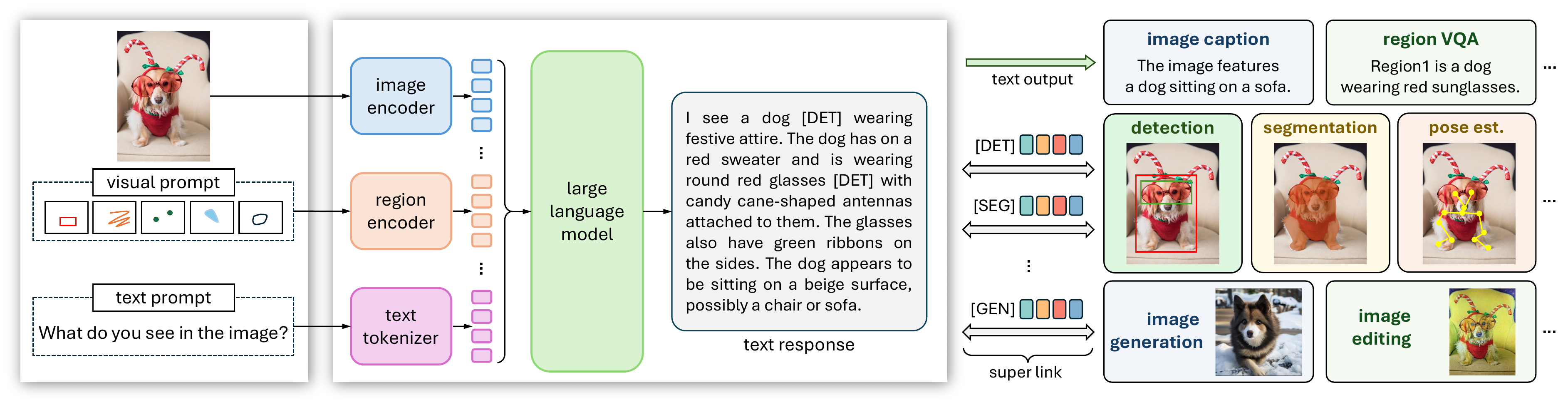}
    \caption{\textbf{Overall architecture of the proposed \modelname.} It receives the image and text/visual prompts as inputs. The central LLM parses the user instructions and generates the textual responses. Besides outputting the plain text, LLM can also output the special routing token such as \texttt{[DET]} when needed. The super-link queries would be automatically appended after the routing token embeddings and further processed by LLM. They play as the bridge for connecting LLM and task-specific decoders. In this way, our generalist model can support hundreds of visual tasks.
    } 
    \label{fig:arch}
    \vspace{-10pt}
\end{figure}

The overall architecture of \modelname is depicted in Figure~\ref{fig:arch}. It mainly consists of four parts: (1) an image encoder and a region encoder that encode the image-level and region-level information; 
(2) a large language model (LLM) that models the multimodal inputs and generates satisfactory textual responses; (3) a series of task-specific decoders for performing downstream tasks;
(4) a super link that uses routing tokens and super-link queries for efficient and conflict-free information transmission. We detail each component in the following.

\noindent
\textbf{Tokenization.} \modelname is flexible for handling multimodal input. (1) \emph{For text prompts}, we employ the text tokenizer to tokenize them into distinct vocabulary indices, which can be further processed by LLM and result in the text features $F_\text{text} \in \mathbb{R}^{L \times C}$, where $L$ denotes the length of input text, and $C$ is the channel dimension of LLM.

(2) \emph{For an image input}, we utilize a pre-trained vision foundation model, such as CLIP~\cite{radford2021clip}, to extract image features. Recognizing that current vision models operate the images at a low resolution,
we adopt the dynamic resolution approach~\cite{chen2024internvl_1_5} to process the input images. Specifically, the input image is first automatically matched to an optimal aspect ratio from a predefined ratio set. Subsequently, the image is scaled up to a higher resolution based on the selected aspect ratio and divided into $P$ square patches, each whose resolutions are 336$\times$336. These local patches, along with a 336$\times$336 global image $I_\text{global}$, are processed by the image encoder to capture both holistic scenes and fine-grained details, resulting in image features $F_\text{img} \in \mathbb{R}^{576(P+1) \times C}$. 

(3) \emph{For a visual prompt}, 
we employ binary masks to flexibly represent the visual prompts, such as point, box, scribble, and mask. To extract the region embedding, we first concatenate the binary mask with the input image along the channel dimension and then process it with three convolutional layers to downsample by a factor of 14 (see appendix for more details).
We further augment this feature map by adding the feature map of the global image $I_\text{global}$.
Finally, grid sampling is used to extract features within the masked regions, and these features are averaged to form the features of the visual prompt $F_\text{vprt} \in \mathbb{R}^{1\times C}$. 

\noindent
\textbf{Large Language Model.} Following previous works~\cite{liu2023llava, zhang2023gpt4roi, guo2024regiongpt}, both the images and visual prompts are projected to the feature space of the LLM. The LLM plays a central role in our model and is used to model multimodal inputs, parse user instructions, and generate appropriate responses. In this work, we adopt the commonly used Vicuna-7B~\cite{zheng2023judging} as the LLM in our network.

\noindent
\textbf{Task-specific Decoders.} To enhance the capacities of MLLM, we equip our model with several task-specific decoders. Specifically, we use Grounding DINO~\cite{liu2023groundingdino} for object-level localization. We additionally add a mask decoder upon it to obtain the segmentation ability. For pose estimation, we adopt UniPose~\cite{yang2023unipose} as the keypoint decoder.
Moreover, we incorporate Stable Diffusion~\cite{rombach2022sd} and InstructPix2Pix~\cite{brooks2023instructpix2pix} as the image decoders, endowing our model with the capability to generate and edit images. We discard these decoders' text encoders and link them with MLLM via the super link technique, which will be detailed explained in Section~\ref{subsec:super-link}. In this way, the decoders can be trained end-to-end with the entire network, ensuring the effective transmission of task information and increasing the 
openness
of these decoders.

\vspace{-5px}
\subsection{Super Link Technique} 
\vspace{-5px}
\label{subsec:super-link}

For the text-only output tasks, such as image-level and region-level VQA, we directly take the plain text generated by LLM as the final output. For visual perception and visual generation tasks, we propose the super link technique to tackle the challenge of selecting the appropriate decoder, avoiding task conflicts, and facilitating effective information transmission between the LLM and the decoders. The super link comprises two parts: 

(1) \emph{Routing Token}. We add the routing tokens, \textit{e.g.}, \texttt{[DET]}, \texttt{[POSE]}, \texttt{[SEG]}, \texttt{[GEN]}, \texttt{[EDIT]}, as special tokens to the original LLM vocabulary. When the model intends to complete the downstream task using one of the decoders, LLM would include the corresponding routing token in its textual response. To enable the model to discern which tasks to perform and which routing tokens to output, we construct a series of instruction templates for different tasks using ChatGPT~\cite{openai2023gpt4}. 

(2) \emph{Super-Link Queries}. For each decoder, we define the super-link  queries as a fixed set of embeddings denoted as $Q_\text{link} \in \mathbb{R}^{N \times C}$,
where $N$ is the number of queries. They are randomly initialized and serve as the bridge between LLM and task-specific decoders. Whenever the LLM predicts the routing token, the super-link queries would be automatically appended after the input embeddings of the routing token. We then extract their corresponding last-layer hidden states $H_\text{link}$ and apply an MLP projection to obtain $\hat{H}_\text{link}$. Finally, $\hat{H}_\text{link}$ is sent into the specific decoders as a condition to perform the downstream tasks. In the following, we illustrate how to integrate $\hat{H}_\text{link}$ into decoders for visual perception and generation, respectively.

\noindent
\textbf{Visual Perception} covers a wide range of visual tasks, such as open-ended/closed-set object detection, instance segmentation, pose estimation, \etc. \modelname supports using both text and visual prompts to define these tasks. We list an example in the following figure. \texttt{<image>} and \texttt{<region>} are the placeholders that will be replaced by image and region embeddings before being fed into the LLM.
Here, we take Example 1 of interactive segmentation for clarification.
The user prompts the model to segment specific regions within a question. MLLM sequentially lists the region names followed by a routing token \texttt{[SEG]} in the response. Remember that the proposed method would automatically append the super-link queries after the routing token. In that way, we can obtain the per-region representations by extracting the output hidden states of MLLM from corresponding super-link queries and pooling them into one embedding. These embeddings are fed into a segmentation decoder as the conditional feature, requiring only a single forward to produce segmentation results for all regions. In the following, we show a template example for interactive segmentation.

\noindent
\begin{minipage}{\textwidth}
\begin{tcolorbox} 
\footnotesize
\vspace{-1mm}

\textbf{\textcolor{blue}{Example1: Text Prompt + Visual Prompt for Interactive Segmentation.}} 

\textbf{USER}: <image> Could you please segment all the corresponding objects according to the visual prompts as region1 <region>, region2 <region>?

\textbf{ASSISTANT}: Sure, these objects are region1 [SEG], region2 [SEG].

\vspace{-1mm}
\end{tcolorbox}
\end{minipage}

\noindent
\textbf{Visual Generation} is also a wide topic covering a number of different tasks, such as generation, editing, variation, personalization, \etc. In \modelname, we focus on two fundamental tasks, \ie, text-to-image generation and instruction-based image editing. We use Stable Diffusion v1.5 (SD) as our tool in the text-to-image generation task. We abandon its text encoder and 
use 
the output hidden states of the MLLM as the image generation condition for SD.
Image editing task~\cite{kirillov2023sam} can also be accomplished in the same paradigm by using both image and text prompts as inputs. In the following, we list a template example for text-to-image generation.

\noindent
\begin{minipage}{\textwidth}
\begin{tcolorbox} 
\footnotesize
\vspace{-1mm}
\textbf{\textcolor{blue}{Example 2: Text Prompt for Text-to-Image Generation.}} 

\textbf{ASSISTANT}: Of course, here it is [GEN].

\vspace{-1mm}
\end{tcolorbox}
\end{minipage}

\noindent
\textbf{\textit{Discussion.}} Some previous works have used the special token or learnable queries independently. InstructBLIP~\cite{instructblip}, ep-ALM~\cite{shukor2023ep-alm}, and MAPL~\cite{manas2022mapl} use learnable queries (i.e., soft prompts) to connect the modality encoders and LLM. FROMAGe~\cite{koh2023fromage} uses a special token for image-text retrieval so as to handle multimodal outputs, where the images are not generated from the network end-to-end. However, these works still remain constrained to text-based outputs. The proposed super link is the seamless integration of the two techniques. Despite the simplicity of our method, it is able to extend MLLMs to handle hundreds of tasks by largely extending the output formats, \eg, box, mask, keypoint and image. Meanwhile, it can address several challenges when scaling up various tasks: (i) precise decoder invocation, (ii) mitigating task conflicts and (iii) efficient message transmission in an end-to-end manner.

\vspace{-5px}
\subsection{Training Strategy} 
\label{subsec:training_strategy}
\vspace{-5px}

Current MLLMs~\cite{lai2023lisa, zhang2024psalm, dong2023dreamllm} face reduced conversational abilities when augmented with additional capacities. To create a generalist model capable of handling hundreds of tasks without compromising vision understanding, we propose a three-stage training strategy, where the first stage focuses on building an MLLM with strong image-level and region-level vision understanding. In the subsequent stages, we add task-specific decoders and continue training to equip the model with advanced capabilities.

\noindent
\textbf{Stage-1: Mutimodal Training.} In the first stage, we follow the training settings of LLaVA~\cite{liu2023llava, liu2023llava1.5}, comprising pre-training and instruction tuning phases. The pre-training phase aims to establish the image-level and region-level vision-language alignment, where only the region encoder and the projections for image embedding and region embedding are trained for efficiency. The instruction tuning phase unfreezes the LLM and trains the model on a wide range of high-quality instruction data. After the training in this stage, we can obtain a strong MLLM with excellent conversation ability, which we term as \textbf{\modelname-Chat}.

\noindent
\textbf{Stage-2: Multi-capacity Fine-tuning.} At this stage, we integrate task-specific decoders into the model and perform multi-task joint training. In addition to the instruction data utilized in stage-1, we incorporate extensive visual datasets such as COCO~\cite{lin2014coco}, ADE20K~\cite{zhou2017ade20k} for their specific tasks. We construct a series of instruction templates for these visual datasets to perform instruction tuning, ensuring that the LLM can accurately invoke the downstream decoders. During this stage, the region encoder and all decoders undergo training, and we only finetune the input and output embeddings of the LLM to maximally preserve its original conversational ability.

\noindent
\textbf{Stage-3: Decoder-only Fine-tuning.} Since the decoders cannot converge within a single epoch, we further train the decoders for 12 epochs using visual datasets while freezing all other components. It is noted that the super-link queries continue to be trained during this stage. After finishing the three-stage training, our model has diverse capacities for visual tasks while maintaining effectiveness in global vision understanding, named \textbf{\modelname}.

\vspace{-5px}
\section{Experiments}
\vspace{-5px}

\subsection{Implementation Details}
\vspace{-5px}

\noindent
\textbf{Dataset Details.} To support the joint training of our model, we meticulously collect and re-organize the datasets across a wide range of tasks from publicly available sources. For the first stage training, we utilize a substantial amount of high-quality instruction data for both image-level and region-level visual question answering, including ShareGPT4V~\cite{chen2023sharegpt4v}, All-Seeing~\cite{wang2023allseeing}, VQAv2~\cite{goyal2017vqav2}, \etc. In the last two stages, we further incorporate extensive visual datasets, \textit{e.g.}, COCO~\cite{lin2014coco}, RefCOCO/+/g~\cite{yu2016refcoco, mao2016refcocog}, LAION-Aesthetics~\cite{laion_aes}, to enhance our model with numerous capacities. These datasets encompass multiple tasks such as object detection, pose estimation, image generation, and span various domains such as natural scenes, remote sensing images, medical images, \etc. To facilitate the training of diverse datasets in our MLLM framework, we construct a series of instruction templates for different tasks, which are completely listed in the Appendix. Additionally, we also collect a multimodal dataset termed \textbf{MMIC}
focusing on visual prompting and in-context learning. The data in our MMIC comes from various sources, including fine-grained visual recognition, object detection, instance segmentation, and pose detection. 
We elaborate on all datasets used in this work as well as the dataset construction of MMIC in the Appendix.

\noindent
\textbf{Model Details.}
We adopt the CLIP-L/14~\cite{radford2021clip} as the image encoder and Vicuna-7B-v1.5~\cite{zheng2023judging} as the language model. Grounding-DINO~\cite{liu2023groundingdino} and UniPose~\cite{yang2023unipose} are selected as object decoder and keypoint decoder, respectively. And for these two decoders, we experiment with Swin-T~\cite{liu2021swin} backbone.
Additionally, image decoders are kept as Stable Diffusion v1.5~\cite{rombach2022sd} for image generation and InstructPix2Pix~\cite{brooks2023instructpix2pix} for image editing. All these components load the pre-trained weights while the region encoder is randomly initialized. 
For visual perception and visual generation tasks, the number $N$ of super-link queries is set to 4 and 64, respectively.
During training, we adjust the dataloader so that each GPU processes samples from only one dataset. More training details are provided in the Appendix.

In the following subsections, we present the experimental results to cover as many tasks, interactive modes, and domains. It is noted that all the results of our method are reported using \textbf{a single generalist model} with the same parameters.
More results can be found in the Appendix.

\vspace{-5px}
\subsection{Mutimodal Benchmarks}
\vspace{-5px}
\begin{table*}[t!]
\scriptsize
\centering
\setlength\tabcolsep{1.7pt}
\renewcommand{\arraystretch}{1}

\begin{tabular}{l|cc|ccccc|cccc}

  &  \multirow{2}{*}{\begin{tabular}[l]{@{}l@{}}visual \\ encoder\end{tabular}}
  & 
  & \multicolumn{5}{c|}{academic-oriented datasets} 
  & \multicolumn{4}{c}{instruction-following datasets} \\
\multirow{-2}{*}{model} & & \multirow{-2}{*}{LLM} & VQA$^{\text{v2}}$ & GQA & VizWiz & SQA$^{\text{I}}$ & VQA$^{\text{T}}$ & POPE & MME & MMB-EN/CN & SEED\\

\hline

InstructBLIP-7B~\cite{instructblip}  & EVA-g   & Vicuna-7B   & -        & 49.2       & 34.5 & 60.5  & 50.1 & -    & -      & 36.0 / 23.7  & 53.4 / 58.8 / 38.1       \\
InstructBLIP-13B~\cite{instructblip} & EVA-g   & Vicuna-13B   & -        & 49.5       & 33.4 & 63.1  & 50.7 & 78.9 & 1212.8 & - / -        & - / - / -          \\
Shikra~\cite{chen2023shikra}         & CLIP-L  & Vicuna-13B  & 77.4$^*$ & -          & -    & -     & -    & -    & -      & 58.8 / -~~~~~~ & - / - / -          \\
IDEFICS-80B~\cite{idefics2023}       & CLIP-H  & LLaMA-65B   & 60.0~~   & 45.2       & 36.0 & -     & 30.9 & -    & -      & 54.5 / 38.1  & - / 53.2 / -          \\
Qwen-VL-Chat~\cite{bai2023qwenvl}    & CLIP-G  & Qwen-7B     & 78.2$^*$ & ~~57.5$^*$ & 38.9 & 68.2  & 61.5 & -    & 1487.5 & 60.6 / 56.7  & 58.2 / 65.4 / 37.8       \\
InternVL-7B~\cite{chen2023internvl}  & ViT-6B & Vicuna-7B   & 79.3$^*$ & ~~62.9$^*$ & 52.5 & 66.2  & 57.0 & 86.4 & 1525.1 & 64.6	/ 57.6  & 60.2 / - / -~~~~~~          \\ 
InternVL-13B~\cite{chen2023internvl} & ViT-6B & Vicuna-13B   & 80.2$^*$ & ~~63.9$^*$ & 54.6 & 70.1  & 58.7 & 87.1 & 1546.9 & 66.5	/ 61.9	& 62.4 / - / -~~~~~~         \\ 
LLaVA-1.5-7B~\cite{liu2023llava1.5}  & CLIP-L & Vicuna-7B    & 78.5$^*$ & ~~62.0$^*$ & 50.0 & 66.8  & 58.2 & 85.9 & 1510.7 & 64.3 / 58.3  & 58.6 / 66.1 / 37.3 \\
LLaVA-NeXT-7B~\cite{liu2024llava-next} & CLIP-L & Vicuna-7B  & 81.8$^*$ & ~~64.2$^*$ & 57.6 & 70.1  & 64.9 & 86.5 & 1519.0 & 67.4 / 60.6  & - / 70.2 / -       \\
LLaVA-NeXT-13B~\cite{liu2024llava-next}& CLIP-L & Vicuna-13B & 82.8$^*$ & ~~65.4$^*$ & 60.5 & 73.6  & 67.1 & 86.2 & 1575.0 & 70.0 / 64.4  & - / 71.9 / -       \\

\hline
\rowcolor{gray!15}
\modelname-Chat     & CLIP-L & Vicuna-7B & 81.4$^*$ & ~~65.1$^*$ & 54.6 & ~~94.4$^*$ & 66.3 & 87.5 & 1512.5 & 77.1 / 67.6 & 65.4 / 71.7 / 41.6  \\
\rowcolor{gray!15}
\modelname  & CLIP-L & Vicuna-7B & 80.8$^*$ & ~~65.1$^*$ & 51.8 & ~~94.2$^*$ & 64.7 & 88.8 & 1495.6 & 76.3 / 66.8 & 65.6 / 71.7 / 42.2  \\

\end{tabular}

\caption{\textbf{Comparison with SoTA models on multimodal dialogue benchmarks.}
The academic-oriented datasets include: VQAv2 test-dev~\cite{goyal2017vqav2}, GQA test-balanced~\cite{hudson2019gqa}, VizWiz test-dev~\cite{gurari2018vizwiz}, ScienceQA test~\cite{saikh2022scienceqa} and TextVQA val~\cite{singh2019textvqa}. The instruction-following datasets include: POPE~\cite{li2023pope}, MME~\cite{fu2023mme}, MMBench-EN/CN~\cite{liu2023mmbench}, SEED-Bench (all/image/video)~\cite{ge2024seed}. 
$^*$The training annotations of the dataset are observed during training. 
}
\label{tab:llava_benchmarks}
\vspace{-10px}
\end{table*}

\noindent
\textbf{Multimodal Dialogue.}
We first evaluate our models on academic-oriented VQA datasets and recent instruction-following datasets for MLLMs, as presented in Table~\ref{tab:llava_benchmarks}. The results clearly demonstrate that our models outperform previous methods under the same parameter scale, particularly on the instruction-following datasets. For instance, \modelname-Chat surpasses LLaVA-NeXT-7B~\cite{liu2024llava-next} by +9.7 and +7.0 points on MMBench-EN/CN~\cite{liu2023mmbench}, respectively. 
Additionally, we find that \modelname achieves comparable performance to \modelname-Chat on these multimodal benchmarks and even performs better on some benchmarks, such as POPE~\cite{li2023pope}, a popular benchmark for evaluating object hallucination. This phenomenon indicates that our framework effectively mitigates the issue of multi-task conflict and maintains proficiency in conversational ability.

\begin{table*}[t!]
\scriptsize
\centering
\renewcommand{\arraystretch}{1}

\begin{subtable}{0.50\linewidth}
    \centering
    \setlength\tabcolsep{4pt}
    \begin{tabular}{l|cc|cc|cc}
           & \multicolumn{2}{c|}{COCO} & \multicolumn{2}{c|}{LVIS} 
           & \multicolumn{2}{c}{PACO}  \\
        \multirow{-2}{*}{method} & mAP & Acc (\%) & SS & S-IoU & SS & S-IoU \\

    \hline
    CLIP~\cite{radford2021clip} & 58.9 & - & - & - & - & - \\
    RegionCLIP~\cite{zhong2022regionclip} & 58.3 & - & - & - & - & - \\
    LLaVA~\cite{liu2023llava} & - & 40.0 & 49.9 & 19.8 & 42.2 & 14.6 \\
    Shikra~\cite{chen2023shikra} & - & 53.9 & 49.7 & 19.8 & 43.6 & 11.4 \\
    GPT4RoI~\cite{zhang2023gpt4roi} & - & 64.0 & 51.3 & 12.0 & 48.0 & 12.1 \\
    ASM~\cite{wang2023allseeing} & 69.3 & - & - & - & - & - \\
    RegionGPT~\cite{guo2024regiongpt} & 70.0 & 80.6 & - & - & - & - \\
    Osprey~\cite{yuan2023osprey} & - & - & 65.2 & 38.2 & 73.1 & 52.7 \\
    
    \hline
    \rowcolor{gray!15}
    \modelname-Chat & 81.8 & 90.5 & 67.3 & 42.7 & 63.8 & 36.3 \\
    \rowcolor{gray!15}
    \modelname & 81.9 & 90.4 & 73.0 & 51.3 & 70.9 & 47.6  \\

    \end{tabular}
    \caption{Region Recognition}
    \label{subtab:region_rec}
\end{subtable}%
\hspace{5.1em}
\begin{subtable}{0.40\linewidth}
    \centering
    \setlength\tabcolsep{4pt}
    \begin{tabular}{l|ccc}
                                 & \multicolumn{3}{c}{Val Acc (\%)} \\
        \multirow{-2}{*}{method} & Q$\rightarrow$A & QA$\rightarrow$R & Q$\rightarrow$AR \\
        \hline
        ViLBERT~\cite{lu2019vilbert}        & 72.4  & 74.5  & 54.0     \\
        Unicoder-VL~\cite{li2020unicoder}   & 72.6  & 74.5  & 54.5     \\
        VLBERT~\cite{su2019vlbert}         & 75.5  & 77.9  & 58.9     \\
        ERNIE-ViL-L~\cite{yu2021ernie}      & 78.5  & 83.4  & 65.8     \\
        VILLA~\cite{gan2020villa}           & 78.5  & 82.6 & 65.2      \\
        GPT4RoI-7B$^*$~\cite{zhang2023gpt4roi}  & 87.4  & 89.6  & 78.6   \\
        ASMv2~\cite{wang2024allseeingv2}    & 87.8  & 88.8  & 78.4     \\ 
        ASMv2$^*$~\cite{wang2024allseeingv2} & 88.4 & 89.9 & 79.4      \\

        \hline
        \rowcolor{gray!15}
        \modelname-Chat                  & 90.0 & 91.9 & 82.9 \\
        \rowcolor{gray!15}
        \modelname               & 89.8 & 91.7 & 82.5 \\
        
    \end{tabular}
    \caption{Visual Commonsense Reasoning}
    \label{subtab:region_vcr}
\end{subtable}%

\vspace{-0.5em}
\caption{\textbf{Comparison of region recognition and visual commonsense reasoning performance.} 
(a) SS and S-IoU represent semantic similarity and semantic IoU, which originated from ~\cite{yuan2023osprey}. (b) Q, A, and R denote question, answer, and rationale, respectively. X$\rightarrow$Y means that the model needs to select option Y conditioned on X. $^*$The model is finetuned on the dataset.
}
\label{tab:region_rec_vcr}
\vspace{-10px}
\end{table*}

\noindent
\textbf{Region Recognition.}
The region recognition task needs the model to identify the object category given the ground-truth bounding box. We compare our method with both feature-based and text-output approaches in Table~\ref{subtab:region_rec}. Feature-based methods, such as RegionCLIP~\cite{zhong2022regionclip} and ASM~\cite{wang2023allseeing}, compute similarity scores between region visual features and candidate category text features. In contrast, text-output~\cite{chen2024allava, guo2024regiongpt, yuan2023osprey} directly predict the category name using a single word or phrase, embracing the advantage of openness. As shown in the table, our models demonstrate the significant superior performance on COCO~\cite{lin2014coco}, long-tail LVIS~\cite{gupta2019lvis} and part-level PACO~\cite{ramanathan2023paco} datasets.

\noindent
\textbf{Visual Commonsense Reasoning.}
Visual commonsense reasoning (VCR) requires the model to possess strong region-level question-answering and reasoning abilities, as it needs to select not only the correct answer but also the correct rationale behind it. We present the comparison results on the VCR dataset~\cite{zellers2019vcr} in Table~\ref{subtab:region_vcr}. Without task-specific fine-tuning, \modelname-Chat
achieves an accuracy of 82.9\% in the crucial Q$\rightarrow$AR task, which precedes the previous best model, ASMv2~\cite{wang2024allseeingv2}, by +3.5 points. \modelname also outperforms the previous methods for all the metrics, highlighting the promising common sense reasoning capability of our model.

\vspace{-5px}
\subsection{Visual Perception Tasks} 
\vspace{-5px}

\begin{table*}[t!]
\scriptsize
\centering
\renewcommand{\arraystretch}{0.98}
\setlength\tabcolsep{4.9pt}

\begin{tabular}{l|c|l|ccc|ccc|ccc}
       &      &    & \multicolumn{3}{c|}{detection (COCO)} 
       & \multicolumn{3}{c|}{instance seg. (COCO)}
       & \multicolumn{3}{c}{detection (CrowdHuman)}
       \\
    \multirow{-2}{*}{method} & \multirow{-2}{*}{type} & \multirow{-2}{*}{backbone} &
    AP & AP$\rm_{50}$ & AP$\rm_{75}$ & AP & AP$\rm_{50}$ & AP$\rm_{75}$ & AP$\rm_{50}$ & mMR$\downarrow$ & Recall \\
    \hline
    Deformable-DETR~\cite{zhu2020deformable} & & ResNet50 & 46.2 & 65.2 & 50.0 & - & - & - & 89.1 & 50.0 & 95.3 \\
    DDQ~\cite{zhang2023ddq} & & ResNet50 & 52.0 & 69.5 & 57.2 & - & - & - & 93.8 & 39.7 & 98.7 \\
    ViTDet-B~\cite{li2022vitdet} &  & ViT-B & 56.0 & - & - & 48.0  & - & - & - & - & - \\
    Grounding DINO~\cite{liu2023groundingdino} &  & Swin-T & 57.2 & - & - & - & - & - & - & - & - \\
    Mask2Former~\cite{cheng2022mask2former}  &  & ResNet50 & - & - & - & 43.7 & - & - & - & - & - \\
    Mask DINO~\cite{li2022maskdino}  & \multirow{-6}{*}{Specialist} & ResNet50 & 51.7 & - & - & 46.3 & - & - & - & - & - \\
    \hline
    UniHCP$^*$~\cite{ci2023unihcp} & & ViT-B & - & - & - & - & - & - & 92.5 & - & - \\
    Hulk~\cite{wang2023hulk} & & ViT-L & - & - & - & - & - & - & 92.2 & - & - \\
    Hulk$^*$~\cite{wang2023hulk} & & ViT-L & - & - & - & - & - & - & 93.0 & - & - \\
    Pix2Seq v2~\cite{chen2022pix2seqv2} & & ViT-B & 46.5 & - & - & 38.2 & - & - & - & - & - \\
    VisionLLM~\cite{wang2023visionllm}  & & ResNet50 & 44.8 & 64.1 & 48.5 & 25.2 & 50.6 & 22.4 & - & - & - \\
    Uni-Perceiver-v2~\cite{li2023uniperceiverv2} & & Swin-B & 58.6 & - & - & 50.6 & - & - & - & - & - \\
    UNINEXT~\cite{yan2023uninext} & & ResNet50 & 51.3 & 68.4 & 56.2 & 44.9 & 67.0 & 48.9 & - & - & - \\
    GLEE-Lite~\cite{wu2023general} & & ResNet50 & 55.0 & - & - & 48.4 & - & - & - & - & - \\
    GLEE-Plus~\cite{wu2023general} & & Swin-L   & 60.4 & - & - & 53.0 & - & - & - & - & - \\

    \rowcolor{gray!15}
    \modelname & \multirow{-10}{*}{Generalist} & Swin-T & 56.7 & 74.5 & 62.2 & 47.8 & 71.8 & 52.0 & 93.1 & 44.7 & 98.5 \\
\end{tabular}

\caption{\textbf{Comparison of object detection and instance segmentation performance.} 
Instance seg. means instance segmentation.
$^*$The model is finetuned on the dataset.
}
\label{tab:detection}
\vspace{-12px}
\end{table*}
\begin{table*}[t!]
\scriptsize
\centering
\renewcommand{\arraystretch}{1}
\setlength\tabcolsep{2.3pt}

\begin{tabular}{l|c|c|ccccc|cccc}
     &  & & \multicolumn{5}{c|}{AP$\uparrow$} & \multicolumn{4}{c}{PCK@0.2$\uparrow$} \\ 
    
    \multirow{-2}{*}{method} & \multirow{-2}{*}{type} & \multirow{-2}{*}{backbone} & 
    COCO & CrowdPose & AP-10K & Human-Art & Macaque& 300W  & AnimalKingdom   & Fly  & Locust \\ \hline

    ViTPose++~\cite{Xu2022ViTPose++}  &  & ViT-S   & 75.8   & - & ~~71.4$^*$   & 23.4      & ~~15.5$^*$   & ~~95.2$^*$   & -    & -    & -        \\ 
    ED-Pose~\cite{Yang2023edpose}  & \multirow{-2}{*}{Specialist} & Swin-T     & 73.3  & - & 45.5   & 71.3      & 51.0   & -      & -    & -    & -       \\
    \hline
    
    UniPose-T ~\cite{yang2023unipose}  & & Swin-T   & 74.4 & - & 74.0   & 72.5      & 78.0   & 98.1   & 95.3 & 99.6 & 99.7  \\
    UniPose-V ~\cite{yang2023unipose} & & Swin-T   & 74.3  & - & 73.6   & 72.1      & 77.3   & 99.4  & 94.3 & 99.8 & 99.6   \\ 

    \rowcolor{gray!15}
    \modelname  & \multirow{-3}{*}{Generalist} & Swin-T  & 74.0 & 79.4 & 76.8   & 72.9      & 81.9   & 91.1   & 94.4 & 99.4 & 97.8 \\
\end{tabular}

\caption{\textbf{Comparison of pose estimation performance.} $^*$ indicates that the results rely on ground-truth bounding boxes for top-down methods.
}
\label{tab:pose}
\vspace{-3px}
\end{table*}

\begin{figure}[t!]
    \centering
    \includegraphics[width=0.95\textwidth]{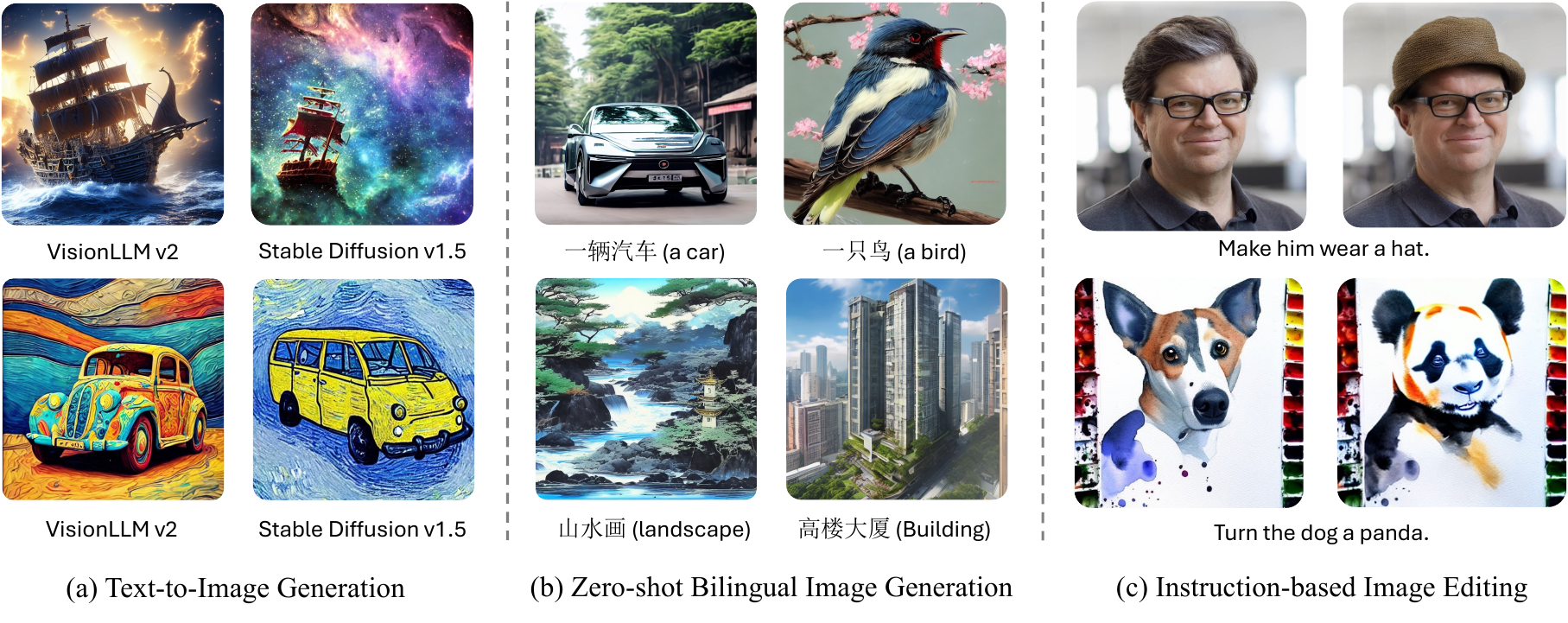}
    \vspace{-10pt}
    \caption{\textbf{Qualitative results of image generation and image editing.} 
    The prompts for text-to-image generation are ``Pirate ship trapped in a cosmic maelstrom nebula'' and ``A car in the style of van Gogh.''
    }
    \vspace{-8pt}
    \label{fig:gen}
    
\end{figure}

\noindent
\textbf{Object Detection and Instance Segmentation.} 
In Table~\ref{tab:detection}, we compare the results of \modelname with state-of-the-art methods on two fundamental vision tasks, \textit{i.e.}, object detection, and instance segmentation. As can be seen, using the lightweight backbone Swin-T, our generalist model achieves the performance of 56.7 AP$_{\rm b}$ and 47.8 AP$_{\rm m}$ on COCO. The results significantly outperform the previous methods using ResNet50~\cite{he2016deep} backbone and are comparable with the specialist model Grounding-DINO-T~\cite{liu2023groundingdino}. Moreover, we also validate our model on the crowded pedestrian detection dataset, \textit{i.e.}, CrowdHuman. \modelname surpasses the previous best generalist model Hulk~\cite{wang2023hulk} by 0.9 points on AP$\rm_{50}$.

\noindent
\textbf{Pose Estimation.} 
We present the results on the multiple pose estimation dataset in Table~\ref{tab:pose}. While most previous methods~\cite{Yang2023edpose, Xu2022ViTPose++, liu2023grouppose} only focus on the person scenes, our \modelname is effective in performing the keypoint detection for multiple objects. As shown in the table, our model achieves competitive performance with UniPose-T~\cite{yang2023unipose} using the same Swin-T backbone. Especially, our model demonstrates the superior performance on AP-10K~\cite{yu2021ap10k} and Macaque~\cite{labuguen2021macaquepose} datasets and sets a new state-of-the-art result on CrowdPose~\cite{li2019crowdpose}. These results prove the effectiveness of our model for pose estimation.


\vspace{-5px}
\subsection{Visual Generation Tasks}
\vspace{-5px}

\noindent
We evaluate the generation capabilities of our model on two tasks, \ie, text-to-image generation and instruction-based image editing. In Figure~\ref{fig:gen}, we demonstrate that even if our model uses Stable Diffusion v1.5 as an image decoder, it achieves better visual quality than SD v1.5 with better conditional embedding produced by LLM. Moreover, the use of LLM for conditional encoding of user instructions makes it possible to benefit from the merits of LLM. For example, our model trained on English data is able to perform zero-shot bilingual image generation. Besides, we show the qualitative results of applying our model for instruction-based image editing, which also achieves appealing performance in a unified approach.

\noindent

\vspace{-10px}
\subsection{Ablation Study}
\vspace{-5px}

\begin{figure}[t]
\centering

    \begin{minipage}[t]{0.55\textwidth}
        \vspace{0pt}
        \centering
        \scriptsize
        \setlength\tabcolsep{2.0pt}
        \resizebox{1.0\linewidth}{!}{

\begin{tabular}{l|c|cccccc}
    & & \multicolumn{2}{c}{inst seg.} & ground. & pose & \multicolumn{2}{c}{interact seg.} \\
    \multirow{-2}{*}{method} 
    & \multirow{-2}{*}{\begin{tabular}[c]{@{}c@{}}query/token\\ number\end{tabular}} 
    & AP$_{\rm b}$ & AP$_{\rm m}$ & P@.5 & AP & mIoU & cIoU \\
    \hline
    
     & 1 & 50.4 & 39.6 & 85.8 & 43.0 & 43.2 & 60.0 \\
     \rowcolor{gray!15}
     & 4 & 52.0 & 41.0 & 85.7 & 71.0 & 44.8 & 60.4 \\
     \multirow{-3}{*}{super-link queries} & 8 & 52.1 & 40.7 & 86.4 & 71.6 & 45.9 & 61.9 \\

\end{tabular}

        }
        \vspace{2pt}
        \captionsetup{width=1\textwidth}
        \captionof{table}{\textbf{Ablation on the super-link queries number.} 
        We evaluate the results on the four crucial visual perception tasks: instance segmentation (COCO), visual grounding (RefCOCO), pose estimation (COCO), and interactive segmentation (COCO using scribble). Our default setting is marked in \graybox{gray}.
        }
        \label{tab:ablation_perception}
    \end{minipage}
    \hspace{1em}
    \begin{minipage}[t]{0.4\textwidth}
        \vspace{-4pt}
        \centering
        \includegraphics[width=1\textwidth]{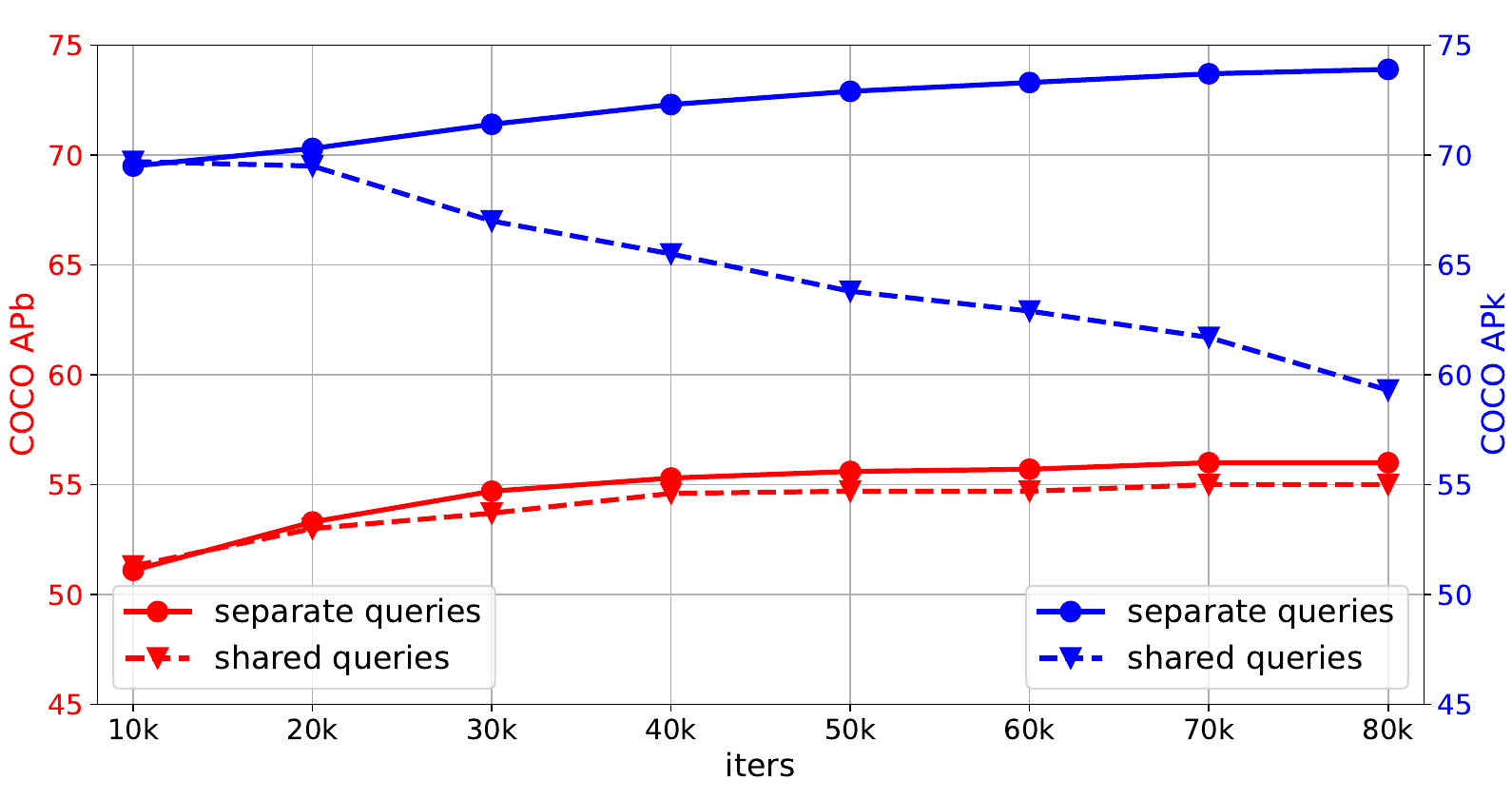}
        \vspace{-3ex}
        \captionsetup{width=1.0\linewidth}
        \caption{\textbf{Shared \vs~ unshared super-link queries for different decoders.}
        We report the box/keypoint AP on COCO.
        }
        \label{fig:shared_queries}
    \end{minipage}
    \vspace{-15px}
\end{figure}

In the ablation studies, we follow the training setup of stage-2 unless otherwise specified. We only train our model on the crucial visual perception tasks, \textit{i.e.}, instance segmentation, visual grounding, pose estimation, and interactive segmentation, for rapid validation.

\noindent
\textbf{Super-Link Queries Number.} We ablate the number $N$ of super-link queries in Table~\ref{tab:ablation_perception}. We observe that the performance of these tasks consistently improves with an increasing number of queries. This is reasonable as more queries can lead to richer and stronger representations.

\noindent
\textbf{Shared \vs  Unshared Super-Link Queries for Different Decoders.} 
To determine if one set of super-link queries is sufficient for all decoders, we conducted an ablation study by either using shared queries for all decoders or defining separate queries for each decoder. In this ablation, we only train the decoders and super-link queries while freezing all other components as the training setting of stage-3. In Figure~\ref{fig:shared_queries}, we plot the performance of box AP (using the object decoder) and keypoint AP (using the keypoint decoder) on COCO.  We observe that the keypoint AP would decrease over training when using shared queries, which may be attributed to the fact that most data are used for object decoder. Besides, the box AP with shared queries is also inferior to decoupled ones. Therefore, we define separate super-link queries for each decoder in our model. 

\begin{table}[t]
    \centering
    \begin{minipage}[t]{0.38\textwidth}
    \scriptsize
    \renewcommand{\arraystretch}{1.2}
    \setlength\tabcolsep{1.8pt}
    \resizebox{1.0\linewidth}{!}{
    
\begin{tabular}{c | c c c}

    Train / Test & Image VQA & Inst. Seg & Image Gen. \\
    \hline
    Image VQA & -0.01 & -0.11 & -0.04 \\
    Inst Seg. & +0.04 & -0.12 & + 0.19 \\
    Image Gen. & +0.03 & +0.02 & -0.04 \\

\end{tabular}
    }
    \vspace{6pt}
    \captionsetup{width=0.96\textwidth}
    \caption{\textbf{Ablation on the multi-task influence. } 
    The numbers denote the loss change when the model is finetuned on a single task.
    }
    \label{tab:ablation_multi-task}
    \end{minipage}
    \hspace{0.01\textwidth}
    \begin{minipage}[t]{0.58\textwidth}
    \scriptsize
    \renewcommand{\arraystretch}{1.2}
    \setlength\tabcolsep{2.5pt}
    \resizebox{1.0\linewidth}{!}{
    
\begin{tabular}{c | c c c | c | c}

     & TextVQA & MME & MMB EN/CN & COCO & COCO-Pose \\
    \hline
    One-stage & 53.2 & 1284.4 & 61.9 / 51.4 & 54.9 / 44.6 & 74.1 \\
    Three-stage & 66.2 & 1507.1 & 77.8 / 68.5 & 56.3 / 47.6 & 74.2 \\

\end{tabular}
    }
    \vspace{6pt}
    \captionsetup{width=0.95\textwidth}
    \caption{\textbf{Ablation on the one-stage and three-stage training.} We evaluate the models on image VQA, instance segmentation and pose estimation.
    }
    \label{tab:ablation_training-stage}
    \end{minipage}
\end{table}

\noindent
\textbf{Multi-Task Influence.} As indicated by previous works~\cite{zhu2022uniperceiver-moe, yu2020gradient}, different tasks with shared parameters may cause conflict with each other. This is mainly due to inconsistent optimization in multi-task learning. To investigate the mutual influence of multi-task joint training in our framework, we start from the same checkpoint and train the model on a single task (image VQA, instance segmentation, or image generation) for 1000 iterations. We record the loss change for all three tasks in Table~\ref{tab:ablation_multi-task}. In the table, a decrease in the loss value indicates beneficial training for the task, while an increase is detrimental. We can observe that training on image VQA is advantageous for all three tasks, which is reasonable as the conversation ability of MLLM is enhanced. Whereas training exclusively on instance segmentation or image generation leads to conflicts with other tasks. This aligns with the findings in Uniperceiver-MoE~\cite{zhu2022uniperceiver-moe}.

\noindent
\textbf{One-Stage \vs Three-Stage Training.} 
Some previous generalist models~\cite{git, bavishi2023fuyu} train the model in one stage. Our model encompasses much more tasks and thus introduces a training conflict: the MLLM requires only 1 epoch of training on chat data to prevent overfitting, whereas the decoders need longer training epochs (e.g., Grounding-DINO need 12 epochs of training on visual data) to achieve convergence. One possible solution for one-stage training is to give a higher sample ratio for the visual data. In the following, we conduct the ablation to study the effect of one-stage v.s. three-stage training. We use image-level chat data, COCO, and COCO-Pose for image understanding, instance segmentation, and pose estimation, respectively. For one-stage training, we repeat the COCO and COCO-Pose datasets 12 times. 
As can be seen from Table~\ref{tab:ablation_training-stage}, the conversation ability of the model is significantly decreased due to extreme data imbalance. And the performance of instance segmentation and pose estimation is also slightly reduced. These results prove the effectiveness of our three-stage training.

\vspace{-5px}
\section{Conclusion \& Limitation}
\vspace{-5px}
In this paper, we presented \modelname, a comprehensive MLLM that unifies visual perception, understanding, and generation within a single framework. 
The proposed super link mechanism facilitates flexible information transmission between the MLLM and task-specific decoders, addressing training conflicts and enhancing gradient feedback. Experiments show that \modelname achieves performance comparable to specialized models while maintaining broad applicability.

Regarding limitations, our model's training encompasses three stages, which are relatively complex. Moreover, the integration of downstream tools has only been preliminarily validated. Future work will further explore solutions to these issues, aiming to enhance the model's performance and efficiency.

\textbf{Broader Impact.} We envision that this work will further promote the fusion of visual and language tasks. In addition, since our work is built on open-source pre-trained vision foundation models and large language models, requiring low training resources, thus reducing the carbon footprint. 
We do not foresee obvious undesirable ethical/social impacts at this moment.

\section*{Acknowledgement}
\vspace{-5px}
This paper is supported by the National Key R\&D Program of China (No.2022ZD0161000), the General Research Fund of Hong Kong (No.17200622, 17209324), and the National Natural Science Foundation of China (No. 62376134, 62372223). Tong Lu and Zhe Chen are supported by the China Mobile Zijin Innovation Insititute (No. NR2310J7M). Zhe Chen is also supported by the Youth PhD Student Research Project under the National Natural Science Foundation (No. 623B2050).


{\small
\bibliographystyle{plain}
\bibliography{references}

\begin{thebibliography}{100}

\bibitem{laion_gpt4v}
Gpt-4v dataset.
\newblock \url{https://huggingface.co/datasets/laion/gpt4v-dataset}.

\bibitem{gpt4v}
Gpt-4v(ision) system card.
\newblock \url{https://cdn.openai.com/papers/GPTV_System_Card.pdf}.

\bibitem{laion_aes}
Laion-aesthetics.
\newblock \url{https://laion.ai/blog/laion-aesthetics/}.

\bibitem{openai2023gpt4}
Josh Achiam, Steven Adler, Sandhini Agarwal, Lama Ahmad, Ilge Akkaya, Florencia~Leoni Aleman, Diogo Almeida, Janko Altenschmidt, Sam Altman, Shyamal Anadkat, et~al.
\newblock Gpt-4 technical report.
\newblock {\em arXiv preprint arXiv:2303.08774}, 2023.

\bibitem{aghajanyan2022cm3}
Armen Aghajanyan, Bernie Huang, Candace Ross, Vladimir Karpukhin, Hu~Xu, Naman Goyal, Dmytro Okhonko, Mandar Joshi, Gargi Ghosh, Mike Lewis, et~al.
\newblock Cm3: A causal masked multimodal model of the internet.
\newblock {\em arXiv preprint arXiv:2201.07520}, 2022.

\bibitem{agrawal2019nocaps}
Harsh Agrawal, Karan Desai, Yufei Wang, Xinlei Chen, Rishabh Jain, Mark Johnson, Dhruv Batra, Devi Parikh, Stefan Lee, and Peter Anderson.
\newblock Nocaps: Novel object captioning at scale.
\newblock In {\em ICCV}, pages 8948--8957, 2019.

\bibitem{meta2024llama3}
Meta AI.
\newblock Introducing meta llama 3: The most capable openly available llm to date.
\newblock \url{https://ai.meta.com/blog/meta-llama-3/}, 2024.

\bibitem{alayrac2022flamingo}
Jean-Baptiste Alayrac, Jeff Donahue, Pauline Luc, Antoine Miech, Iain Barr, Yana Hasson, Karel Lenc, Arthur Mensch, Katherine Millican, Malcolm Reynolds, et~al.
\newblock Flamingo: a visual language model for few-shot learning.
\newblock {\em NeurIPS}, 35:23716--23736, 2022.

\bibitem{falcon40b}
Ebtesam Almazrouei, Hamza Alobeidli, Abdulaziz Alshamsi, Alessandro Cappelli, Ruxandra Cojocaru, Merouane Debbah, Etienne Goffinet, Daniel Heslow, Julien Launay, Quentin Malartic, Badreddine Noune, Baptiste Pannier, and Guilherme Penedo.
\newblock {Falcon-40B}: an open large language model with state-of-the-art performance.
\newblock 2023.

\bibitem{claude3series2024}
{Anthropic}.
\newblock The claude 3 model family: Opus, sonnet, haiku.
\newblock \url{https://www.anthropic.com}, 2024.

\bibitem{awadalla2023openflamingo}
Anas Awadalla, Irena Gao, Josh Gardner, Jack Hessel, Yusuf Hanafy, Wanrong Zhu, Kalyani Marathe, Yonatan Bitton, Samir Gadre, Shiori Sagawa, et~al.
\newblock Openflamingo: An open-source framework for training large autoregressive vision-language models.
\newblock {\em arXiv preprint arXiv:2308.01390}, 2023.

\bibitem{ba2016layer}
Jimmy~Lei Ba, Jamie~Ryan Kiros, and Geoffrey~E Hinton.
\newblock Layer normalization.
\newblock {\em arXiv preprint arXiv:1607.06450}, 2016.

\bibitem{qwen}
Jinze Bai, Shuai Bai, Yunfei Chu, Zeyu Cui, Kai Dang, Xiaodong Deng, Yang Fan, Wenbin Ge, Yu~Han, Fei Huang, Binyuan Hui, Luo Ji, Mei Li, Junyang Lin, Runji Lin, Dayiheng Liu, Gao Liu, Chengqiang Lu, Keming Lu, Jianxin Ma, Rui Men, Xingzhang Ren, Xuancheng Ren, Chuanqi Tan, Sinan Tan, Jianhong Tu, Peng Wang, Shijie Wang, Wei Wang, Shengguang Wu, Benfeng Xu, Jin Xu, An~Yang, Hao Yang, Jian Yang, Shusheng Yang, Yang Yao, Bowen Yu, Hongyi Yuan, Zheng Yuan, Jianwei Zhang, Xingxuan Zhang, Yichang Zhang, Zhenru Zhang, Chang Zhou, Jingren Zhou, Xiaohuan Zhou, and Tianhang Zhu.
\newblock Qwen technical report.
\newblock {\em arXiv preprint arXiv:2309.16609}, 2023.

\bibitem{bai2023qwenvl}
Jinze Bai, Shuai Bai, Shusheng Yang, Shijie Wang, Sinan Tan, Peng Wang, Junyang Lin, Chang Zhou, and Jingren Zhou.
\newblock Qwen-vl: A frontier large vision-language model with versatile abilities.
\newblock {\em arXiv preprint arXiv:2308.12966}, 2023.

\bibitem{bavishi2023fuyu}
Rohan Bavishi, Erich Elsen, Curtis Hawthorne, Maxwell Nye, Augustus Odena, Arushi Somani, and Sa{\u{g}}nak Ta{\c{s}}{\i}rlar.
\newblock Fuyu-8b: A multimodal architecture for ai agents, 2023.

\bibitem{berg2014birdsnap}
Thomas Berg, Jiongxin Liu, Seung Woo~Lee, Michelle~L Alexander, David~W Jacobs, and Peter~N Belhumeur.
\newblock Birdsnap: Large-scale fine-grained visual categorization of birds.
\newblock In {\em CVPR}, pages 2011--2018, 2014.

\bibitem{betker2023improving}
James Betker, Gabriel Goh, Li~Jing, Tim Brooks, Jianfeng Wang, Linjie Li, Long Ouyang, Juntang Zhuang, Joyce Lee, Yufei Guo, et~al.
\newblock Improving image generation with better captions.
\newblock {\em Computer Science. https://cdn. openai. com/papers/dall-e-3. pdf}, 2(3):8, 2023.

\bibitem{biten2019stvqa}
Ali~Furkan Biten, Ruben Tito, Andres Mafla, Lluis Gomez, Mar{\c{c}}al Rusinol, Ernest Valveny, CV~Jawahar, and Dimosthenis Karatzas.
\newblock Scene text visual question answering.
\newblock In {\em ICCV}, pages 4291--4301, 2019.

\bibitem{bossard2014food}
Lukas Bossard, Matthieu Guillaumin, and Luc Van~Gool.
\newblock Food-101--mining discriminative components with random forests.
\newblock In {\em ECCV}, pages 446--461. Springer, 2014.

\bibitem{brooks2023instructpix2pix}
Tim Brooks, Aleksander Holynski, and Alexei~A Efros.
\newblock Instructpix2pix: Learning to follow image editing instructions.
\newblock In {\em CVPR}, pages 18392--18402, 2023.

\bibitem{brown2020gpt3}
Tom Brown, Benjamin Mann, Nick Ryder, Melanie Subbiah, Jared~D Kaplan, Prafulla Dhariwal, Arvind Neelakantan, Pranav Shyam, Girish Sastry, Amanda Askell, et~al.
\newblock Language models are few-shot learners.
\newblock {\em NeurIPS}, 33:1877--1901, 2020.

\bibitem{cai2024internlm2}
Zheng Cai, Maosong Cao, Haojiong Chen, Kai Chen, Keyu Chen, Xin Chen, Xun Chen, Zehui Chen, Zhi Chen, Pei Chu, et~al.
\newblock Internlm2 technical report.
\newblock {\em arXiv preprint arXiv:2403.17297}, 2024.

\bibitem{cao2022geoqa+}
Jie Cao and Jing Xiao.
\newblock An augmented benchmark dataset for geometric question answering through dual parallel text encoding.
\newblock In {\em Proceedings of the 29th International Conference on Computational Linguistics}, pages 1511--1520, 2022.

\bibitem{chen2023understanding}
Aochuan Chen, Yuguang Yao, Pin-Yu Chen, Yihua Zhang, and Sijia Liu.
\newblock Understanding and improving visual prompting: A label-mapping perspective.
\newblock In {\em CVPR}, pages 19133--19143, 2023.

\bibitem{chen2024allava}
Guiming~Hardy Chen, Shunian Chen, Ruifei Zhang, Junying Chen, Xiangbo Wu, Zhiyi Zhang, Zhihong Chen, Jianquan Li, Xiang Wan, and Benyou Wang.
\newblock Allava: Harnessing gpt4v-synthesized data for a lite vision-language model.
\newblock {\em arXiv preprint arXiv:2402.11684}, 2024.

\bibitem{chen2023minigpt-v2}
Jun Chen, Deyao Zhu, Xiaoqian Shen, Xiang Li, Zechun Liu, Pengchuan Zhang, Raghuraman Krishnamoorthi, Vikas Chandra, Yunyang Xiong, and Mohamed Elhoseiny.
\newblock Minigpt-v2: large language model as a unified interface for vision-language multi-task learning.
\newblock {\em arXiv preprint arXiv:2310.09478}, 2023.

\bibitem{chen2023shikra}
Keqin Chen, Zhao Zhang, Weili Zeng, Richong Zhang, Feng Zhu, and Rui Zhao.
\newblock Shikra: Unleashing multimodal llm's referential dialogue magic.
\newblock {\em arXiv preprint arXiv:2306.15195}, 2023.

\bibitem{chen2023sharegpt4v}
Lin Chen, Jisong Li, Xiaoyi Dong, Pan Zhang, Conghui He, Jiaqi Wang, Feng Zhao, and Dahua Lin.
\newblock Sharegpt4v: Improving large multi-modal models with better captions.
\newblock {\em arXiv preprint arXiv:2311.12793}, 2023.

\bibitem{chen2023generalist}
Ting Chen, Lala Li, Saurabh Saxena, Geoffrey Hinton, and David~J Fleet.
\newblock A generalist framework for panoptic segmentation of images and videos.
\newblock In {\em ICCV}, pages 909--919, 2023.

\bibitem{chen2022pix2seqv2}
Ting Chen, Saurabh Saxena, Lala Li, Tsung-Yi Lin, David~J Fleet, and Geoffrey~E Hinton.
\newblock A unified sequence interface for vision tasks.
\newblock {\em NeurIPS}, 35:31333--31346, 2022.

\bibitem{chen2015cococaption}
Xinlei Chen, Hao Fang, Tsung-Yi Lin, Ramakrishna Vedantam, Saurabh Gupta, Piotr Doll{\'a}r, and C~Lawrence Zitnick.
\newblock Microsoft coco captions: Data collection and evaluation server.
\newblock {\em arXiv preprint arXiv:1504.00325}, 2015.

\bibitem{chen2020uniter}
Yen-Chun Chen, Linjie Li, Licheng Yu, Ahmed El~Kholy, Faisal Ahmed, Zhe Gan, Yu~Cheng, and Jingjing Liu.
\newblock Uniter: Universal image-text representation learning.
\newblock In {\em ECCV}, pages 104--120. Springer, 2020.

\bibitem{chen2024internvl_1_5}
Zhe Chen, Weiyun Wang, Hao Tian, Shenglong Ye, Zhangwei Gao, Erfei Cui, Wenwen Tong, Kongzhi Hu, Jiapeng Luo, Zheng Ma, et~al.
\newblock How far are we to gpt-4v? closing the gap to commercial multimodal models with open-source suites.
\newblock {\em arXiv preprint arXiv:2404.16821}, 2024.

\bibitem{chen2023internvl}
Zhe Chen, Jiannan Wu, Wenhai Wang, Weijie Su, Guo Chen, Sen Xing, Zhong Muyan, Qinglong Zhang, Xizhou Zhu, Lewei Lu, et~al.
\newblock Internvl: Scaling up vision foundation models and aligning for generic visual-linguistic tasks.
\newblock {\em arXiv preprint arXiv:2312.14238}, 2023.

\bibitem{cheng2022mask2former}
Bowen Cheng, Ishan Misra, Alexander~G Schwing, Alexander Kirillov, and Rohit Girdhar.
\newblock Masked-attention mask transformer for universal image segmentation.
\newblock In {\em CVPR}, pages 1290--1299, 2022.

\bibitem{cheng2014msra10k}
Ming-Ming Cheng, Niloy~J Mitra, Xiaolei Huang, Philip~HS Torr, and Shi-Min Hu.
\newblock Global contrast based salient region detection.
\newblock {\em TPAMI}, 37(3):569--582, 2014.

\bibitem{vicuna2023}
Wei-Lin Chiang, Zhuohan Li, Zi~Lin, Ying Sheng, Zhanghao Wu, Hao Zhang, Lianmin Zheng, Siyuan Zhuang, Yonghao Zhuang, Joseph~E. Gonzalez, Ion Stoica, and Eric~P. Xing.
\newblock Vicuna: An open-source chatbot impressing gpt-4 with 90\%* chatgpt quality, March 2023.

\bibitem{Chu2024MobileVLM}
Xiangxiang Chu, Limeng Qiao, Xinyu Zhang, Shuang Xu, Fei Wei, Yang Yang, Xiaofei Sun, Yiming Hu, Xinyang Lin, Bo~Zhang, and Chunhua Shen.
\newblock Mobilevlm v2: Faster and stronger baseline for vision language model.
\newblock {\em ArXiv}, abs/2402.03766, 2024.

\bibitem{ci2023unihcp}
Yuanzheng Ci, Yizhou Wang, Meilin Chen, Shixiang Tang, Lei Bai, Feng Zhu, Rui Zhao, Fengwei Yu, Donglian Qi, and Wanli Ouyang.
\newblock Unihcp: A unified model for human-centric perceptions.
\newblock In {\em CVPR}, pages 17840--17852, 2023.

\bibitem{clark2017docqa}
Christopher Clark and Matt Gardner.
\newblock Simple and effective multi-paragraph reading comprehension.
\newblock In {\em ACL}, pages 845--855, 2018.

\bibitem{cordts2016cityscapes}
Marius Cordts, Mohamed Omran, Sebastian Ramos, Timo Rehfeld, Markus Enzweiler, Rodrigo Benenson, Uwe Franke, Stefan Roth, and Bernt Schiele.
\newblock The cityscapes dataset for semantic urban scene understanding.
\newblock In {\em CVPR}, pages 3213--3223, 2016.

\bibitem{instructblip}
Wenliang Dai, Junnan Li, Dongxu Li, Anthony Meng~Huat Tiong, Junqi Zhao, Weisheng Wang, Boyang Li, Pascale~N Fung, and Steven Hoi.
\newblock Instructblip: Towards general-purpose vision-language models with instruction tuning.
\newblock {\em NeurIPS}, 36, 2024.

\bibitem{deepseek-llm}
DeepSeek-AI.
\newblock Deepseek llm: Scaling open-source language models with longtermism.
\newblock {\em arXiv preprint arXiv:2401.02954}, 2024.

\bibitem{dong2023dreamllm}
Runpei Dong, Chunrui Han, Yuang Peng, Zekun Qi, Zheng Ge, Jinrong Yang, Liang Zhao, Jianjian Sun, Hongyu Zhou, Haoran Wei, et~al.
\newblock Dreamllm: Synergistic multimodal comprehension and creation.
\newblock In {\em ICLR}, 2024.

\bibitem{CNFOOD241}
Bokun Fan.
\newblock Cnfood-241.
\newblock \url{https://data.mendeley.com/datasets/fspyss5zbb/1}.
\newblock [Accessed 12-08-2022].

\bibitem{fan2021cod10k}
Deng-Ping Fan, Ge-Peng Ji, Ming-Ming Cheng, and Ling Shao.
\newblock Concealed object detection.
\newblock {\em TPAMI}, 44(10):6024--6042, 2021.

\bibitem{fei2024vitron}
Hao Fei, Shengqiong Wu, Hanwang Zhang, Tat-Seng Chua, and Shuicheng Yan.
\newblock Vitron: A unified pixel-level vision llm for understanding, generating, segmenting, editing.

\bibitem{hao2024vitron}
Hao Fei, Shengqiong Wu, Hanwang Zhang, Tat-Seng Chua, and Shuicheng Yan.
\newblock Vitron: A unified pixel-level vision llm for understanding, generating, segmenting, editing.

\bibitem{fu2023mme}
Chaoyou Fu, Peixian Chen, Yunhang Shen, Yulei Qin, Mengdan Zhang, Xu~Lin, Zhenyu Qiu, Wei Lin, Jinrui Yang, Xiawu Zheng, et~al.
\newblock Mme: A comprehensive evaluation benchmark for multimodal large language models.
\newblock {\em arXiv preprint arXiv:2306.13394}, 2023.

\bibitem{fu2023guiding}
Tsu-Jui Fu, Wenze Hu, Xianzhi Du, William~Yang Wang, Yinfei Yang, and Zhe Gan.
\newblock Guiding instruction-based image editing via multimodal large language models.
\newblock {\em arXiv preprint arXiv:2309.17102}, 2023.

\bibitem{gan2023instructcv}
Yulu Gan, Sungwoo Park, Alexander Schubert, Anthony Philippakis, and Ahmed~M Alaa.
\newblock Instructcv: Instruction-tuned text-to-image diffusion models as vision generalists.
\newblock {\em arXiv preprint arXiv:2310.00390}, 2023.

\bibitem{gan2020villa}
Zhe Gan, Yen-Chun Chen, Linjie Li, Chen Zhu, Yu~Cheng, and Jingjing Liu.
\newblock Large-scale adversarial training for vision-and-language representation learning.
\newblock {\em NeurIPS}, 33:6616--6628, 2020.

\bibitem{ge2023planting}
Yuying Ge, Yixiao Ge, Ziyun Zeng, Xintao Wang, and Ying Shan.
\newblock Planting a seed of vision in large language model.
\newblock {\em arXiv preprint arXiv:2307.08041}, 2023.

\bibitem{ge2023making}
Yuying Ge, Sijie Zhao, Ziyun Zeng, Yixiao Ge, Chen Li, Xintao Wang, and Ying Shan.
\newblock Making llama see and draw with seed tokenizer.
\newblock {\em arXiv preprint arXiv:2310.01218}, 2023.

\bibitem{ge2024seed}
Yuying Ge, Sijie Zhao, Jinguo Zhu, Yixiao Ge, Kun Yi, Lin Song, Chen Li, Xiaohan Ding, and Ying Shan.
\newblock Seed-x: Multimodal models with unified multi-granularity comprehension and generation.
\newblock {\em arXiv preprint arXiv:2404.14396}, 2024.

\bibitem{geng2023instructdiffusion}
Zigang Geng, Binxin Yang, Tiankai Hang, Chen Li, Shuyang Gu, Ting Zhang, Jianmin Bao, Zheng Zhang, Han Hu, Dong Chen, et~al.
\newblock Instructdiffusion: A generalist modeling interface for vision tasks.
\newblock {\em arXiv preprint arXiv:2309.03895}, 2023.

\bibitem{goyal2017vqav2}
Yash Goyal, Tejas Khot, Douglas Summers-Stay, Dhruv Batra, and Devi Parikh.
\newblock Making the v in vqa matter: Elevating the role of image understanding in visual question answering.
\newblock In {\em CVPR}, pages 6904--6913, 2017.

\bibitem{graving2019deepposekit}
Jacob~M Graving, Daniel Chae, Hemal Naik, Liang Li, Benjamin Koger, Blair~R Costelloe, and Iain~D Couzin.
\newblock Deepposekit, a software toolkit for fast and robust animal pose estimation using deep learning.
\newblock {\em Elife}, 8:e47994, 2019.

\bibitem{Gunasekar2023phi}
Suriya Gunasekar, Yi~Zhang, Jyoti Aneja, Caio C'esar~Teodoro Mendes, Allison~Del Giorno, Sivakanth Gopi, Mojan Javaheripi, Piero Kauffmann, Gustavo de~Rosa, Olli Saarikivi, Adil Salim, S.~Shah, Harkirat~Singh Behl, Xin Wang, S{\'e}bastien Bubeck, Ronen Eldan, Adam~Tauman Kalai, Yin~Tat Lee, and Yuan-Fang Li.
\newblock Textbooks are all you need.
\newblock {\em ArXiv}, abs/2306.11644, 2023.

\bibitem{guo2024regiongpt}
Qiushan Guo, Shalini De~Mello, Hongxu Yin, Wonmin Byeon, Ka~Chun Cheung, Yizhou Yu, Ping Luo, and Sifei Liu.
\newblock Regiongpt: Towards region understanding vision language model.
\newblock {\em arXiv preprint arXiv:2403.02330}, 2024.

\bibitem{gupta2019lvis}
Agrim Gupta, Piotr Dollar, and Ross Girshick.
\newblock Lvis: A dataset for large vocabulary instance segmentation.
\newblock In {\em CVPR}, pages 5356--5364, 2019.

\bibitem{gurari2018vizwiz}
Danna Gurari, Qing Li, Abigale~J Stangl, Anhong Guo, Chi Lin, Kristen Grauman, Jiebo Luo, and Jeffrey~P Bigham.
\newblock Vizwiz grand challenge: Answering visual questions from blind people.
\newblock In {\em CVPR}, pages 3608--3617, 2018.

\bibitem{he2024anyref}
Junwen He, Yifan Wang, Lijun Wang, Huchuan Lu, Jun-Yan He, Jin-Peng Lan, Bin Luo, and Xuansong Xie.
\newblock Multi-modal instruction tuned llms with fine-grained visual perception.
\newblock {\em arXiv preprint arXiv:2403.02969}, 2024.

\bibitem{he2016deep}
Kaiming He, Xiangyu Zhang, Shaoqing Ren, and Jian Sun.
\newblock Deep residual learning for image recognition.
\newblock In {\em CVPR}, pages 770--778, 2016.

\bibitem{hendrycks2016gelu}
Dan Hendrycks and Kevin Gimpel.
\newblock Gaussian error linear units (gelus).
\newblock {\em arXiv preprint arXiv:1606.08415}, 2016.

\bibitem{hou2017vegfru}
Saihui Hou, Yushan Feng, and Zilei Wang.
\newblock Vegfru: A domain-specific dataset for fine-grained visual categorization.
\newblock In {\em ICCV}, pages 541--549, 2017.

\bibitem{hu2023high}
Xiaobin Hu, Shuo Wang, Xuebin Qin, Hang Dai, Wenqi Ren, Donghao Luo, Ying Tai, and Ling Shao.
\newblock High-resolution iterative feedback network for camouflaged object detection.
\newblock In {\em AAAI}, volume~37, pages 881--889, 2023.

\bibitem{huang2024dialoggen}
Minbin Huang, Yanxin Long, Xinchi Deng, Ruihang Chu, Jiangfeng Xiong, Xiaodan Liang, Hong Cheng, Qinglin Lu, and Wei Liu.
\newblock Dialoggen: Multi-modal interactive dialogue system for multi-turn text-to-image generation.
\newblock {\em arXiv preprint arXiv:2403.08857}, 2024.

\bibitem{huang2023smartedit}
Yuzhou Huang, Liangbin Xie, Xintao Wang, Ziyang Yuan, Xiaodong Cun, Yixiao Ge, Jiantao Zhou, Chao Dong, Rui Huang, Ruimao Zhang, et~al.
\newblock Smartedit: Exploring complex instruction-based image editing with multimodal large language models.
\newblock {\em arXiv preprint arXiv:2312.06739}, 2023.

\bibitem{huang2023feature}
Zhou Huang, Hang Dai, Tian-Zhu Xiang, Shuo Wang, Huai-Xin Chen, Jie Qin, and Huan Xiong.
\newblock Feature shrinkage pyramid for camouflaged object detection with transformers.
\newblock In {\em CVPR}, pages 5557--5566, 2023.

\bibitem{hudson2019gqa}
Drew~A Hudson and Christopher~D Manning.
\newblock Gqa: A new dataset for real-world visual reasoning and compositional question answering.
\newblock In {\em CVPR}, pages 6700--6709, 2019.

\bibitem{idefics2023}
{IDEFICS}.
\newblock Introducing idefics: An open reproduction of state-of-the-art visual language model.
\newblock \url{https://huggingface.co/blog/idefics}, 2023.

\bibitem{jiang2023t-rex}
Qing Jiang, Feng Li, Tianhe Ren, Shilong Liu, Zhaoyang Zeng, Kent Yu, and Lei Zhang.
\newblock T-rex: Counting by visual prompting.
\newblock {\em arXiv preprint arXiv:2311.13596}, 2023.

\bibitem{ju2023humanart}
Xuan Ju, Ailing Zeng, Jianan Wang, Qiang Xu, and Lei Zhang.
\newblock Human-art: A versatile human-centric dataset bridging natural and artificial scenes.
\newblock In {\em CVPR}, pages 618--629, 2023.

\bibitem{kafle2018dvqa}
Kushal Kafle, Brian Price, Scott Cohen, and Christopher Kanan.
\newblock Dvqa: Understanding data visualizations via question answering.
\newblock In {\em CVPR}, pages 5648--5656, 2018.

\bibitem{kamath2021mdetr}
Aishwarya Kamath, Mannat Singh, Yann LeCun, Gabriel Synnaeve, Ishan Misra, and Nicolas Carion.
\newblock Mdetr-modulated detection for end-to-end multi-modal understanding.
\newblock In {\em ICCV}, pages 1780--1790, 2021.

\bibitem{kawano2015automatic}
Yoshiyuki Kawano and Keiji Yanai.
\newblock Automatic expansion of a food image dataset leveraging existing categories with domain adaptation.
\newblock In {\em ECCVW}, pages 3--17. Springer, 2015.

\bibitem{kembhavi2016ai2d}
Aniruddha Kembhavi, Mike Salvato, Eric Kolve, Minjoon Seo, Hannaneh Hajishirzi, and Ali Farhadi.
\newblock A diagram is worth a dozen images.
\newblock In {\em ECCV}, pages 235--251, 2016.

\bibitem{khan2020animalweb}
Muhammad~Haris Khan, John McDonagh, Salman Khan, Muhammad Shahabuddin, Aditya Arora, Fahad~Shahbaz Khan, Ling Shao, and Georgios Tzimiropoulos.
\newblock Animalweb: A large-scale hierarchical dataset of annotated animal faces.
\newblock In {\em CVPR}, pages 6939--6948, 2020.

\bibitem{KhoslaYaoJayadevaprakashFeiFei_FGVC2011}
Aditya Khosla, Nityananda Jayadevaprakash, Bangpeng Yao, and Li~Fei-Fei.
\newblock Novel dataset for fine-grained image categorization.
\newblock In {\em First Workshop on Fine-Grained Visual Categorization, IEEE Conference on Computer Vision and Pattern Recognition}, Colorado Springs, CO, June 2011.

\bibitem{kim2022synthdog}
Geewook Kim, Teakgyu Hong, Moonbin Yim, JeongYeon Nam, Jinyoung Park, Jinyeong Yim, Wonseok Hwang, Sangdoo Yun, Dongyoon Han, and Seunghyun Park.
\newblock Ocr-free document understanding transformer.
\newblock In {\em ECCV}, pages 498--517. Springer, 2022.

\bibitem{kirillov2023sam}
Alexander Kirillov, Eric Mintun, Nikhila Ravi, Hanzi Mao, Chloe Rolland, Laura Gustafson, Tete Xiao, Spencer Whitehead, Alexander~C Berg, Wan-Yen Lo, et~al.
\newblock Segment anything.
\newblock In {\em ICCV}, pages 4015--4026, 2023.

\bibitem{koh2024gill}
Jing~Yu Koh, Daniel Fried, and Russ~R Salakhutdinov.
\newblock Generating images with multimodal language models.
\newblock {\em NeurIPS}, 36, 2024.

\bibitem{koh2023fromage}
Jing~Yu Koh, Ruslan Salakhutdinov, and Daniel Fried.
\newblock Grounding language models to images for multimodal inputs and outputs.
\newblock In {\em International Conference on Machine Learning}, pages 17283--17300. PMLR, 2023.

\bibitem{krause20133d}
Jonathan Krause, Michael Stark, Jia Deng, and Li~Fei-Fei.
\newblock 3d object representations for fine-grained categorization.
\newblock In {\em ICCVW}, pages 554--561, 2013.

\bibitem{krishna2017visual}
Ranjay Krishna, Yuke Zhu, Oliver Groth, Justin Johnson, Kenji Hata, Joshua Kravitz, Stephanie Chen, Yannis Kalantidis, Li-Jia Li, David~A Shamma, et~al.
\newblock Visual genome: Connecting language and vision using crowdsourced dense image annotations.
\newblock {\em IJCV}, 123:32--73, 2017.

\bibitem{kuznetsova2020openimage}
Alina Kuznetsova, Hassan Rom, Neil Alldrin, Jasper Uijlings, Ivan Krasin, Jordi Pont-Tuset, Shahab Kamali, Stefan Popov, Matteo Malloci, Alexander Kolesnikov, et~al.
\newblock The open images dataset v4: Unified image classification, object detection, and visual relationship detection at scale.
\newblock {\em IJCV}, 128(7):1956--1981, 2020.

\bibitem{labuguen2021macaquepose}
Rollyn Labuguen, Jumpei Matsumoto, Salvador~Blanco Negrete, Hiroshi Nishimaru, Hisao Nishijo, Masahiko Takada, Yasuhiro Go, Ken-ichi Inoue, and Tomohiro Shibata.
\newblock Macaquepose: a novel “in the wild” macaque monkey pose dataset for markerless motion capture.
\newblock {\em Frontiers in behavioral neuroscience}, 14:581154, 2021.

\bibitem{lai2023lisa}
Xin Lai, Zhuotao Tian, Yukang Chen, Yanwei Li, Yuhui Yuan, Shu Liu, and Jiaya Jia.
\newblock Lisa: Reasoning segmentation via large language model.
\newblock {\em arXiv preprint arXiv:2308.00692}, 2023.

\bibitem{le2019camo}
Trung-Nghia Le, Tam~V Nguyen, Zhongliang Nie, Minh-Triet Tran, and Akihiro Sugimoto.
\newblock Anabranch network for camouflaged object segmentation.
\newblock {\em Computer vision and image understanding}, 184:45--56, 2019.

\bibitem{li2023visual}
Feng Li, Qing Jiang, Hao Zhang, Tianhe Ren, Shilong Liu, Xueyan Zou, Huaizhe Xu, Hongyang Li, Chunyuan Li, Jianwei Yang, et~al.
\newblock Visual in-context prompting.
\newblock {\em arXiv preprint arXiv:2311.13601}, 2023.

\bibitem{li2022maskdino}
Feng Li, Hao Zhang, Shilong Liu, Lei Zhang, Lionel~M Ni, Heung-Yeung Shum, et~al.
\newblock Mask dino: Towards a unified transformer-based framework for object detection and segmentation.
\newblock {\em arXiv preprint arXiv:2206.02777}, 2022.

\bibitem{li2023semantic}
Feng Li, Hao Zhang, Peize Sun, Xueyan Zou, Shilong Liu, Jianwei Yang, Chunyuan Li, Lei Zhang, and Jianfeng Gao.
\newblock Semantic-sam: Segment and recognize anything at any granularity.
\newblock {\em arXiv preprint arXiv:2307.04767}, 2023.

\bibitem{li2020unicoder}
Gen Li, Nan Duan, Yuejian Fang, Ming Gong, and Daxin Jiang.
\newblock Unicoder-vl: A universal encoder for vision and language by cross-modal pre-training.
\newblock In {\em AAAI}, volume~34, pages 11336--11344, 2020.

\bibitem{li2023uniperceiverv2}
Hao Li, Jinguo Zhu, Xiaohu Jiang, Xizhou Zhu, Hongsheng Li, Chun Yuan, Xiaohua Wang, Yu~Qiao, Xiaogang Wang, Wenhai Wang, et~al.
\newblock Uni-perceiver v2: A generalist model for large-scale vision and vision-language tasks.
\newblock In {\em CVPR}, pages 2691--2700, 2023.

\bibitem{li2019crowdpose}
Jiefeng Li, Can Wang, Hao Zhu, Yihuan Mao, Hao-Shu Fang, and Cewu Lu.
\newblock Crowdpose: Efficient crowded scenes pose estimation and a new benchmark.
\newblock In {\em CVPR}, pages 10863--10872, 2019.

\bibitem{li2023blip2}
Junnan Li, Dongxu Li, Silvio Savarese, and Steven Hoi.
\newblock Blip-2: Bootstrapping language-image pre-training with frozen image encoders and large language models.
\newblock In {\em ICML}, pages 19730--19742. PMLR, 2023.

\bibitem{li2022glip}
Liunian~Harold Li, Pengchuan Zhang, Haotian Zhang, Jianwei Yang, Chunyuan Li, Yiwu Zhong, Lijuan Wang, Lu~Yuan, Lei Zhang, Jenq-Neng Hwang, et~al.
\newblock Grounded language-image pre-training.
\newblock In {\em CVPR}, pages 10965--10975, 2022.

\bibitem{li2022vitdet}
Yanghao Li, Hanzi Mao, Ross Girshick, and Kaiming He.
\newblock Exploring plain vision transformer backbones for object detection.
\newblock In {\em ECCV}, pages 280--296. Springer, 2022.

\bibitem{Li2024MiniGeminiMT}
Yanwei Li, Yuechen Zhang, Chengyao Wang, Zhisheng Zhong, Yixin Chen, Ruihang Chu, Shaoteng Liu, and Jiaya Jia.
\newblock Mini-gemini: Mining the potential of multi-modality vision language models.
\newblock {\em ArXiv}, abs/2403.18814, 2024.

\bibitem{li2023pope}
Yifan Li, Yifan Du, Kun Zhou, Jinpeng Wang, Wayne~Xin Zhao, and Ji-Rong Wen.
\newblock Evaluating object hallucination in large vision-language models.
\newblock In {\em EMNLP}, pages 292--305, 2023.

\bibitem{li2024sardet}
Yuxuan Li, Xiang Li, Weijie Li, Qibin Hou, Li~Liu, Ming-Ming Cheng, and Jian Yang.
\newblock Sardet-100k: Towards open-source benchmark and toolkit for large-scale sar object detection.
\newblock {\em arXiv preprint arXiv:2403.06534}, 2024.

\bibitem{lian2023mistralorca1}
Wing Lian, Bleys Goodson, Guan Wang, Eugene Pentland, Austin Cook, Chanvichet Vong, and "Teknium".
\newblock Mistralorca: Mistral-7b model instruct-tuned on filtered openorcav1 gpt-4 dataset.
\newblock \url{https://huggingface.co/Open-Orca/Mistral-7B-OpenOrca}, 2023.

\bibitem{lin2014coco}
Tsung-Yi Lin, Michael Maire, Serge Belongie, James Hays, Pietro Perona, Deva Ramanan, Piotr Doll{\'a}r, and C~Lawrence Zitnick.
\newblock Microsoft coco: Common objects in context.
\newblock In {\em ECCV}, pages 740--755, 2014.

\bibitem{liu2023llava1.5}
Haotian Liu, Chunyuan Li, Yuheng Li, and Yong~Jae Lee.
\newblock Improved baselines with visual instruction tuning.
\newblock {\em arXiv preprint arXiv:2310.03744}, 2023.

\bibitem{liu2024llava-next}
Haotian Liu, Chunyuan Li, Yuheng Li, Bo~Li, Yuanhan Zhang, Sheng Shen, and Yong~Jae Lee.
\newblock Llava-next: Improved reasoning, ocr, and world knowledge, 2024.

\bibitem{liu2023llava}
Haotian Liu, Chunyuan Li, Qingyang Wu, and Yong~Jae Lee.
\newblock Visual instruction tuning.
\newblock {\em NeurIPS}, 36, 2023.

\bibitem{liu2023grouppose}
Huan Liu, Qiang Chen, Zichang Tan, Jiang-Jiang Liu, Jian Wang, Xiangbo Su, Xiaolong Li, Kun Yao, Junyu Han, Errui Ding, et~al.
\newblock Group pose: A simple baseline for end-to-end multi-person pose estimation.
\newblock In {\em ICCV}, pages 15029--15038, 2023.

\bibitem{liu2019simple}
Jiang-Jiang Liu, Qibin Hou, Ming-Ming Cheng, Jiashi Feng, and Jianmin Jiang.
\newblock A simple pooling-based design for real-time salient object detection.
\newblock In {\em CVPR}, pages 3917--3926, 2019.

\bibitem{liu2021visual}
Nian Liu, Ni~Zhang, Kaiyuan Wan, Ling Shao, and Junwei Han.
\newblock Visual saliency transformer.
\newblock In {\em ICCV}, pages 4722--4732, 2021.

\bibitem{liu2023llava-plus}
Shilong Liu, Hao Cheng, Haotian Liu, Hao Zhang, Feng Li, Tianhe Ren, Xueyan Zou, Jianwei Yang, Hang Su, Jun Zhu, et~al.
\newblock Llava-plus: Learning to use tools for creating multimodal agents.
\newblock {\em arXiv preprint arXiv:2311.05437}, 2023.

\bibitem{liu2023groundingdino}
Shilong Liu, Zhaoyang Zeng, Tianhe Ren, Feng Li, Hao Zhang, Jie Yang, Chunyuan Li, Jianwei Yang, Hang Su, Jun Zhu, et~al.
\newblock Grounding dino: Marrying dino with grounded pre-training for open-set object detection.
\newblock {\em arXiv preprint arXiv:2303.05499}, 2023.

\bibitem{liu2023explicit}
Weihuang Liu, Xi~Shen, Chi-Man Pun, and Xiaodong Cun.
\newblock Explicit visual prompting for universal foreground segmentations.
\newblock {\em arXiv preprint arXiv:2305.18476}, 2023.

\bibitem{liu2023mmbench}
Yuan Liu, Haodong Duan, Yuanhan Zhang, Bo~Li, Songyang Zhang, Wangbo Zhao, Yike Yuan, Jiaqi Wang, Conghui He, Ziwei Liu, et~al.
\newblock Mmbench: Is your multi-modal model an all-around player?
\newblock {\em arXiv preprint arXiv:2307.06281}, 2023.

\bibitem{liu2021swin}
Ze~Liu, Yutong Lin, Yue Cao, Han Hu, Yixuan Wei, Zheng Zhang, Stephen Lin, and Baining Guo.
\newblock Swin transformer: Hierarchical vision transformer using shifted windows.
\newblock In {\em ICCV}, pages 10012--10022, 2021.

\bibitem{2023interngpt}
Zhaoyang Liu, Yinan He, Wenhai Wang, Weiyun Wang, Yi~Wang, Shoufa Chen, Qinglong Zhang, Zeqiang Lai, Yang Yang, Qingyun Li, Jiashuo Yu, et~al.
\newblock Interngpt: Solving vision-centric tasks by interacting with chatgpt beyond language.
\newblock {\em arXiv preprint arXiv:2305.05662}, 2023.

\bibitem{liu2023controlllm}
Zhaoyang Liu, Zeqiang Lai, Zhangwei Gao, Erfei Cui, Xizhou Zhu, Lewei Lu, Qifeng Chen, Yu~Qiao, Jifeng Dai, and Wenhai Wang.
\newblock Controlllm: Augment language models with tools by searching on graphs.
\newblock {\em arXiv preprint arXiv:2310.17796}, 2023.

\bibitem{long2022hiertext}
Shangbang Long, Siyang Qin, Dmitry Panteleev, Alessandro Bissacco, Yasuhisa Fujii, and Michalis Raptis.
\newblock Towards end-to-end unified scene text detection and layout analysis.
\newblock In {\em CVPR}, pages 1049--1059, 2022.

\bibitem{loshchilov2017adamw}
Ilya Loshchilov and Frank Hutter.
\newblock Decoupled weight decay regularization.
\newblock {\em arXiv preprint arXiv:1711.05101}, 2017.

\bibitem{lu2019vilbert}
Jiasen Lu, Dhruv Batra, Devi Parikh, and Stefan Lee.
\newblock Vilbert: Pretraining task-agnostic visiolinguistic representations for vision-and-language tasks.
\newblock {\em NeurIPS}, 32, 2019.

\bibitem{lu2022unified-io}
Jiasen Lu, Christopher Clark, Rowan Zellers, Roozbeh Mottaghi, and Aniruddha Kembhavi.
\newblock Unified-io: A unified model for vision, language, and multi-modal tasks.
\newblock In {\em The Eleventh International Conference on Learning Representations}, 2022.

\bibitem{lv2021nc4k}
Yunqiu Lv, Jing Zhang, Yuchao Dai, Aixuan Li, Bowen Liu, Nick Barnes, and Deng-Ping Fan.
\newblock Simultaneously localize, segment and rank the camouflaged objects.
\newblock In {\em CVPR}, pages 11591--11601, 2021.

\bibitem{ma2024groma}
Chuofan Ma, Yi~Jiang, Jiannan Wu, Zehuan Yuan, and Xiaojuan Qi.
\newblock Groma: Localized visual tokenization for grounding multimodal large language models.
\newblock {\em arXiv preprint arXiv:2404.13013}, 2024.

\bibitem{ma2023boosting}
Mingcan Ma, Changqun Xia, Chenxi Xie, Xiaowu Chen, and Jia Li.
\newblock Boosting broader receptive fields for salient object detection.
\newblock {\em TIP}, 32:1026--1038, 2023.

\bibitem{manas2022mapl}
Oscar Ma{\~n}as, Pau Rodriguez, Saba Ahmadi, Aida Nematzadeh, Yash Goyal, and Aishwarya Agrawal.
\newblock Mapl: Parameter-efficient adaptation of unimodal pre-trained models for vision-language few-shot prompting.
\newblock {\em arXiv preprint arXiv:2210.07179}, 2022.

\bibitem{mao2016refcocog}
Junhua Mao, Jonathan Huang, Alexander Toshev, Oana Camburu, Alan~L Yuille, and Kevin Murphy.
\newblock Generation and comprehension of unambiguous object descriptions.
\newblock In {\em CVPR}, pages 11--20, 2016.

\bibitem{marino2019okvqa}
Kenneth Marino, Mohammad Rastegari, Ali Farhadi, and Roozbeh Mottaghi.
\newblock Ok-vqa: A visual question answering benchmark requiring external knowledge.
\newblock In {\em CVPR}, pages 3195--3204, 2019.

\bibitem{masry2022chartqa}
Ahmed Masry, Xuan~Long Do, Jia~Qing Tan, Shafiq Joty, and Enamul Hoque.
\newblock Chartqa: A benchmark for question answering about charts with visual and logical reasoning.
\newblock In {\em ACL}, pages 2263--2279, 2022.

\bibitem{mathew2022infographicvqa}
Minesh Mathew, Viraj Bagal, Rub{\`e}n Tito, Dimosthenis Karatzas, Ernest Valveny, and CV~Jawahar.
\newblock Infographicvqa.
\newblock In {\em WACV}, pages 1697--1706, 2022.

\bibitem{mishra2019ocrvqa}
Anand Mishra, Shashank Shekhar, Ajeet~Kumar Singh, and Anirban Chakraborty.
\newblock Ocr-vqa: Visual question answering by reading text in images.
\newblock In {\em ICDAR}, pages 947--952. IEEE, 2019.

\bibitem{nasiriany2024pivot}
Soroush Nasiriany, Fei Xia, Wenhao Yu, Ted Xiao, Jacky Liang, Ishita Dasgupta, Annie Xie, Danny Driess, Ayzaan Wahid, Zhuo Xu, et~al.
\newblock Pivot: Iterative visual prompting elicits actionable knowledge for vlms.
\newblock {\em arXiv preprint arXiv:2402.07872}, 2024.

\bibitem{neuhold2017mapillary}
Gerhard Neuhold, Tobias Ollmann, Samuel Rota~Bulo, and Peter Kontschieder.
\newblock The mapillary vistas dataset for semantic understanding of street scenes.
\newblock In {\em ICCV}, pages 4990--4999, 2017.

\bibitem{ng2022animal}
Xun~Long Ng, Kian~Eng Ong, Qichen Zheng, Yun Ni, Si~Yong Yeo, and Jun Liu.
\newblock Animal kingdom: A large and diverse dataset for animal behavior understanding.
\newblock In {\em CVPR}, pages 19023--19034, 2022.

\bibitem{nilsback2008automated}
Maria-Elena Nilsback and Andrew Zisserman.
\newblock Automated flower classification over a large number of classes.
\newblock In {\em 2008 Sixth Indian conference on computer vision, graphics \& image processing}, pages 722--729. IEEE, 2008.

\bibitem{pan2023journeydb}
Junting Pan, Keqiang Sun, Yuying Ge, Hao Li, Haodong Duan, Xiaoshi Wu, Renrui Zhang, Aojun Zhou, Zipeng Qin, Yi~Wang, Jifeng Dai, Yu~Qiao, and Hongsheng Li.
\newblock Journeydb: A benchmark for generative image understanding, 2023.

\bibitem{pan2023kosmos}
Xichen Pan, Li~Dong, Shaohan Huang, Zhiliang Peng, Wenhu Chen, and Furu Wei.
\newblock Kosmos-g: Generating images in context with multimodal large language models.
\newblock {\em arXiv preprint arXiv:2310.02992}, 2023.

\bibitem{pang2022zoomnet}
Youwei Pang, Xiaoqi Zhao, Tian-Zhu Xiang, Lihe Zhang, and Huchuan Lu.
\newblock Zoom in and out: A mixed-scale triplet network for camouflaged object detection.
\newblock In {\em CVPR}, pages 2160--2170, 2022.

\bibitem{pang2023zoomnext}
Youwei Pang, Xiaoqi Zhao, Tian-Zhu Xiang, Lihe Zhang, and Huchuan Lu.
\newblock Zoomnext: A unified collaborative pyramid network for camouflaged object detection.
\newblock {\em arXiv preprint arXiv:2310.20208}, 2023.

\bibitem{parkhi2012cats}
Omkar~M Parkhi, Andrea Vedaldi, Andrew Zisserman, and CV~Jawahar.
\newblock Cats and dogs.
\newblock In {\em CVPR}, pages 3498--3505, 2012.

\bibitem{peng2023kosmos2}
Zhiliang Peng, Wenhui Wang, Li~Dong, Yaru Hao, Shaohan Huang, Shuming Ma, and Furu Wei.
\newblock Kosmos-2: Grounding multimodal large language models to the world.
\newblock {\em arXiv preprint arXiv:2306.14824}, 2023.

\bibitem{pereira2019fly}
Talmo~D Pereira, Diego~E Aldarondo, Lindsay Willmore, Mikhail Kislin, Samuel S-H Wang, Mala Murthy, and Joshua~W Shaevitz.
\newblock Fast animal pose estimation using deep neural networks.
\newblock {\em Nature methods}, 16(1):117--125, 2019.

\bibitem{plummer2015flickr30k}
Bryan~A Plummer, Liwei Wang, Chris~M Cervantes, Juan~C Caicedo, Julia Hockenmaier, and Svetlana Lazebnik.
\newblock Flickr30k entities: Collecting region-to-phrase correspondences for richer image-to-sentence models.
\newblock In {\em ICCV}, pages 2641--2649, 2015.

\bibitem{pramanick2023vistallm}
Shraman Pramanick, Guangxing Han, Rui Hou, Sayan Nag, Ser-Nam Lim, Nicolas Ballas, Qifan Wang, Rama Chellappa, and Amjad Almahairi.
\newblock Jack of all tasks, master of many: Designing general-purpose coarse-to-fine vision-language model.
\newblock {\em arXiv preprint arXiv:2312.12423}, 2023.

\bibitem{radford2021clip}
Alec Radford, Jong~Wook Kim, Chris Hallacy, Aditya Ramesh, Gabriel Goh, Sandhini Agarwal, Girish Sastry, Amanda Askell, Pamela Mishkin, Jack Clark, et~al.
\newblock Learning transferable visual models from natural language supervision.
\newblock In {\em ICML}, pages 8748--8763, 2021.

\bibitem{radford2018improving}
Alec Radford, Karthik Narasimhan, Tim Salimans, and Ilya Sutskever.
\newblock Improving language understanding by generative pre-training.
\newblock 2018.

\bibitem{radford2019language}
Alec Radford, Jeff Wu, Rewon Child, David Luan, Dario Amodei, and Ilya Sutskever.
\newblock Language models are unsupervised multitask learners.
\newblock 2019.

\bibitem{ramanathan2023paco}
Vignesh Ramanathan, Anmol Kalia, Vladan Petrovic, Yi~Wen, Baixue Zheng, Baishan Guo, Rui Wang, Aaron Marquez, Rama Kovvuri, Abhishek Kadian, et~al.
\newblock Paco: Parts and attributes of common objects.
\newblock In {\em CVPR}, pages 7141--7151, 2023.

\bibitem{rasheed2023glamm}
Hanoona Rasheed, Muhammad Maaz, Sahal Shaji, Abdelrahman Shaker, Salman Khan, Hisham Cholakkal, Rao~M Anwer, Erix Xing, Ming-Hsuan Yang, and Fahad~S Khan.
\newblock Glamm: Pixel grounding large multimodal model.
\newblock {\em arXiv preprint arXiv:2311.03356}, 2023.

\bibitem{rasley2020deepspeed}
Jeff Rasley, Samyam Rajbhandari, Olatunji Ruwase, and Yuxiong He.
\newblock Deepspeed: System optimizations enable training deep learning models with over 100 billion parameters.
\newblock In {\em SIGKDD}, pages 3505--3506, 2020.

\bibitem{reid2024gemini1_5}
Machel Reid, Nikolay Savinov, Denis Teplyashin, Dmitry Lepikhin, Timothy Lillicrap, Jean-baptiste Alayrac, Radu Soricut, Angeliki Lazaridou, Orhan Firat, Julian Schrittwieser, et~al.
\newblock Gemini 1.5: Unlocking multimodal understanding across millions of tokens of context.
\newblock {\em arXiv preprint arXiv:2403.05530}, 2024.

\bibitem{ren2023pixellm}
Zhongwei Ren, Zhicheng Huang, Yunchao Wei, Yao Zhao, Dongmei Fu, Jiashi Feng, and Xiaojie Jin.
\newblock Pixellm: Pixel reasoning with large multimodal model.
\newblock {\em arXiv preprint arXiv:2312.02228}, 2023.

\bibitem{rombach2022sd}
Robin Rombach, Andreas Blattmann, Dominik Lorenz, Patrick Esser, and Bj{\"o}rn Ommer.
\newblock High-resolution image synthesis with latent diffusion models.
\newblock In {\em CVPR}, pages 10684--10695, 2022.

\bibitem{sagonas2013face}
Christos Sagonas, Georgios Tzimiropoulos, Stefanos Zafeiriou, and Maja Pantic.
\newblock 300 faces in-the-wild challenge: The first facial landmark localization challenge.
\newblock In {\em ICCVW}, pages 397--403, 2013.

\bibitem{saikh2022scienceqa}
Tanik Saikh, Tirthankar Ghosal, Amish Mittal, Asif Ekbal, and Pushpak Bhattacharyya.
\newblock Scienceqa: A novel resource for question answering on scholarly articles.
\newblock {\em International Journal on Digital Libraries}, 23(3):289--301, 2022.

\bibitem{shao2019objects365}
Shuai Shao, Zeming Li, Tianyuan Zhang, Chao Peng, Gang Yu, Xiangyu Zhang, Jing Li, and Jian Sun.
\newblock Objects365: A large-scale, high-quality dataset for object detection.
\newblock In {\em ICCV}, pages 8430--8439, 2019.

\bibitem{shao2018crowdhuman}
Shuai Shao, Zijian Zhao, Boxun Li, Tete Xiao, Gang Yu, Xiangyu Zhang, and Jian Sun.
\newblock Crowdhuman: A benchmark for detecting human in a crowd.
\newblock {\em arXiv preprint arXiv:1805.00123}, 2018.

\bibitem{shtedritski2023does}
Aleksandar Shtedritski, Christian Rupprecht, and Andrea Vedaldi.
\newblock What does clip know about a red circle? visual prompt engineering for vlms.
\newblock In {\em ICCV}, pages 11987--11997, 2023.

\bibitem{shukor2023ep-alm}
Mustafa Shukor, Corentin Dancette, and Matthieu Cord.
\newblock ep-alm: Efficient perceptual augmentation of language models.
\newblock In {\em Proceedings of the IEEE/CVF International Conference on Computer Vision}, pages 22056--22069, 2023.

\bibitem{sidorov2020textcaps}
Oleksii Sidorov, Ronghang Hu, Marcus Rohrbach, and Amanpreet Singh.
\newblock Textcaps: a dataset for image captioning with reading comprehension.
\newblock In {\em ECCV}, pages 742--758, 2020.

\bibitem{singh2019textvqa}
Amanpreet Singh, Vivek Natarajan, Meet Shah, Yu~Jiang, Xinlei Chen, Dhruv Batra, Devi Parikh, and Marcus Rohrbach.
\newblock Towards vqa models that can read.
\newblock In {\em CVPR}, pages 8317--8326, 2019.

\bibitem{Singh2021TextOCRTL}
Amanpreet Singh, Guan Pang, Mandy Toh, Jing Huang, Wojciech Galuba, and Tal Hassner.
\newblock Textocr: Towards large-scale end-to-end reasoning for arbitrary-shaped scene text.
\newblock In {\em CVPR}, pages 8802--8812, 2021.

\bibitem{su2019vlbert}
Weijie Su, Xizhou Zhu, Yue Cao, Bin Li, Lewei Lu, Furu Wei, and Jifeng Dai.
\newblock Vl-bert: Pre-training of generic visual-linguistic representations.
\newblock {\em arXiv preprint arXiv:1908.08530}, 2019.

\bibitem{sun2023emu2}
Quan Sun, Yufeng Cui, Xiaosong Zhang, Fan Zhang, Qiying Yu, Zhengxiong Luo, Yueze Wang, Yongming Rao, Jingjing Liu, Tiejun Huang, et~al.
\newblock Generative multimodal models are in-context learners.
\newblock {\em arXiv preprint arXiv:2312.13286}, 2023.

\bibitem{sun2023emu}
Quan Sun, Qiying Yu, Yufeng Cui, Fan Zhang, Xiaosong Zhang, Yueze Wang, Hongcheng Gao, Jingjing Liu, Tiejun Huang, and Xinlong Wang.
\newblock Generative pretraining in multimodality.
\newblock In {\em ICLR}, 2024.

\bibitem{Sun2019ICDAR2C}
Yipeng Sun, Zihan Ni, Chee-Kheng Chng, Yuliang Liu, Canjie Luo, Chun~Chet Ng, Junyu Han, Errui Ding, Jingtuo Liu, Dimosthenis Karatzas, et~al.
\newblock Icdar 2019 competition on large-scale street view text with partial labeling-rrc-lsvt.
\newblock In {\em ICDAR}, pages 1557--1562. IEEE, 2019.

\bibitem{suris2023vipergpt}
D{\'\i}dac Sur{\'\i}s, Sachit Menon, and Carl Vondrick.
\newblock Vipergpt: Visual inference via python execution for reasoning.
\newblock In {\em ICCV}, pages 11888--11898, 2023.

\bibitem{tang2019deeppcb}
Sanli Tang, Fan He, Xiaolin Huang, and Jie Yang.
\newblock Online pcb defect detector on a new pcb defect dataset.
\newblock {\em arXiv preprint arXiv:1902.06197}, 2019.

\bibitem{team2024chameleon}
Chameleon Team.
\newblock Chameleon: Mixed-modal early-fusion foundation models.
\newblock {\em arXiv preprint arXiv:2405.09818}, 2024.

\bibitem{team2023gemini}
Gemini Team, Rohan Anil, Sebastian Borgeaud, Yonghui Wu, Jean-Baptiste Alayrac, Jiahui Yu, Radu Soricut, Johan Schalkwyk, Andrew~M Dai, Anja Hauth, et~al.
\newblock Gemini: a family of highly capable multimodal models.
\newblock {\em arXiv preprint arXiv:2312.11805}, 2023.

\bibitem{2023internlm}
InternLM Team.
\newblock Internlm: A multilingual language model with progressively enhanced capabilities.
\newblock \url{https://github.com/InternLM/InternLM}, 2023.

\bibitem{touvron2023llama}
Hugo Touvron, Thibaut Lavril, Gautier Izacard, Xavier Martinet, Marie-Anne Lachaux, Timoth{\'e}e Lacroix, Baptiste Rozi{\`e}re, Naman Goyal, Eric Hambro, Faisal Azhar, Aurelien Rodriguez, Armand Joulin, Edouard Grave, and Guillaume Lample.
\newblock Llama: Open and efficient foundation language models.
\newblock {\em arXiv preprint arXiv:2302.13971}, 2023.

\bibitem{touvron2023llama2}
Hugo Touvron, Louis Martin, Kevin Stone, Peter Albert, Amjad Almahairi, Yasmine Babaei, Nikolay Bashlykov, Soumya Batra, Prajjwal Bhargava, Shruti Bhosale, et~al.
\newblock Llama 2: Open foundation and fine-tuned chat models.
\newblock {\em arXiv preprint arXiv:2307.09288}, 2023.

\bibitem{tsao2023autovp}
Hsi-Ai Tsao, Lei Hsiung, Pin-Yu Chen, Sijia Liu, and Tsung-Yi Ho.
\newblock Autovp: An automated visual prompting framework and benchmark.
\newblock {\em arXiv preprint arXiv:2310.08381}, 2023.

\bibitem{van2021benchmarking}
Grant Van~Horn, Elijah Cole, Sara Beery, Kimberly Wilber, Serge Belongie, and Oisin Mac~Aodha.
\newblock Benchmarking representation learning for natural world image collections.
\newblock In {\em CVPR}, pages 12884--12893, 2021.

\bibitem{WahCUB_200_2011}
C.~Wah, S.~Branson, P.~Welinder, P.~Perona, and S.~Belongie.
\newblock Technical Report CNS-TR-2011-001, California Institute of Technology, 2011.

\bibitem{git}
Haiyang Wang, Hao Tang, Li~Jiang, Shaoshuai Shi, Muhammad~Ferjad Naeem, Hongsheng Li, Bernt Schiele, and Liwei Wang.
\newblock Git:towards generalist vision transformer through universal language interface.
\newblock {\em arXiv preprint arXiv:2403.09394}, 2024.

\bibitem{wang2023v3det}
Jiaqi Wang, Pan Zhang, Tao Chu, Yuhang Cao, Yujie Zhou, Tong Wu, Bin Wang, Conghui He, and Dahua Lin.
\newblock V3det: Vast vocabulary visual detection dataset.
\newblock In {\em ICCV}, pages 19844--19854, 2023.

\bibitem{wang2021loveda}
Junjue Wang, Zhuo Zheng, Ailong Ma, Xiaoyan Lu, and Yanfei Zhong.
\newblock Loveda: A remote sensing land-cover dataset for domain adaptive semantic segmentation.
\newblock {\em arXiv preprint arXiv:2110.08733}, 2021.

\bibitem{wang2017duts}
Lijun Wang, Huchuan Lu, Yifan Wang, Mengyang Feng, Dong Wang, Baocai Yin, and Xiang Ruan.
\newblock Learning to detect salient objects with image-level supervision.
\newblock In {\em CVPR}, pages 136--145, 2017.

\bibitem{wang2024allseeingv2}
Weiyun Wang, Yiming Ren, Haowen Luo, Tiantong Li, Chenxiang Yan, Zhe Chen, Wenhai Wang, Qingyun Li, Lewei Lu, Xizhou Zhu, et~al.
\newblock The all-seeing project v2: Towards general relation comprehension of the open world.
\newblock {\em arXiv preprint arXiv:2402.19474}, 2024.

\bibitem{wang2023allseeing}
Weiyun Wang, Min Shi, Qingyun Li, Wenhai Wang, Zhenhang Huang, Linjie Xing, Zhe Chen, Hao Li, Xizhou Zhu, Zhiguo Cao, et~al.
\newblock The all-seeing project: Towards panoptic visual recognition and understanding of the open world.
\newblock In {\em ICLR}, 2024.

\bibitem{wang2023visionllm}
Wenhai Wang, Zhe Chen, Xiaokang Chen, Jiannan Wu, Xizhou Zhu, Gang Zeng, Ping Luo, Tong Lu, Jie Zhou, Yu~Qiao, et~al.
\newblock Visionllm: Large language model is also an open-ended decoder for vision-centric tasks.
\newblock {\em NeurIPS}, 36, 2023.

\bibitem{wang2023images}
Xinlong Wang, Wen Wang, Yue Cao, Chunhua Shen, and Tiejun Huang.
\newblock Images speak in images: A generalist painter for in-context visual learning.
\newblock In {\em CVPR}, pages 6830--6839, 2023.

\bibitem{wang2023seggpt}
Xinlong Wang, Xiaosong Zhang, Yue Cao, Wen Wang, Chunhua Shen, and Tiejun Huang.
\newblock Seggpt: Segmenting everything in context.
\newblock {\em arXiv preprint arXiv:2304.03284}, 2023.

\bibitem{wang2023hulk}
Yizhou Wang, Yixuan Wu, Shixiang Tang, Weizhen He, Xun Guo, Feng Zhu, Lei Bai, Rui Zhao, Jian Wu, Tong He, et~al.
\newblock Hulk: A universal knowledge translator for human-centric tasks.
\newblock {\em arXiv preprint arXiv:2312.01697}, 2023.

\bibitem{wei2020label}
Jun Wei, Shuhui Wang, Zhe Wu, Chi Su, Qingming Huang, and Qi~Tian.
\newblock Label decoupling framework for salient object detection.
\newblock In {\em CVPR}, pages 13025--13034, 2020.

\bibitem{wu2023visual-chatgpt}
Chenfei Wu, Shengming Yin, Weizhen Qi, Xiaodong Wang, Zecheng Tang, and Nan Duan.
\newblock Visual chatgpt: Talking, drawing and editing with visual foundation models.
\newblock {\em arXiv preprint arXiv:2303.04671}, 2023.

\bibitem{wu2022grit}
Jialian Wu, Jianfeng Wang, Zhengyuan Yang, Zhe Gan, Zicheng Liu, Junsong Yuan, and Lijuan Wang.
\newblock Grit: A generative region-to-text transformer for object understanding.
\newblock {\em arXiv preprint arXiv:2212.00280}, 2022.

\bibitem{wu2023omni}
Jialin Wu, Xia Hu, Yaqing Wang, Bo~Pang, and Radu Soricut.
\newblock Omni-smola: Boosting generalist multimodal models with soft mixture of low-rank experts.
\newblock {\em arXiv preprint arXiv:2312.00968}, 2023.

\bibitem{wu2023general}
Junfeng Wu, Yi~Jiang, Qihao Liu, Zehuan Yuan, Xiang Bai, and Song Bai.
\newblock General object foundation model for images and videos at scale.
\newblock {\em arXiv preprint arXiv:2312.09158}, 2023.

\bibitem{xia2023llmga}
Bin Xia, Shiyin Wang, Yingfan Tao, Yitong Wang, and Jiaya Jia.
\newblock Llmga: Multimodal large language model based generation assistant.
\newblock {\em arXiv preprint arXiv:2311.16500}, 2023.

\bibitem{xia2018dota}
Gui-Song Xia, Xiang Bai, Jian Ding, Zhen Zhu, Serge Belongie, Jiebo Luo, Mihai Datcu, Marcello Pelillo, and Liangpei Zhang.
\newblock Dota: A large-scale dataset for object detection in aerial images.
\newblock In {\em CVPR}, pages 3974--3983, 2018.

\bibitem{xu2023pixelllm}
Jiarui Xu, Xingyi Zhou, Shen Yan, Xiuye Gu, Anurag Arnab, Chen Sun, Xiaolong Wang, and Cordelia Schmid.
\newblock Pixel aligned language models.
\newblock {\em arXiv preprint arXiv:2312.09237}, 2023.

\bibitem{Xu2022ViTPose++}
Yufei Xu, Jing Zhang, Qiming Zhang, and Dacheng Tao.
\newblock Vitpose++: Vision transformer for generic body pose estimation.
\newblock {\em TPAMI}, 46:1212--1230, 2022.

\bibitem{yan2023uninext}
Bin Yan, Yi~Jiang, Jiannan Wu, Dong Wang, Ping Luo, Zehuan Yuan, and Huchuan Lu.
\newblock Universal instance perception as object discovery and retrieval.
\newblock In {\em CVPR}, pages 15325--15336, 2023.

\bibitem{yang2023set}
Jianwei Yang, Hao Zhang, Feng Li, Xueyan Zou, Chunyuan Li, and Jianfeng Gao.
\newblock Set-of-mark prompting unleashes extraordinary visual grounding in gpt-4v.
\newblock {\em arXiv preprint arXiv:2310.11441}, 2023.

\bibitem{Yang2023edpose}
Jie Yang, Ailing Zeng, Siyi Liu, Feng Li, Ruimao Zhang, and Lei Zhang.
\newblock Explicit box detection unifies end-to-end multi-person pose estimation.
\newblock {\em ArXiv}, abs/2302.01593, 2023.

\bibitem{yang2023unipose}
Jie Yang, Ailing Zeng, Ruimao Zhang, and Lei Zhang.
\newblock Unipose: Detecting any keypoints.
\newblock {\em arXiv preprint arXiv:2310.08530}, 2023.

\bibitem{yang2022apt36k}
Yuxiang Yang, Junjie Yang, Yufei Xu, Jing Zhang, Long Lan, and Dacheng Tao.
\newblock Apt-36k: A large-scale benchmark for animal pose estimation and tracking.
\newblock {\em NeurIPS}, 35:17301--17313, 2022.

\bibitem{ye2023samed2d20m}
Jin Ye, Junlong Cheng, Jianpin Chen, Zhongying Deng, Tianbin Li, Haoyu Wang, Yanzhou Su, Ziyan Huang, Jilong Chen, Lei Jiang, et~al.
\newblock Sa-med2d-20m dataset: Segment anything in 2d medical imaging with 20 million masks.
\newblock {\em arXiv preprint arXiv:2311.11969}, 2023.

\bibitem{you2023ferret}
Haoxuan You, Haotian Zhang, Zhe Gan, Xianzhi Du, Bowen Zhang, Zirui Wang, Liangliang Cao, Shih-Fu Chang, and Yinfei Yang.
\newblock Ferret: Refer and ground anything anywhere at any granularity.
\newblock {\em arXiv preprint arXiv:2310.07704}, 2023.

\bibitem{yu2021ernie}
Fei Yu, Jiji Tang, Weichong Yin, Yu~Sun, Hao Tian, Hua Wu, and Haifeng Wang.
\newblock Ernie-vil: Knowledge enhanced vision-language representations through scene graphs.
\newblock In {\em AAAI}, volume~35, pages 3208--3216, 2021.

\bibitem{yu2021ap10k}
Hang Yu, Yufei Xu, Jing Zhang, Wei Zhao, Ziyu Guan, and Dacheng Tao.
\newblock Ap-10k: A benchmark for animal pose estimation in the wild.
\newblock {\em arXiv preprint arXiv:2108.12617}, 2021.

\bibitem{yu2016refcoco}
Licheng Yu, Patrick Poirson, Shan Yang, Alexander~C Berg, and Tamara~L Berg.
\newblock Modeling context in referring expressions.
\newblock In {\em ECCV}, pages 69--85, 2016.

\bibitem{yu2023cm3leon}
Lili Yu, Bowen Shi, Ramakanth Pasunuru, Benjamin Muller, Olga Golovneva, Tianlu Wang, Arun Babu, Binh Tang, Brian Karrer, Shelly Sheynin, et~al.
\newblock Scaling autoregressive multi-modal models: Pretraining and instruction tuning.
\newblock {\em arXiv preprint arXiv:2309.02591}, 2(3), 2023.

\bibitem{yu2020gradient}
Tianhe Yu, Saurabh Kumar, Abhishek Gupta, Sergey Levine, Karol Hausman, and Chelsea Finn.
\newblock Gradient surgery for multi-task learning.
\newblock {\em Advances in Neural Information Processing Systems}, 33:5824--5836, 2020.

\bibitem{yuan2023osprey}
Yuqian Yuan, Wentong Li, Jian Liu, Dongqi Tang, Xinjie Luo, Chi Qin, Lei Zhang, and Jianke Zhu.
\newblock Osprey: Pixel understanding with visual instruction tuning.
\newblock {\em arXiv preprint arXiv:2312.10032}, 2023.

\bibitem{yun2022selfreformer}
Yi~Ke Yun and Weisi Lin.
\newblock Selfreformer: Self-refined network with transformer for salient object detection.
\newblock {\em arXiv preprint arXiv:2205.11283}, 2022.

\bibitem{zellers2019vcr}
Rowan Zellers, Yonatan Bisk, Ali Farhadi, and Yejin Choi.
\newblock From recognition to cognition: Visual commonsense reasoning.
\newblock In {\em CVPR}, pages 6720--6731, 2019.

\bibitem{zhan2024anygpt}
Jun Zhan, Junqi Dai, Jiasheng Ye, Yunhua Zhou, Dong Zhang, Zhigeng Liu, Xin Zhang, Ruibin Yuan, Ge~Zhang, Linyang Li, et~al.
\newblock Anygpt: Unified multimodal llm with discrete sequence modeling.
\newblock {\em arXiv preprint arXiv:2402.12226}, 2024.

\bibitem{zhang2023openseed}
Hao Zhang, Feng Li, Xueyan Zou, Shilong Liu, Chunyuan Li, Jianwei Yang, and Lei Zhang.
\newblock A simple framework for open-vocabulary segmentation and detection.
\newblock In {\em ICCV}, pages 1020--1031, 2023.

\bibitem{zhang2024ferretv2}
Haotian Zhang, Haoxuan You, Philipp Dufter, Bowen Zhang, Chen Chen, Hong-You Chen, Tsu-Jui Fu, William~Yang Wang, Shih-Fu Chang, Zhe Gan, et~al.
\newblock Ferret-v2: An improved baseline for referring and grounding with large language models.
\newblock {\em arXiv preprint arXiv:2404.07973}, 2024.

\bibitem{zhang2023gpt4roi}
Shilong Zhang, Peize Sun, Shoufa Chen, Min Xiao, Wenqi Shao, Wenwei Zhang, Kai Chen, and Ping Luo.
\newblock Gpt4roi: Instruction tuning large language model on region-of-interest.
\newblock {\em arXiv preprint arXiv:2307.03601}, 2023.

\bibitem{zhang2023ddq}
Shilong Zhang, Xinjiang Wang, Jiaqi Wang, Jiangmiao Pang, Chengqi Lyu, Wenwei Zhang, Ping Luo, and Kai Chen.
\newblock Dense distinct query for end-to-end object detection.
\newblock In {\em CVPR}, pages 7329--7338, 2023.

\bibitem{zhang2022opt}
Susan Zhang, Stephen Roller, Naman Goyal, Mikel Artetxe, Moya Chen, Shuohui Chen, Christopher Dewan, Mona Diab, Xian Li, Xi~Victoria Lin, et~al.
\newblock Opt: Open pre-trained transformer language models.
\newblock {\em arXiv preprint arXiv:2205.01068}, 2022.

\bibitem{zhang2023llavar}
Yanzhe Zhang, Ruiyi Zhang, Jiuxiang Gu, Yufan Zhou, Nedim Lipka, Diyi Yang, and Tong Sun.
\newblock Llavar: Enhanced visual instruction tuning for text-rich image understanding.
\newblock {\em arXiv preprint arXiv:2306.17107}, 2023.

\bibitem{zhang2024groundhog}
Yichi Zhang, Ziqiao Ma, Xiaofeng Gao, Suhaila Shakiah, Qiaozi Gao, and Joyce Chai.
\newblock Groundhog: Grounding large language models to holistic segmentation.
\newblock {\em arXiv preprint arXiv:2402.16846}, 2024.

\bibitem{zhang2024psalm}
Zheng Zhang, Yeyao Ma, Enming Zhang, and Xiang Bai.
\newblock Psalm: Pixelwise segmentation with large multi-modal model.
\newblock {\em arXiv preprint arXiv:2403.14598}, 2024.

\bibitem{zheng2023judging}
Lianmin Zheng, Wei-Lin Chiang, Ying Sheng, Siyuan Zhuang, Zhanghao Wu, Yonghao Zhuang, Zi~Lin, Zhuohan Li, Dacheng Li, Eric Xing, et~al.
\newblock Judging llm-as-a-judge with mt-bench and chatbot arena.
\newblock {\em NeurIPS}, 36, 2024.

\bibitem{zheng2018cpd1k}
Yunfei Zheng, Xiongwei Zhang, Feng Wang, Tieyong Cao, Meng Sun, and Xiaobing Wang.
\newblock Detection of people with camouflage pattern via dense deconvolution network.
\newblock {\em IEEE Signal Processing Letters}, 26(1):29--33, 2018.

\bibitem{zhong2022regionclip}
Yiwu Zhong, Jianwei Yang, Pengchuan Zhang, Chunyuan Li, Noel Codella, Liunian~Harold Li, Luowei Zhou, Xiyang Dai, Lu~Yuan, Yin Li, et~al.
\newblock Regionclip: Region-based language-image pretraining.
\newblock In {\em CVPR}, pages 16793--16803, 2022.

\bibitem{zhou2017ade20k}
Bolei Zhou, Hang Zhao, Xavier Puig, Sanja Fidler, Adela Barriuso, and Antonio Torralba.
\newblock Scene parsing through ade20k dataset.
\newblock In {\em CVPR}, pages 633--641, 2017.

\bibitem{zhu2023minigpt4}
Deyao Zhu, Jun Chen, Xiaoqian Shen, Xiang Li, and Mohamed Elhoseiny.
\newblock Minigpt-4: Enhancing vision-language understanding with advanced large language models.
\newblock In {\em ICLR}, 2024.

\bibitem{zhu2022uni}
Jinguo Zhu, Xizhou Zhu, Wenhai Wang, Xiaohua Wang, Hongsheng Li, Xiaogang Wang, and Jifeng Dai.
\newblock Uni-perceiver-moe: Learning sparse generalist models with conditional moes.
\newblock {\em arXiv preprint arXiv:2206.04674}, 2022.

\bibitem{zhu2022uniperceiver-moe}
Jinguo Zhu, Xizhou Zhu, Wenhai Wang, Xiaohua Wang, Hongsheng Li, Xiaogang Wang, and Jifeng Dai.
\newblock Uni-perceiver-moe: Learning sparse generalist models with conditional moes.
\newblock {\em Advances in Neural Information Processing Systems}, 35:2664--2678, 2022.

\bibitem{zhu2020deformable}
Xizhou Zhu, Weijie Su, Lewei Lu, Bin Li, Xiaogang Wang, and Jifeng Dai.
\newblock Deformable detr: Deformable transformers for end-to-end object detection.
\newblock In {\em ICLR}, 2020.

\bibitem{zhu2022uniperceiver}
Xizhou Zhu, Jinguo Zhu, Hao Li, Xiaoshi Wu, Hongsheng Li, Xiaohua Wang, and Jifeng Dai.
\newblock Uni-perceiver: Pre-training unified architecture for generic perception for zero-shot and few-shot tasks.
\newblock In {\em CVPR}, pages 16804--16815, 2022.

\bibitem{Zhu2024llavaPhi}
Yichen Zhu, Minjie Zhu, Ning Liu, Zhicai Ou, Xiaofeng Mou, and Jian Tang.
\newblock Llava-phi: Efficient multi-modal assistant with small language model.
\newblock {\em ArXiv}, abs/2401.02330, 2024.

\bibitem{zou2023xdecoder}
Xueyan Zou, Zi-Yi Dou, Jianwei Yang, Zhe Gan, Linjie Li, Chunyuan Li, Xiyang Dai, Harkirat Behl, Jianfeng Wang, Lu~Yuan, et~al.
\newblock Generalized decoding for pixel, image, and language.
\newblock In {\em CVPR}, pages 15116--15127, 2023.

\bibitem{zou2024segment}
Xueyan Zou, Jianwei Yang, Hao Zhang, Feng Li, Linjie Li, Jianfeng Wang, Lijuan Wang, Jianfeng Gao, and Yong~Jae Lee.
\newblock Segment everything everywhere all at once.
\newblock {\em NeurIPS}, 36, 2024.

\end{thebibliography}
}


\clearpage
\appendix

\setcounter{table}{0}   
\setcounter{figure}{0}
\renewcommand{\thetable}{A\arabic{table}}
\renewcommand{\thefigure}{A\arabic{figure}}

\begin{center}
    {\huge{The Appendix of VisionLLM v2}}
\end{center}

\section{More Results}
\label{sec:more_exps}

\subsection{More Experimental Results}

\noindent
\textbf{Region Captioning.} 
To access the region understanding capabilities of \modelname, we evaluate our models on two prominent benchmarks: RefCOCOg~\cite{mao2016refcocog} and Visual Genome~\cite{krishna2017visual}. The results are presented in Table~\ref{subtab:region_captioning}. Notably, \modelname-Chat significantly outperforms the state-of-the-art methods, with the improvements of +8.6 and +8.4 points in CIDEr scores on RefCOCOg~\cite{mao2016refcocog} and Visual Genome (validation subset)~\cite{krishna2017visual, rasheed2023glamm}, respectively. The generalist \modelname also shows promising performance on RefCOCOg. These results demonstrate the strong fine-grained region captioning capabilities of our model.

\begin{table*}[b]
\scriptsize
\centering

\setlength\tabcolsep{1.1mm}
\renewcommand{\arraystretch}{1.2}

\begin{subtable}{0.3\linewidth}
    \begin{tabular}{l|cc}
        method & Flickr30K & NoCaps \\
        \hline
        Flamingo-80B~\cite{alayrac2022flamingo} & 67.2 & - \\
        Kosmos-2~\cite{peng2023kosmos2}         & 66.7 & - \\
        BLIP-2~\cite{li2023blip2}               & 71.6 & 103.9 \\
        InstructBLIP~\cite{instructblip}        & 82.8 & 121.9 \\
        Shikra-13B~\cite{chen2023shikra}        & 73.9 & - \\
        ASM~\cite{wang2023allseeing}            & 87.7 & 117.2 \\
        InternVL-G~\cite{chen2023internvl}      & 79.2 & 113.7 \\
        Qwen-VL~\cite{bai2023qwenvl}            & 85.8 & 121.4 \\
        Qwen-VL-Chat~\cite{bai2023qwenvl}       & 81.0 & 120.2 \\

        \hline
        \rowcolor{gray!15}
        \modelname-Chat                     & 88.7 & 118.1 \\
        \rowcolor{gray!15}
        \modelname                  & 90.0 & 116.2 \\
        
    \end{tabular}
    \caption{Zero-shot image captioning.}
    \label{subtab:image_captioning}
\end{subtable}%
\hspace{2.5em}
\begin{subtable}{0.60\linewidth}
    \begin{tabular}{l|cc|cc|cc}
           & \multicolumn{2}{c|}{RefCOCOg} & \multicolumn{2}{c|}{VG (full set)} & \multicolumn{2}{c}{VG (subset)} \\
        \multirow{-2}{*}{method} & METEOR & CIDEr & METEOR & CIDEr & METEOR & CIDEr  \\

    \hline
    GRiT~\cite{wu2022grit} & 15.2 & 71.6 & 17.1 & 142.0 & - & - \\
    Kosmos-2~\cite{peng2023kosmos2} & 14.1 & 62.3 & - & - & - & - \\
    GPT4RoI~\cite{zhang2023gpt4roi} & - & - & 17.4 & 145.2 & - & - \\
    ASM~\cite{wang2023allseeing} & 20.8 & 103.0 & 18.0 & 145.1 & - & -\\
    RegionGPT~\cite{guo2024regiongpt} & 16.9 & 109.9 & 17.0 & 145.6 & - & - \\
    PixelLLM~\cite{xu2023pixelllm} & 14.3 & 82.3 & 19.9 & 148.9 & - & - \\
    GLaMM~\cite{rasheed2023glamm} & 16.1 & 107.3 & - & - & 19.0 & 163.9 \\
    Groma~\cite{ma2024groma} & 16.8 & 107.3 & - & - & 19.0 & 158.4 \\

    \hline
    \rowcolor{gray!15}
    \modelname-Chat & 21.2 & 118.5 & 17.8 & 149.2 & 20.0 & 172.3 \\
    \rowcolor{gray!15}
    \modelname & 21.1 & 116.6 & 17.5 & 146.7 & 19.8 & 170.1 \\

    \end{tabular}
    \caption{Region captioning.}
    \label{subtab:region_captioning}
\end{subtable}%

\caption{\textbf{Comparison of zero-shot image captioning and region captioning performance.} Zero-shot image captioning is evaluated on Flickr30K test set~\cite{plummer2015flickr30k} and NoCaps validation set~\cite{agrawal2019nocaps}, using CIDEr as evaluation metric. For region captioning on Visual Genome~\cite{krishna2017visual}, full set refers to the use of all validation samples for evaluation, while subset denotes the 5000 samples specified by~\cite{rasheed2023glamm}.
}
\label{tab:captioning}
\end{table*}

\noindent
\textbf{Visual Grounding.} Visual grounding is a crucial vision task that associates the language description with the specific object within an image. Using the box or mask as the output format, visual grounding can be further categorized into referring expression comprehension (REC) and referring expression segmentation (RES) tasks. We comprehensively list the comparison results for the two tasks in Table~\ref{tab:rec} and Table~\ref{tab:res}, respectively. From Table~\ref{tab:rec}, it is found that \modelname achieves the best performance on RefCOCO~\cite{yu2016refcoco} among MLLMs. We also showcase that \modelname exhibits remarkable pixel-level segmentation capacities in Table~\ref{tab:res}. Without further fine-tuning, our model demonstrates the good gIoU result of 51.0 on the challenging ReasonSeg dataset~\cite{lai2023lisa}.

\begin{table*}[t]
\scriptsize
\centering
\renewcommand{\arraystretch}{1.2}
\setlength\tabcolsep{3.2mm}

\begin{tabular}{l|c|ccc|ccc|cc}
      & & \multicolumn{3}{c|}{RefCOCO} & \multicolumn{3}{c|}{RefCOCO+} & \multicolumn{2}{c}{RefCOCOg}  \\
    \multirow{-2}{*}{method} & \multirow{-2}{*}{type} & val & testA & testB  & val & testA & testB & val & test \\
    \hline
    
    UNITER~\cite{chen2020uniter} & & 81.4 & 87.0 & 74.2 & 75.9 & 81.5 & 66.7 & 74.0 & 68.7 \\
    VILLA~\cite{gan2020villa} &   &  82.4 & 87.5 & 74.8 & 76.2 & 81.5 & 66.8 & 76.2 & 76.7 \\
    MDETR~\cite{kamath2021mdetr}  &    &       86.8 & 89.6 & 81.4 & 79.5 & 84.1 & 70.6 & 81.6 & 80.9 \\
    Grounding DINO T$^*$~\cite{liu2023groundingdino} & & 89.2 & 91.9 & 86.0 & 81.1 & 87.4 & 74.7 & 85.2 & 84.9 \\
    Grounding DINO L$^*$~\cite{liu2023groundingdino} & \multirow{-5}{*}{VGM} & 90.6 & 93.2 &  88.2 & 82.8 & 89.0 & 75.9 & 86.1 & 87.0 \\
    \hline

    Shikra-7B~\cite{chen2023shikra} &  & 87.0 & 90.6 & 80.2 & 81.6 & 87.4 & 72.1 & 82.3 & 82.2 \\
    Shikra~13B~\cite{chen2023shikra} & & 87.8 & 91.1 & 81.8 & 82.9 & 87.8 & 74.4 & 82.6 & 83.2 \\
    MiniGPT-v2-7B~\cite{chen2023minigpt-v2} &   & 88.1 & 91.3 & 84.3 & 79.6 & 85.5 & 73.3 & 84.2 & 84.3 \\
    Qwen-VL-7B~\cite{bai2023qwenvl} & &  88.6 & 92.3 & 84.5 & 82.8 & 88.6 & 76.8 & 86.0 & 86.3 \\
    VistaLLM~\cite{pramanick2023vistallm} & &         88.1 & 91.5 & 83.0 & 82.9 & 89.8 & 74.8 & 83.6 & 84.4 \\
    Ferret-7B~\cite{you2023ferret} &  &  87.5 & 91.4 & 82.5 & 80.8 & 87.4 & 73.1 & 83.9 & 84.8 \\

    \rowcolor{gray!15}
    \modelname & \multirow{-7}{*}{MLLM} & 87.9 & 91.2 & 84.3 & 77.6 & 83.8 & 70.2 & 82.9 & 84.1 \\

\end{tabular}

\caption{\textbf{Comparison of referring expression comprehension performance.} 
The results are reported based on P@0.5.
VGM and MLLM represent vision generalist model and multimodal large language model, respectively. $^*$The model is finetuned on the dataset.
}
\label{tab:rec}
\end{table*}

\begin{table*}[t]
\scriptsize
\centering
\renewcommand{\arraystretch}{1.2}
\setlength\tabcolsep{2.5mm}

\begin{tabular}{l|c|ccc|ccc|cc|cc}
      & & \multicolumn{3}{c|}{RefCOCO} & \multicolumn{3}{c|}{RefCOCO+} & \multicolumn{2}{c|}{RefCOCOg} & ReasonSeg  \\
    \multirow{-2}{*}{method} & \multirow{-2}{*}{type} & val & testA & testB  & val & testA & testB & val & test & gIoU  \\
    \hline

    X-Decoder (L)~\cite{zou2023xdecoder} & & - & - & - & - & - & - & 64.6 & - & -  \\
    SEEM (L)~\cite{zou2024segment}          & & - & - & - & - & - & - & 65.6 & - & -  \\
    UNINEXT (R50)~\cite{yan2023uninext} & & 77.9 & 79.7 & 75.8 & 66.2 & 71.2 & 59.0 & 70.0 & 70.5 & - \\
    UNINEXT (H)~\cite{yan2023uninext} & & 82.2 & 83.4 & 81.3 & 72.5 & 76.4 & 66.2 & 74.7 & 76.4 & -  \\
    GLEE-Pro~\cite{wu2023general}  & \multirow{-5}{*}{VGM}  & 80.0 & - & - & 69.6 & - & - & 72.9 & - & -  \\
    \hline

    LISA-7B~\cite{lai2023lisa} & & 74.1 & 76.5 & 71.1 & 62.4 & 67.4 & 56.5 & 66.4 & 68.5 & 44.4  \\
    LISA-7B$^*$~\cite{lai2023lisa} & & 74.9 & 79.1 & 72.3 & 65.1 & 70.8 & 58.1 & 67.9 & 70.6 & 52.9  \\
    PixelLM~\cite{ren2023pixellm} & & 73.0 & 76.5 & 68.2 & 66.3 & 71.7 & 58.3 & 69.3 & 70.5 & -  \\
    PixelLLM~\cite{xu2023pixelllm} & & 76.9 & 78.5 & 74.4 & 69.2 & 72.1 & 64.5 & 70.7 & 72.4 & -  \\
    AnyRef$^*$~\cite{he2024anyref} & & 76.9 & 79.9 & 74.2 & 70.3 & 73.5 & 61.8 & 70.0 & 70.7 & -  \\
    GROUNDINGHOG~\cite{zhang2024groundhog} & & 78.5 & 79.9 & 75.7 & 70.5 & 75.0 & 64.9 & 74.1 & 74.6 & 56.2  \\
    GLaMM~\cite{rasheed2023glamm} &  & 79.5 & 83.2 & 76.9 & 72.6 & 78.7 & 64.6 & 74.2 & 74.9 & -  \\
    
    \rowcolor{gray!15} 
    \modelname & \multirow{-8}{*}{MLLM} & 76.6 & 79.3 & 74.3 & 64.5 & 69.8 & 61.5 & 70.7 & 71.2 & 51.0  \\

\end{tabular}

\caption{\textbf{Comparison of referring expression segmentation performance.} 
gIoU denotes the general IoU. The results for RefCOCO/+/g~\cite{yu2016refcoco, mao2016refcocog} are reported based on cumulative IoU (cIoU).
VGM and MLLM represent vision generalist models and multimodal large language models, respectively. $^*$The model is finetuned on the dataset.
}
\label{tab:res}
\vspace{-2mm}
\end{table*}

\begin{table}[b]
    \centering
    \begin{minipage}[t]{0.38\textwidth}
    \scriptsize
    \renewcommand{\arraystretch}{1.2}
    \setlength\tabcolsep{1.8pt}
    \resizebox{1.0\linewidth}{!}{

\begin{tabular}{l|c|c|c}
    method & backbone & iters & mIoU  \\
    \hline
    
    Mask2Former (T)$^*$~\cite{cheng2022mask2former} & Swin-T & & 47.7 \\
    X-Decoder (T)$^*$~\cite{zou2023xdecoder} & Focal-T & & 51.0 \\
    OpenSeeD (T)$^*$~\cite{zhang2023openseed} & Swin-T &  \multirow{-3}{*}{160k} & 52.2 \\
    \hline

    \rowcolor{gray!15} 
    \modelname & Swin-T & - & 38.9 \\
    \rowcolor{gray!15} 
    \modelname$^*$ & Swin-T & 45k & 52.3 \\

\end{tabular}

    }
    \vspace{6pt}
    \captionsetup{width=0.96\textwidth}
    \caption{\textbf{Comparison of semantic segmentation performance on ADE20K.} 
    $^*$The model is finetuned on the dataset.
    }
    \label{tab:semseg}
    \end{minipage}
    \hspace{0.01\textwidth}
    \begin{minipage}[t]{0.58\textwidth}
    \scriptsize
    \renewcommand{\arraystretch}{1.2}
    \setlength\tabcolsep{2.5pt}
    \resizebox{1.0\linewidth}{!}{

\begin{tabular}{l|cccccccc}
      & \multicolumn{2}{c}{Point} & \multicolumn{2}{c}{Scribble}
      & \multicolumn{2}{c}{Box} & \multicolumn{2}{c}{Mask} \\
     \multirow{-2}{*}{method} & mIoU & cIoU & mIoU & cIoU & mIoU & cIoU & mIoU & cIoU \\
     \hline

     SAM-B~\cite{kirillov2023sam} & 48.7 & 33.6 & - & - & 73.7 & 68.7 & - & - \\
     SAM-L~\cite{kirillov2023sam} & 51.8 & 37.7 & - & - & 76.6 & 71.6 & - & - \\
     SEEM-B~\cite{zou2024segment} & 47.8 & 57.8 & 43.0 & 44.0 & 44.9 & 42.1 & 48.4 & 65.0 \\
     PSALM~\cite{zhang2024psalm} & 64.3 & 74.0 & 66.9 & 80.0 & 67.3 & 80.9 & 67.6 & 82.4 \\
     \hline

     \rowcolor{gray!15}
     \modelname & 49.1  & 60.7  & 54.7 & 72.3 & 59.1 & 78.2 & 59.6 & 81.0 \\
     \rowcolor{gray!15}
     \modelname$^*$ & 65.4 & 70.9 & 66.8 &  77.2 &  74.2 &  83.2 &  67.9 &  83.8 \\
\end{tabular}

    }
    \vspace{6pt}
    \captionsetup{width=0.95\textwidth}
    \caption{\textbf{Comparison of interactive segmentation performance.} 
    The task is evaluated on the COCO-interactive dataset proposed by~\cite{zhang2024psalm}. $^*$The model is finetuned on the task.
    }
    \label{tab:interactive_seg}
    \end{minipage}
\end{table}

\noindent
\textbf{Semantic Segmentation.} In addition to the instance-level segmentation, our model also has the capacity to address the task of semantic segmentation. We present the results on ADE20K~\cite{zhou2017ade20k} in Table~\ref{tab:semseg}. The previous works mainly follow the standard training setting for 160k iterations on 8 GPUs with a total batch size of 16. As ADE20K only constitutes a small proportion of our joint-training dataset, our generalist model has a slightly inferior performance on this dataset. By fine-tuning this dataset with fewer training iterations, \textit{i.e.}, 45k, \modelname can achieve a mIoU of 52.3 points, surpassing the previous methods under the same backbone.

\noindent
\textbf{Interactive Segmentation.} Interactive segmentation~\cite{kirillov2023sam} is an emerging task that uses visual prompts as conditions for instance segmentation. We compare our method with state-of-the-art approaches on the COCO-interactive dataset~\cite{zhang2024psalm} in Table~\ref{tab:interactive_seg}. This dataset, proposed by~\cite{zhang2024psalm}, utilizes points, scribbles, boxes, and masks as visual prompts and is annotated on the COCO dataset~\cite{lin2014coco}. As shown in the table, our generalist model \modelname demonstrates performance advantages over SEEM-B~\cite{zou2024segment} across all metrics but falls behind the recently proposed MLLM method PSALM~\cite{zhang2024psalm}. We hypothesize that this is due to our region encoder being frozen during stage-3 of training, which constrains the model's performance. Therefore, we further fine-tune our model on this task by unfreezing the region encoder. It is observed that the performance of our model is significantly improved, as illustrated in the last row of Table~\ref{tab:interactive_seg}.

\begin{table*}[t]
\scriptsize
\centering
\renewcommand{\arraystretch}{1.2}
\setlength\tabcolsep{0.6pt}

\resizebox{1.0\linewidth}{!}{
\begin{tabular}{l|cccc|cccc|cccc|cccc|cccc}
      & \multicolumn{4}{c|}{DUTS} & \multicolumn{4}{c|}{DUT-OMRON}
      & \multicolumn{4}{c|}{HKU-IS} & \multicolumn{4}{c|}{ECSSD} 
      & \multicolumn{4}{c}{PASCAl-S}
      \\
     \multirow{-2}{*}{method} 
     & $S_m\uparrow$ & $E_m\uparrow$ & $F_{\beta}^{\omega}\uparrow$ & $\mathcal{M}\downarrow$ 
     & $S_m\uparrow$ & $E_m\uparrow$ & $F_{\beta}^{\omega}\uparrow$ & $\mathcal{M}\downarrow$  
     & $S_m\uparrow$ & $E_m\uparrow$ & $F_{\beta}^{\omega}\uparrow$ & $\mathcal{M}\downarrow$  
     & $S_m\uparrow$ & $E_m\uparrow$ & $F_{\beta}^{\omega}\uparrow$ & $\mathcal{M}\downarrow$  
     & $S_m\uparrow$ & $E_m\uparrow$ & $F_{\beta}^{\omega}\uparrow$ & $\mathcal{M}\downarrow$  \\
     \hline

     PoolNet~\cite{liu2019simple} & .878 & .889 & .880 & .040 & .828 & .863 & .808 & .056 & .910 & .949 & .933 & .032 & .922 & .924 & .944 & .039 & .847 & .850 & .869 & .074 \\
    LDF~\cite{wei2020label}  & .892 & .910 & .898 & .034 & .838 & .873 & .820 & .051 & .919 & .954 & .939 & .027 & .924 & .925 & .950 & .034 & .856 & .865 & .874 & .059\\
    VST~\cite{liu2021visual}  & .896 & .892 & .890 & .037 & .850 & .861 & .825 & .058 & .928 & .953 & .942 & .029 & .932 & .918 & .951 & .033 & .865 & .837 & .875 & .061\\
    SelfReformer~\cite{yun2022selfreformer} & .911 & .920 & .916 & .026 & .856 & .886 & .836 & .041 & .930 & .959 & .947 & .024 & .935 & .928 & .957 & .027 & .874 & .872 & .894 & .050 \\
    BBRF~\cite{ma2023boosting} & .908 & .927 & .916 & .025 & .855 & .887 & .843 & .042 & .935 & .965 & .958 & .020 & .939 & .934 & .963 & .022 & .871 & .867 & .891 & .049\\
    EVPv2~\cite{liu2023explicit} & .915 & .948 & .923 & .027 & .862 & .895 & .857 & .047 & .932 & .963 & .953 & .023 & .935 & .957 & .958 & .028 & .879 & .917 & .869 & .053\\
    \hline

    \rowcolor{gray!15} 
    \modelname & .921 & .955 & .911 & .024 & .882 & .920 & .846 & .041 & .941 & .975 & .946 & .016 & .950 & .974 & .959 & .018 & .892 & .933 & .877 & .044 \\
    
\end{tabular}
}

\caption{\textbf{Comparison of salient object detection performance.} The  metrics include S-measure ($S_m$), weighted F-measure ($F_{\beta}^{\omega}$), E-measure($E_m$) and mean absolute error ($\mathcal{M}$).
}
\label{tab:sod}
\vspace{-2mm}
\end{table*}

\begin{table*}[t]
\scriptsize
\centering
\renewcommand{\arraystretch}{1.2}
\setlength\tabcolsep{4mm}

\begin{tabular}{l|cccc|cccc}
      & \multicolumn{4}{c|}{CAMO} & \multicolumn{4}{c}{COD10K} \\
     \multirow{-2}{*}{method} 
     & $S_m\uparrow$ & $F_{\beta}^{\omega}\uparrow$ & $E_m\uparrow$ & $\mathcal{M}\downarrow$ 
     & $S_m\uparrow$ & $F_{\beta}^{\omega}\uparrow$ & $E_m\uparrow$ & $\mathcal{M}\downarrow$  \\
     \hline
     ZoomNet~\cite{pang2022zoomnet} & .820 & .752 & .878 & .066 & .838 & .729 & .888 & .029 \\
     HitNet~\cite{hu2023high} & .849 & .809 & .906 & .055 & .871 & .806  & .935 & .023 \\
     FSPNet~\cite{huang2023feature} & .856 & .799 & .899 & .050 & .851 & .735  & .895 & .026 \\
     ZoomNeXt~\cite{pang2023zoomnext} & .889 & .857 & .945 & .041 & .898 & .827 & .956 & .018 \\

     \hline
     \rowcolor{gray!15}
     \modelname & .856 & .829 & .914 & .057 & .877 & .818 & .934 & .024 \\
     
\end{tabular}

\caption{\textbf{Comparison of camouflaged object detection performance.} The evaluation metrics including S-measure ($S_m$), weighted F-measure ($F_{\beta}^{\omega}$), E-measure($E_m$) and mean absolute error ($\mathcal{M}$).
}
\label{tab:cod}
\vspace{-4mm}
\end{table*}

\begin{table*}[t]
\scriptsize
\centering
\renewcommand{\arraystretch}{1.2}
\setlength\tabcolsep{1.5pt}

\resizebox{1.0\linewidth}{!}{
\begin{tabular}{l|c|ccccccccccccc|c}
Method & Backbone & PascalVOC & AerialDrone & Aquarium & Rabbits & EgoHands & Mushrooms & Packages & Raccoon & Shellfish & Vehicles & Pistols & Pothole & Thermal & AP$_{\mathrm{avg}}$ \\
\hline
GLIP-T 
& Swin-T 
& 56.2
& 12.5
& 18.4
& 70.2
& 50.0
& 73.8
& 72.3
& 57.8
& 26.3
& 56.0
& 49.6
& 17.7
& 44.1
& 46.5
  
\\

GLEE-Lite 
& ResNet50
& 61.7
& 7.9 
&23.2 
& 72.6 
& 41.9 
& 51.6 
& 32.9 
& 51.1 
& 35.0 
& 59.4 
& 45.6 
& 21.8 
& 56.9 
& 43.2
\\

GLEE-Plus 
& Swin-L 
& 67.8 
& 10.8 
& 38.3 
& 76.1 
& 47.4 
& 19.2 
& 29.4 
& 63.8 
& 66.7 
& 63.8 
& 62.6 
& 15.3 
& 66.5 
& 48.3 
\\

\hline
\rowcolor{gray!15}
\modelname & Swin-T & 54.2 & 16.5 & 27.0 & 79.6 & 44.7 & 29.0 & 64.7  & 54.2 & 49.7 & 61.2 & 64.8 & 14.6 & 57.1 & 48.3 \\

\end{tabular}
}

\caption{\textbf{Comparison of zero-shot object detection performance on OdinW13.} 
}
\label{tab:odinw13}
\end{table*}

\vspace{-2mm}
\subsection{Evaluation on Various Domains.}

\noindent
\textbf{Salient Object Detection.} We compare the results of \modelname with state-of-the-art methods for salient object detection (SOD) in Table~\ref{tab:sod}. Our model clearly achieves the highest performance on 4 of the 5 classical benchmarks, demonstrating its strong object discovery capabilities.

\noindent
\textbf{Camouflaged Object Detection.} The performance comparisons for camouflaged object detection (COD) are presented in Table~\ref{tab:cod}. It is observed that \modelname exhibits competitive performance with state-of-the-art expert models that undergo longer training schedule, \textit{e.g.}, 150 epochs.

\begin{figure}[b]
    \centering
    \includegraphics[width=\textwidth]{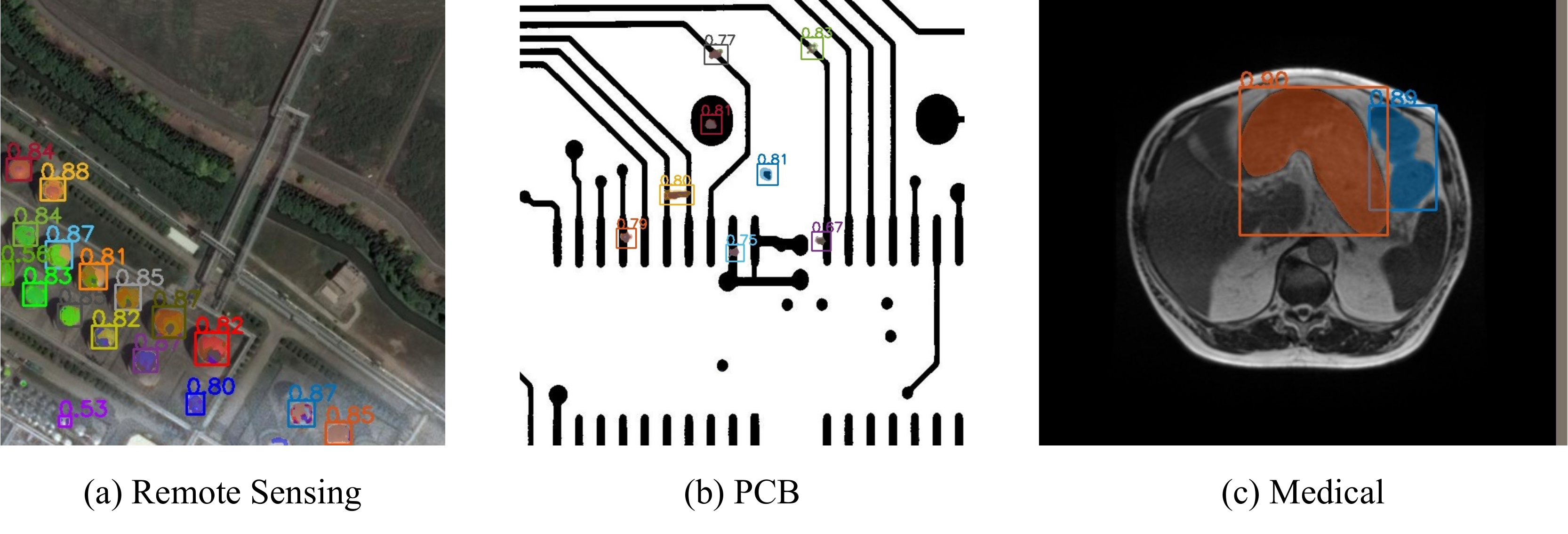}
    \vspace{-12pt}
    \caption{\textbf{Visualization results across various domains.} 
    \modelname shows a strong generalization ability for remote sensing, PCB, and medical images.
    }
    \vspace{-4pt}
    \label{fig:vis_domain}
    
\end{figure}

\noindent
\textbf{Visualization across various domains.} 
Besides the quantitative results, we also display the visualization results of \modelname across various domains. As illustrated in Figure~\ref{fig:vis_domain}, our model also shows strong perception capacities for remote sensing, PCB, and medical images.

\subsection{Zero-shot Evaluation}

\noindent
\textbf{Zero-shot Image Captioning.} 
Benefiting from the joint training on large-scale vision-language datasets, \modelname exhibits promising capacities for zero-shot image captioning. As shown in Table~\ref{subtab:image_captioning}, both \modelname-Chat and \modelname achieve competitive performance on Flickr30K~\cite{plummer2015flickr30k} and NoCaps~\cite{agrawal2019nocaps} compared with previous methods.

\noindent
\textbf{Zero-shot Object Detection on OdinW13.}
We conduct the zero-shot object detection evaluation on OdinW13 dataset~\cite{li2022glip}, as shown in Table~\ref{tab:odinw13}. The results demonstrate that our \modelname with a Swin-Tiny backbone is even on par with GLEE-Plus~\cite{wu2023general} with a Swin-Large backbone in AP$\rm_\text{avg}$. This indicates that our model benefits from the extensive dataset joint training, thereby providing robust general object detection capabilities.

\begin{wraptable}{R}{0.30\textwidth}
\vspace{-2.5ex}
\centering
\scriptsize
\renewcommand{\arraystretch}{1.2}
\setlength\tabcolsep{2.2pt}

\begin{tabular}{c| c}

    \hline
    \multicolumn{2}{l}{\textbf{\textit{In-Context Segmentation}}} \\
    Method & mIoU \\
    \hline
    Painter~\cite{wang2023images} & 44.26 \\
    SegGPT~\cite{wang2023seggpt} & 54.25 \\
    \rowcolor{gray!15}
    \modelname & 67.51 \\

    \hline

    \multicolumn{2}{l}{\textbf{\textit{In-Context Image Captioning}}} \\
    Method & METEROR / CIDEr \\
    OpenFlamingo~\cite{awadalla2023openflamingo} & 13.80 / 104.61 \\
    \rowcolor{gray!15}
    \modelname & 18.56 / 152.74 \\
    \hline

\end{tabular}
\caption{\textbf{Comparison of in-context segmentation and in-context image captioning performance.}}
\label{tab:in_context}
\vspace{-2.5ex}
\end{wraptable}

\noindent
\textbf{In-Context Segmentation \& In-Context Image Captioning.} To evaluate the in-context learning ability of \modelname, we compare the results of in-context segmentation and in-context image captioning in Table~\ref{tab:in_context}. For in-context segmentation, we construct a benchmark based on the validation set of COCO2017, where the number of in-context examples used during inference ranges from 1 to 5. For in-context image-captioning, we follow the same evaluation protocol as OpenFlamingo~\cite{awadalla2023openflamingo} and use 4-shot to assess the performance between different methods. The validation set is built upon COCO2017. From the table, VisionLLM v2 exhibits clear performance advantages compared with state-of-the-art methods in both in-context learning settings, which demonstrates the superior in-context capacities of our method.

\begin{table*}[t]
\scriptsize
\centering
\renewcommand{\arraystretch}{1.2}
\setlength\tabcolsep{4mm}

\begin{tabular}{l|c|cccccc}
    & & \multicolumn{2}{c}{inst seg.} & ground. & pose & \multicolumn{2}{c}{interact seg.} \\
    \multirow{-2}{*}{method} 
    & \multirow{-2}{*}{\begin{tabular}[c]{@{}c@{}}query/token\\ number\end{tabular}} 
    & AP$_{\rm b}$ & AP$_{\rm m}$ & P@.5 & AP & mIoU & cIoU \\
    \hline
    
     & 1 & 50.4 & 39.6 & 85.8 & 43.0 & 43.2 & 60.0 \\
     \rowcolor{gray!15}
     & 4 & 52.0 & 41.0 & 85.7 & 71.0 & 44.8 & 60.4 \\
     \multirow{-3}{*}{super-link queries} & 8 & 52.1 & 40.7 & 86.4 & 71.6 & 45.9 & 61.9 \\
     \hline

     & 1 & 50.8 & 39.3 & 85.4 & 42.2 & 42.1 & 57.5 \\
     & 4 & 51.5 & 41.0 & 86.2 & 71.3 & 43.7 & 59.7 \\
     \multirow{-3}{*}{token embeddings} & 8 & 52.1 & 41.1 & 86.5 & 71.5 & 44.0 & 59.1 \\

\end{tabular}

\caption{\textbf{Ablation on the comparison between super-link queries and token embeddings.} 
We evaluate the results on the four crucial visual perception tasks: instance segmentation, visual grounding, pose estimation and interactive segmentation. Our default setting is marked in \graybox{gray}.
}
\label{tab:ablation_super-link}
\vspace{-4mm}
\end{table*}

\subsection{More Ablation Studies}

\noindent
\textbf{Super-Link Queries v.s. Token Embeddings in LISA~\cite{lai2023lisa}.}
Current MLLMs~\cite{lai2023lisa, rasheed2023glamm, ren2023pixellm} introduce a segmentation token \texttt{[SEG]} into the LLM vocabulary and directly use its corresponding token embedding as a condition for SAM~\cite{kirillov2023sam} to achieve pixel-level segmentation, which we refer to as the token embedding method. We also ablate this method for linking the LLM with task-specific decoders, as shown in Table~\ref{tab:ablation_super-link}. The performance difference between the two methods is negligible for tasks using text prompts, such as instance segmentation. We hypothesize that this is because the category names are seen during training, allowing the token embeddings to effectively capture category semantics. However, our super-link queries method outperforms the token embedding method for more open-ended tasks, such as interactive segmentation with visual prompts, demonstrating the greater flexibility of our approach.

We emphasize two fundamental differences between the two methods: (1) The token embedding method requires sequential prediction of the special tokens during inference, which is time-consuming when the number of tokens is large. In contrast, our super-link technique requires only a single forward pass and the super-link queries would be automatically appended. This is efficient for cases requiring many tokens, such as image generation. (2) The super-link queries are not constrained by the cross-entropy loss of the LLM, allowing for more flexible and stronger representations for open-ended tasks.

\vspace{-2mm}
\subsection{Qualitative Results}

\begin{figure}[t]
    \centering
    \includegraphics[width=1.0\textwidth]{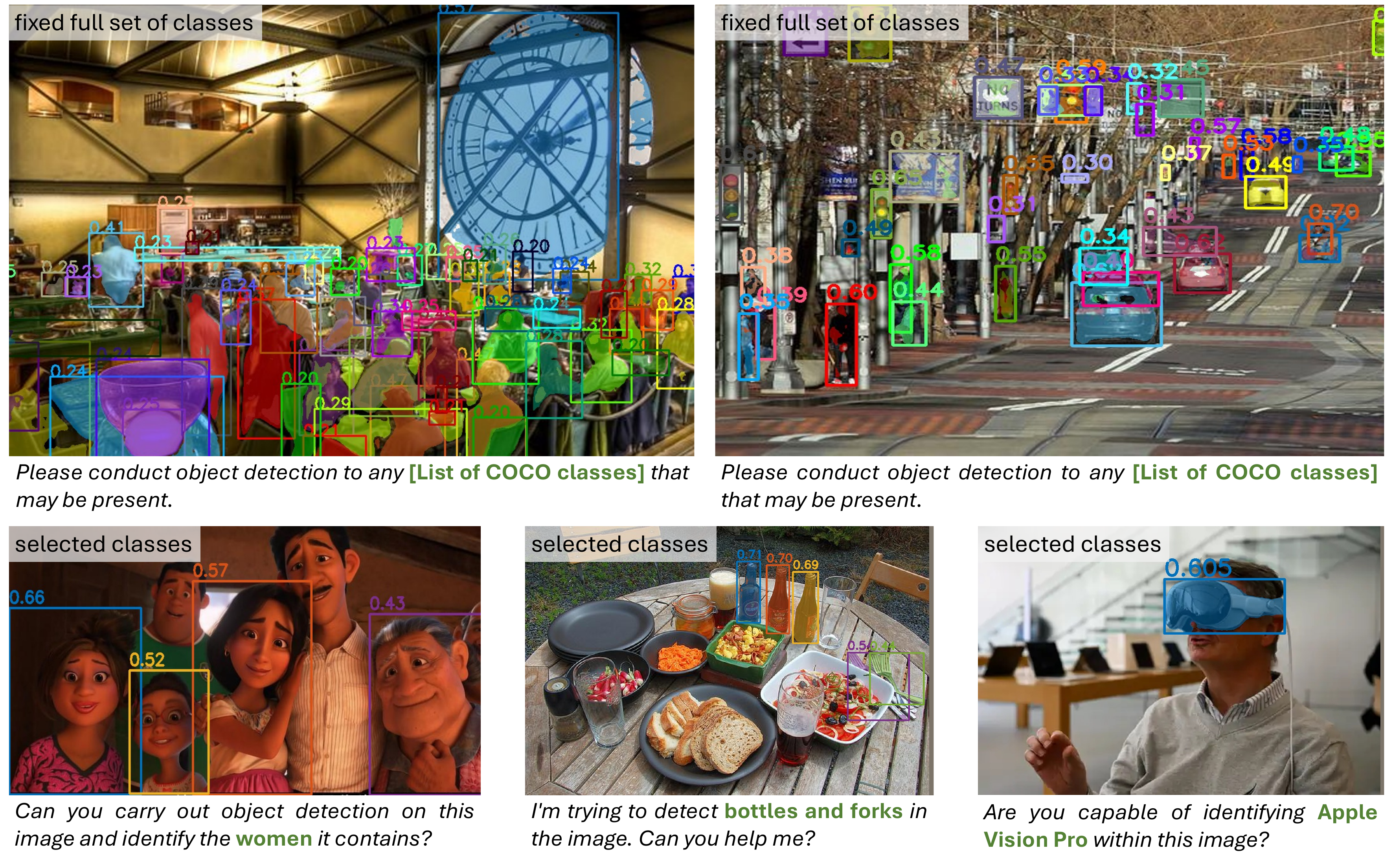}
    \caption{\textbf{Object detection and instance segmentation.} The model excels in various environments, supporting the detection of a large number of instances. Its flexibility is highlighted by its ability to detect only user-selected categories and identify novel classes.} 
    \label{fig:demo_det}
    \vspace{-5pt}
\end{figure}

\begin{figure}[t]
    \centering
    \includegraphics[width=1.0\textwidth]{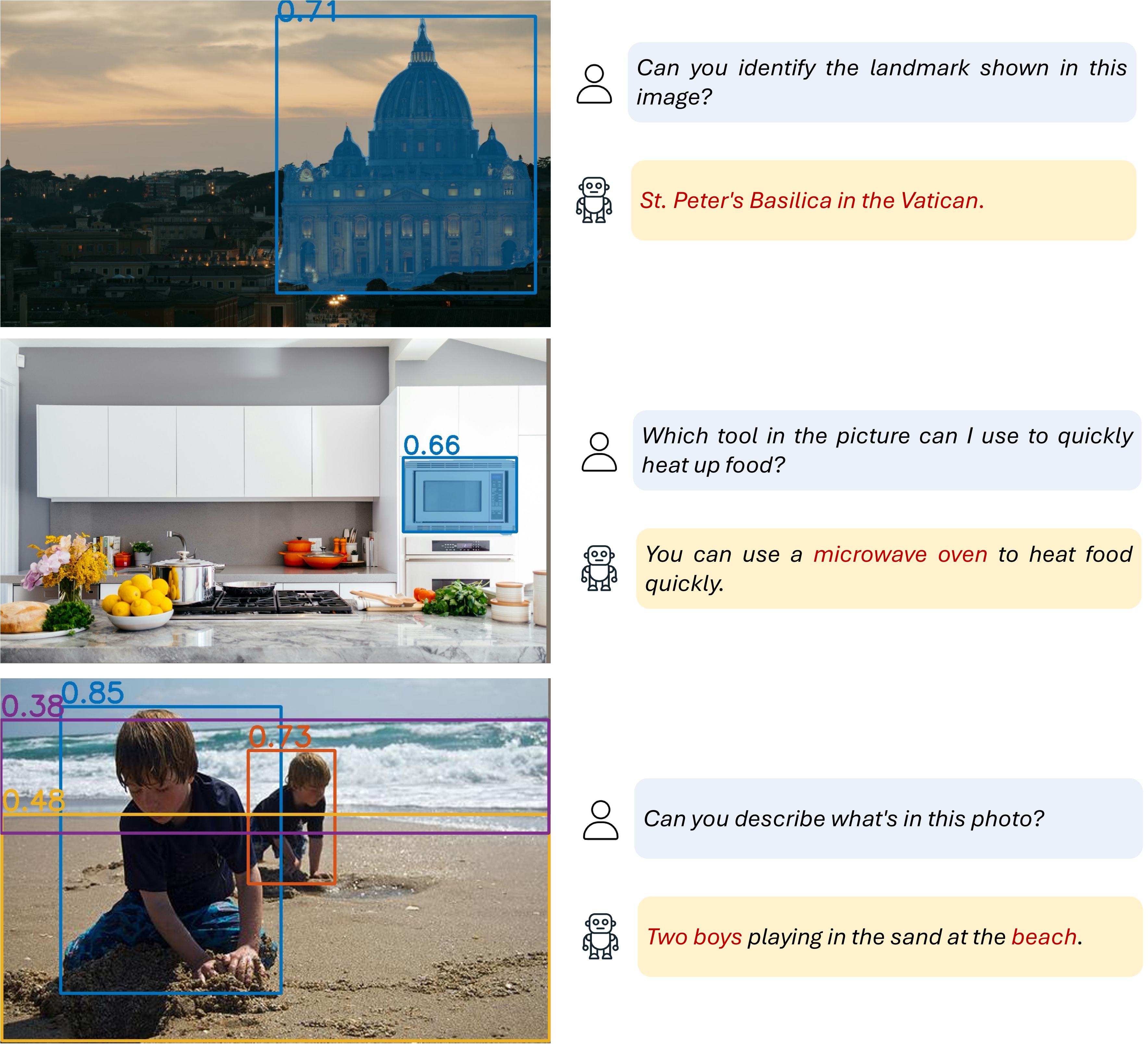}
    \caption{\textbf{Grounded caption.} The model accurately locates objects based on user prompts, outputs bounding boxes, and provides answers to user queries.} 
     \label{fig:demo_grd_cap}
    \vspace{-5pt}
\end{figure}

\noindent
\textbf{Visual Perception.}
We evaluate \modelname on various visual perception tasks and display the visualization results from Figure~\ref{fig:demo_det} to Figure~\ref{fig:demo_grd_cap}. The qualitative examples showcase that \modelname exhibits strong visual perception capacities, from coarse to fine-grained perception (box, keypoint, pixel), from basic to novel classes, from commonly-seen domains to long-tailed domains (natural scenes, industry, agriculture, \etc).

\begin{figure}[t]
    \centering
    \includegraphics[width=1\textwidth]{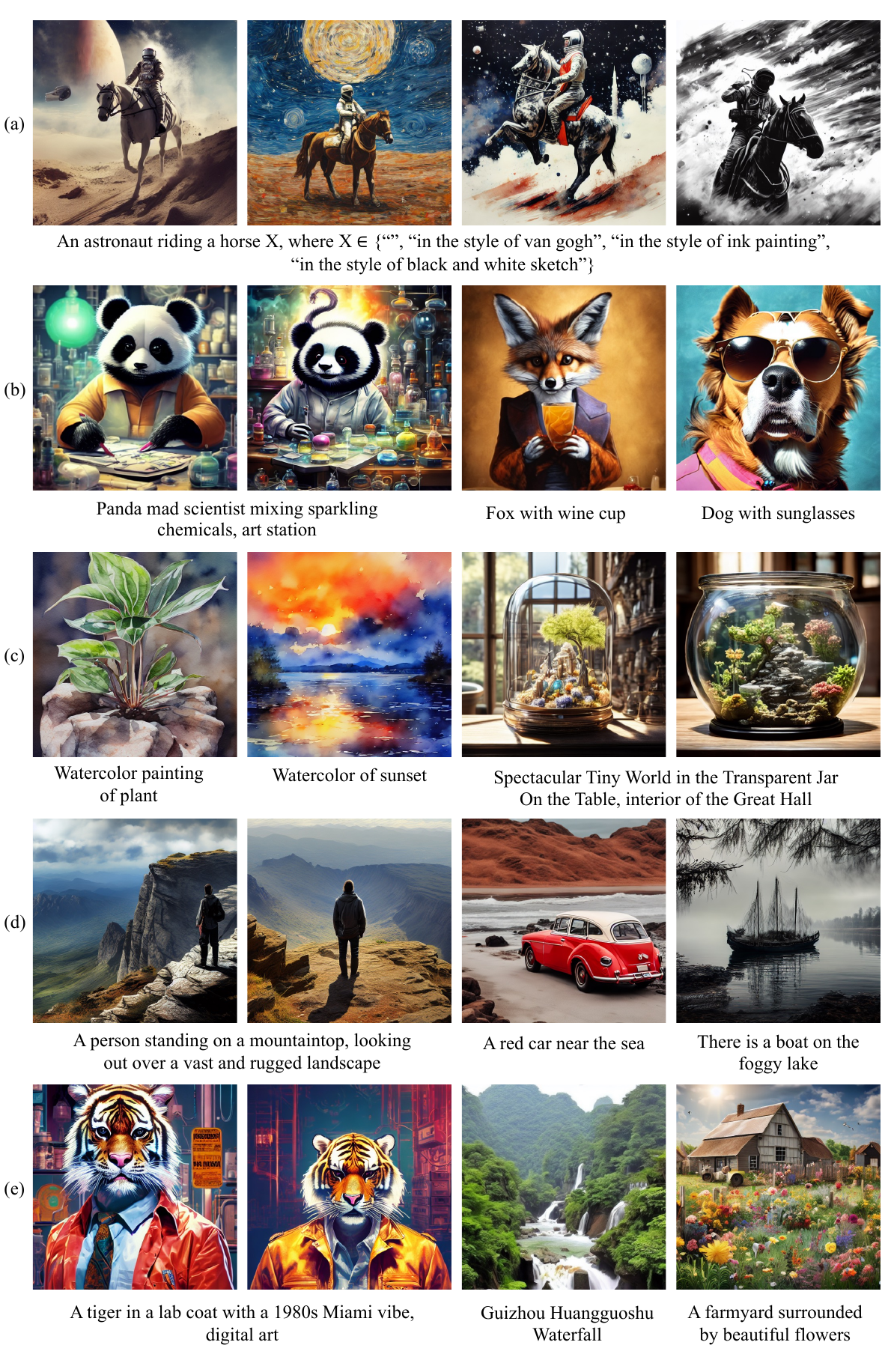}
    \caption{\textbf{VisionLLM v2 text-to-image generation examples.} VisionLLM v2 could generate high-quality images that not only properly follow the concepts and relations, but also different styles specified in the instructions. . } 
    \label{fig:t2i_more}
\end{figure}

\begin{figure}[t]
    \centering
    \includegraphics[width=1\textwidth]{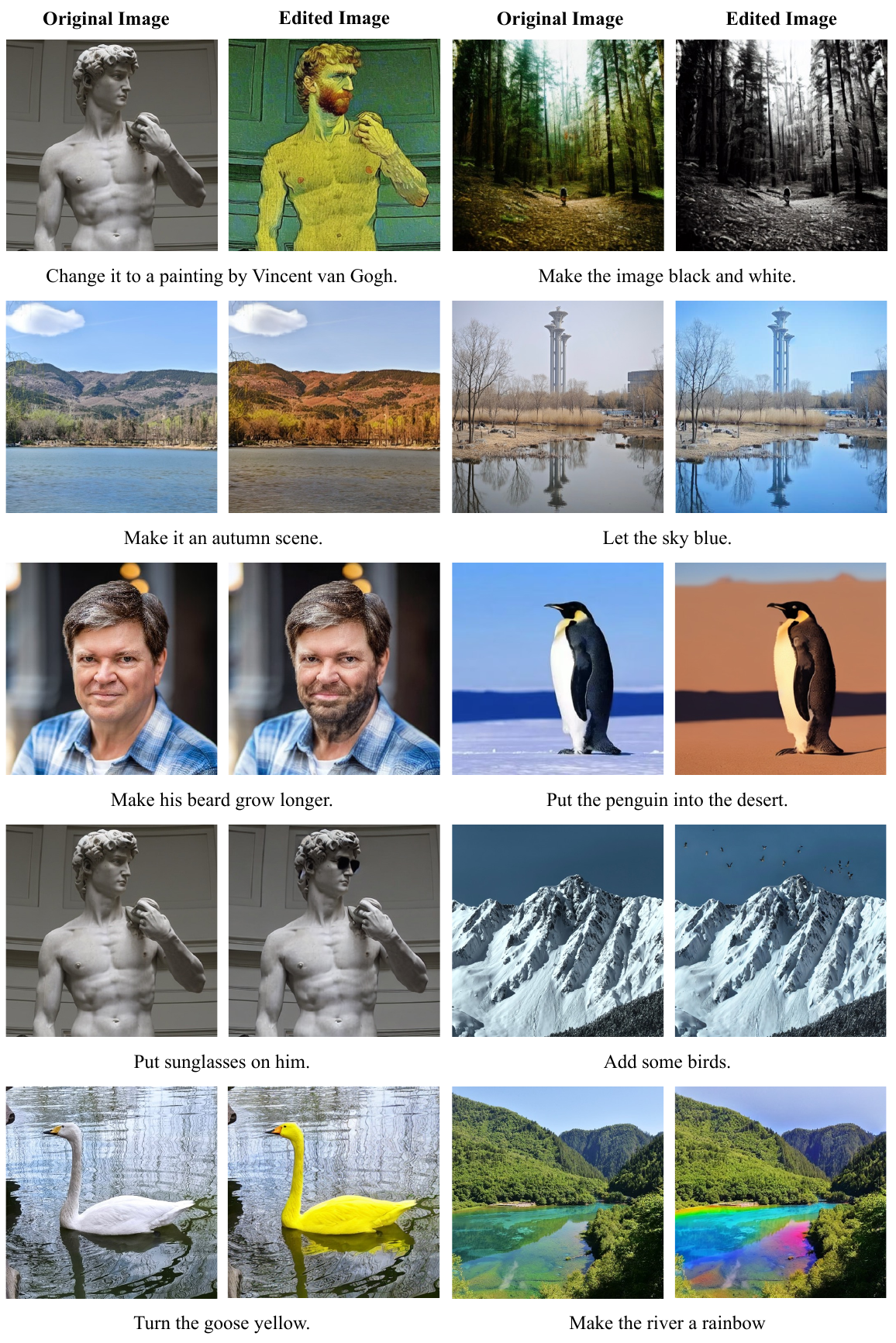}
    \caption{\textbf{VisionLLM v2 instructed-based image editing examples.} VisionLLM v2 can understand a variety of instructions such as style transfer, object replacement, object addition, attribute change, and more to generate high-quality edited images.} 
    \label{fig:editing_more}
\end{figure}

\noindent
\textbf{Visual Generation.} Figure~\ref{fig:t2i_more} shows more text-to-image generation results of VisionLLM v2. It could be observed that our model could generate high-quality images that not only properly follow the concepts and relations but also different styles specified in the instructions. 
Figure~\ref{fig:editing_more} shows more instructed-based image editing results of VisionLLM v2. Our model could successfully perform image editing for over five types of editing instructions, such as style transfer, object replacement, object addition, and attribute change.

\clearpage

\begin{figure}[t]
    \centering
    \includegraphics[width=1.0\textwidth]{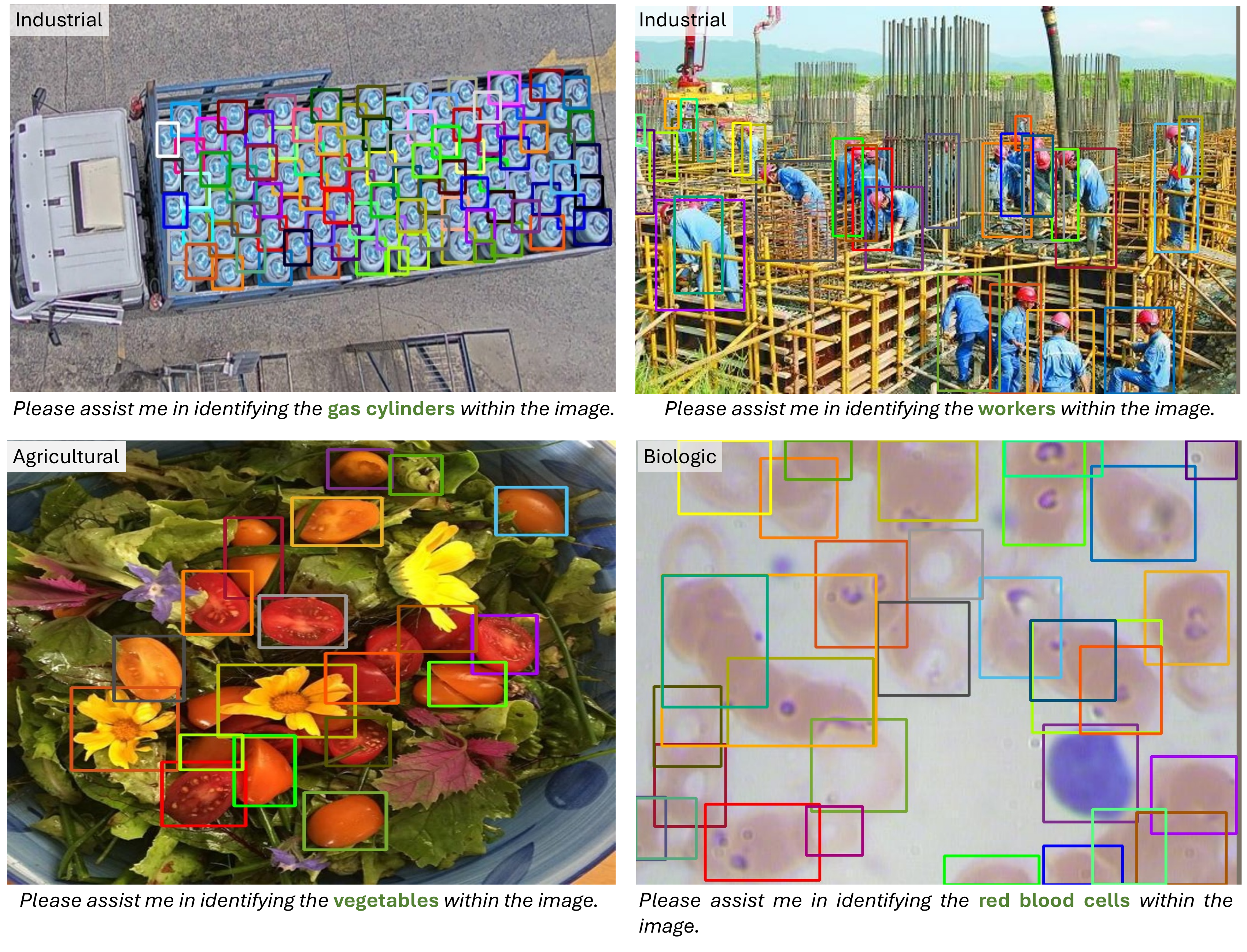}
    \caption{\textbf{Object detection on multiple domains.} The image illustrates the domain adaptability of our model, which supports perception across multiple fields such as industrial, agricultural, and biological environments.} 
    \vspace{-5pt}
\end{figure}

\begin{figure}[t]
    \centering
    \includegraphics[width=1.0\textwidth]{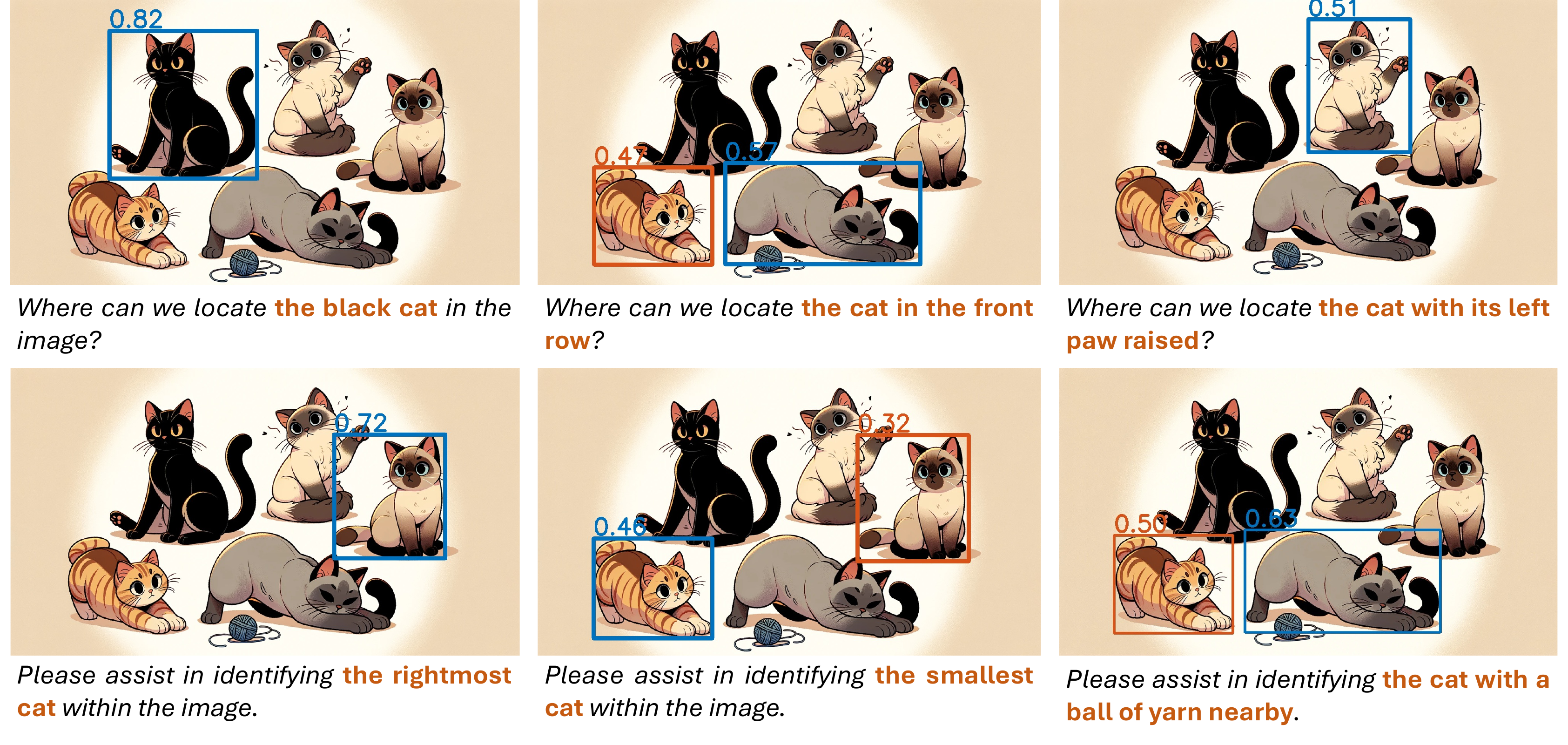}
    \caption{\textbf{Visual Grounding.} On the visual grounding task, our model demonstrates good accuracy and a certain level of reasoning capability.} 
    \vspace{-5pt}
\end{figure}

\begin{figure}[t]
    \centering
    \includegraphics[width=1.0\textwidth]{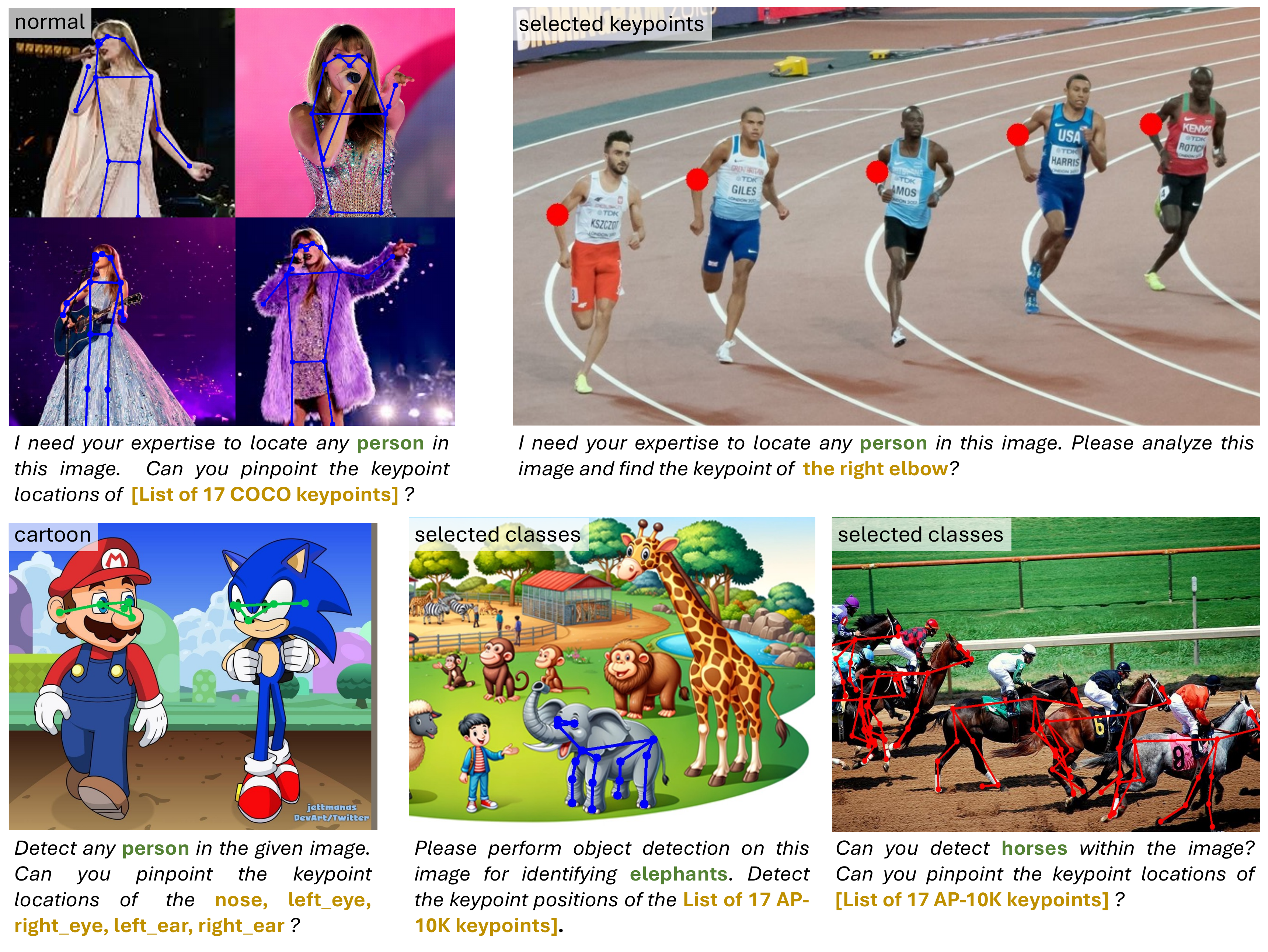}
    \caption{\textbf{Pose estimation.} Our model is capable of detecting keypoints in humans and animals with flexibility. The model allows users to select specific instance categories for detection, as well as choose individual keypoints.} 
    \vspace{-5pt}
\end{figure}

\newpage

\clearpage

\begin{figure}[t]
    \centering
    \includegraphics[width=1.0\textwidth]{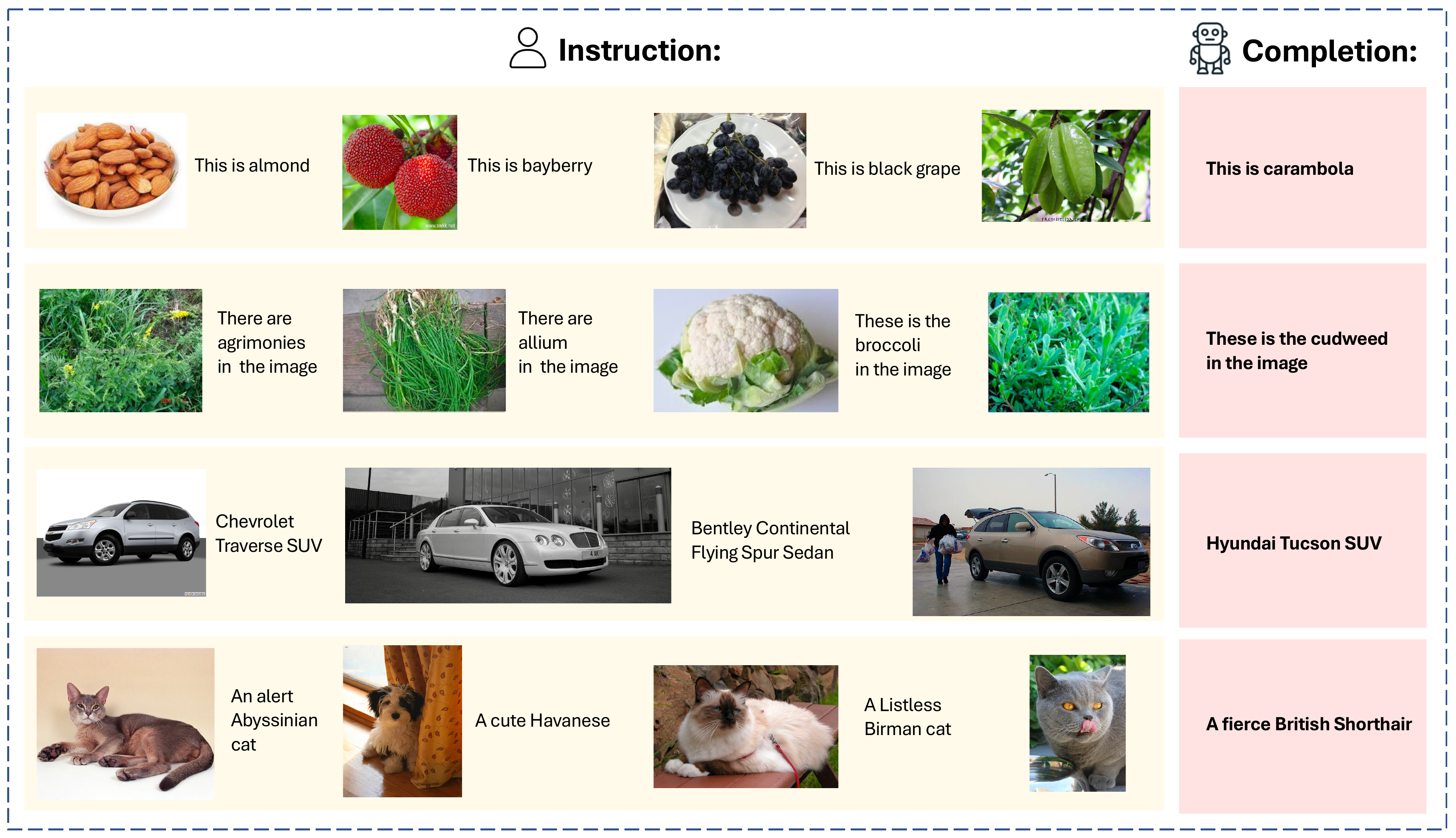}
    \caption{\textbf{In-context fine-grained visual recognition.} It demonstrates that our model has the strong capability of fine-grained recognition.} 
     \label{fig:ic_fg}
\end{figure}

\begin{figure}[t]
    \centering
    \includegraphics[width=1.0\textwidth]{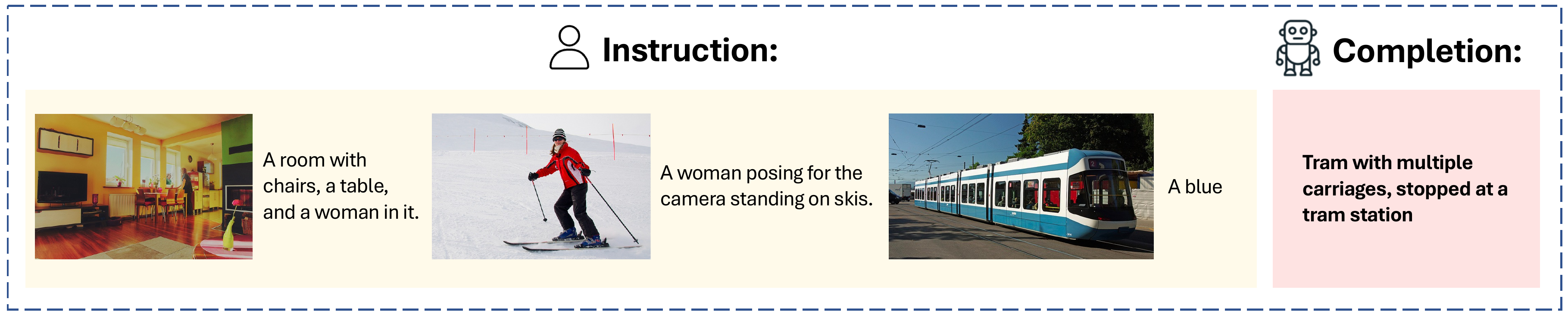}
    \caption{\textbf{In-context image captioning.} Our model is able to perform text completion based on in-context examples.} 
     \label{fig:ic_captioning}
\end{figure}

\begin{figure}[t]
    \centering
    \includegraphics[width=1.0\textwidth]{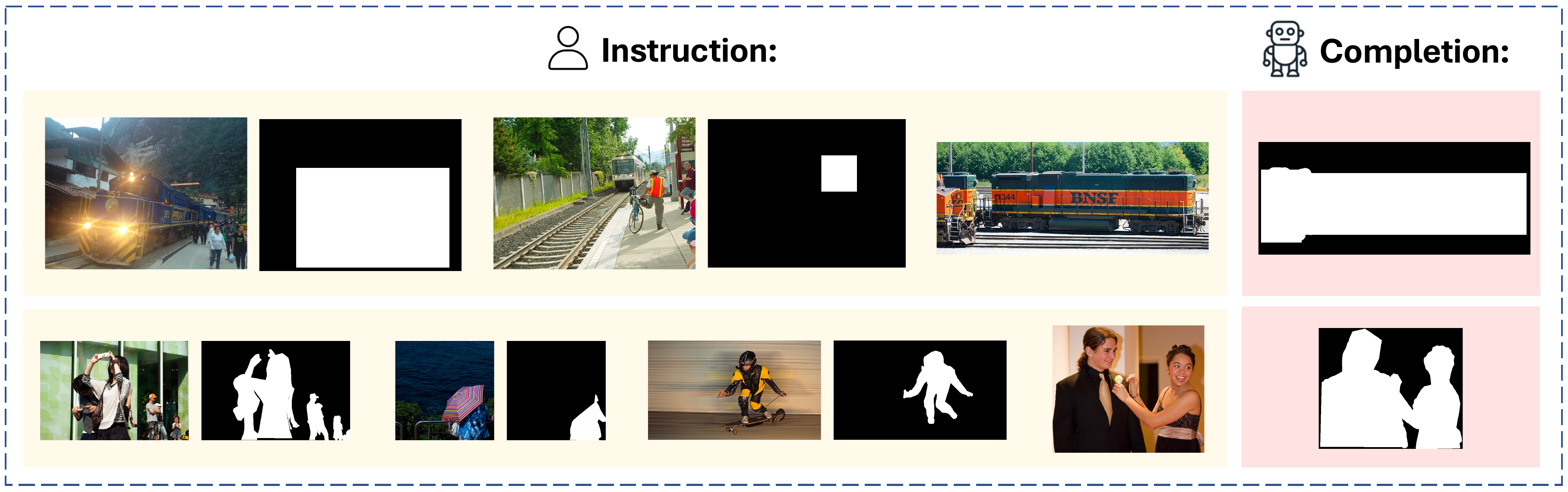}
    \caption{\textbf{In-context detection and segmentation.} We just need to provide some examples where the instances falling into the same class are highlighted. Then our model can learn from the example and use the too of detection or segmentation to process the input image.}
     \label{fig:ic_det_seg}
\end{figure}

\begin{figure}[t]
    \centering
    \includegraphics[width=1.0\textwidth]{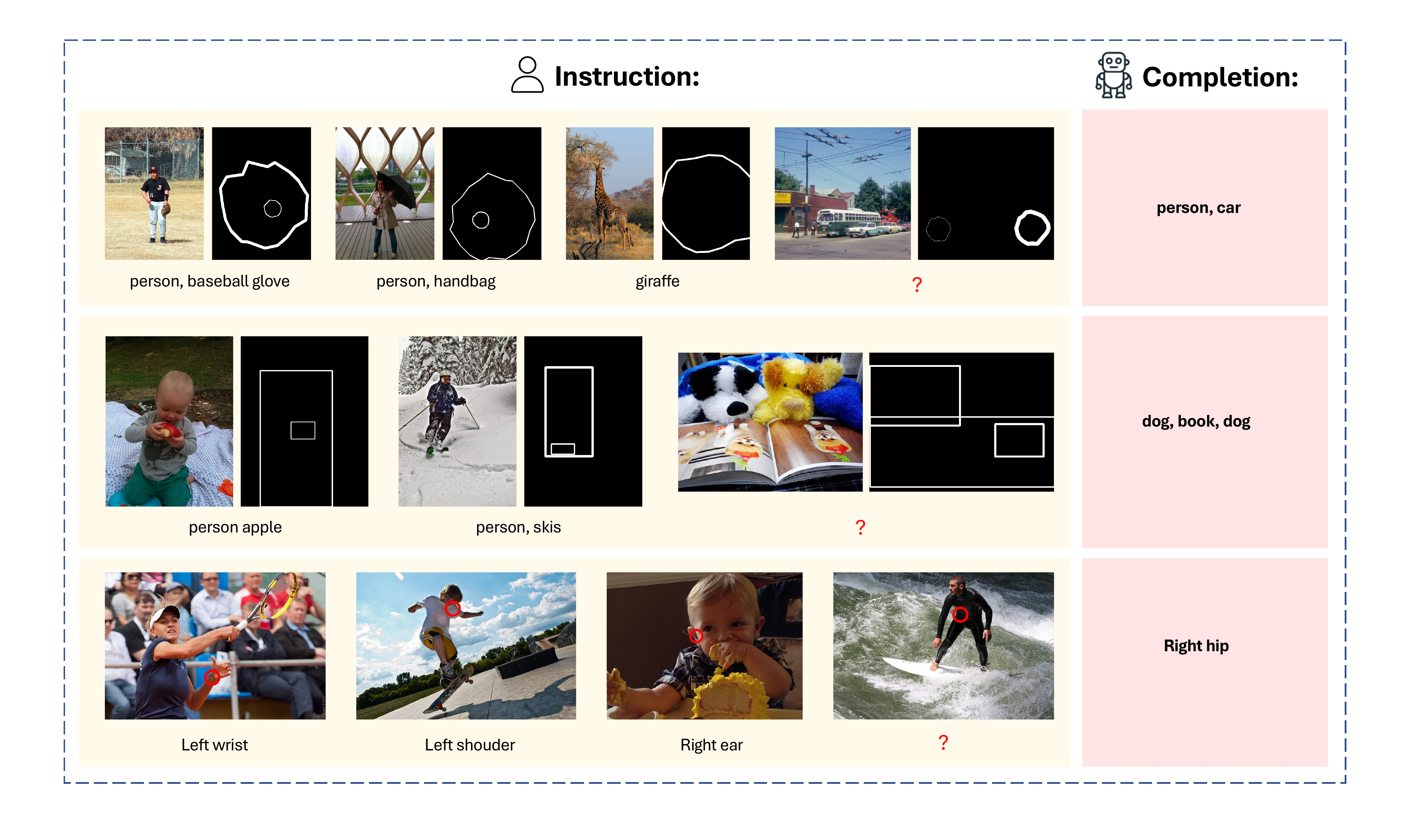}
    \caption{\textbf{In-context regional perception.} In our dataset, we construct various visual masks in input prompts. Our models are required to infer from the given examples and complete the text for the last image.} 
     \label{fig:ic_imt}
\end{figure}

\noindent
\textbf{Multimodal In-context Learning Ability.} 
To qualitatively verify the in-context capabilities of our model after trained on MMIC, we provide comprehensive visualizations across different tasks. As demonstrated in Figure~\ref{fig:ic_fg},~\ref{fig:ic_captioning},~\ref{fig:ic_det_seg} and ~\ref{fig:ic_imt}, our method can handle both visual and textual prompts, enabling it to perform tasks that require understanding and integration of information from different modalities.
In addition, our models can distinguish between different prompting strategies and can correctly use the corresponding detection or segmentation tools to obtain the expected output based on given in-context examples.

\clearpage

\section{More Architecture Details}
\label{sec:detail_arch}

\subsection{Region Encoder} 

The region encoder is designed to encode various shaped visual prompts such as points, scribbles, boxes, \etc. Each visual prompt is represented by a binary mask. We first concatenate the binary mask with the image along the channel dimension, resulting in a 4-channel input, denoted as $I_\text{vprt} \in \mathbb{R}^{4 \times H \times W}$. The region encoder is implemented with three convolutional layers: the first layer uses a kernel size of 7 and a stride of 7, the second layer employs a kernel size of 2, and a stride of 2, and the final layer features a kernel size of 1 and a stride of 1. Each convolutional layer is followed by layer normalization~\cite{ba2016layer} and GELU activation~\cite{hendrycks2016gelu}. This process downsamples the input $I_\text{vprt} \in \mathbb{R}^{4 \times H \times W}$ by a factor of 14. We further augment this feature map by adding the feature map of the global image $I_\text{global}$. Finally, we use grid sampling to extract features within the masked regions and pool them into a single region embedding $F_\text{vprt} \in \mathbb{R}^{1 \times C}$.

\subsection{Task-specific Decoders}

\begin{figure}[t]
\hsize=\textwidth
\centering

\begin{subfigure}{\textwidth}
    \centering
    \includegraphics[width=0.98\textwidth]{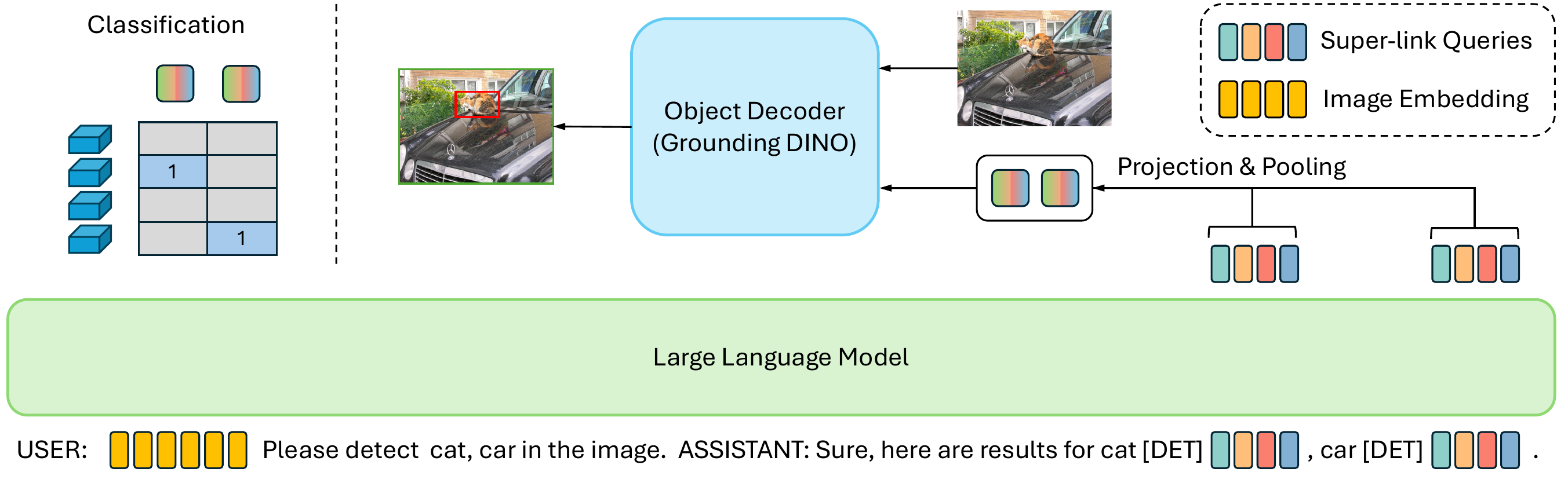}
    \caption{Connecting with object decoder for visual perception.}
    \label{fig:arch_perception}
\end{subfigure}
\\
\vspace{0.15in}
\begin{subfigure}{\textwidth}
    \centering
    \includegraphics[width=0.98\textwidth]{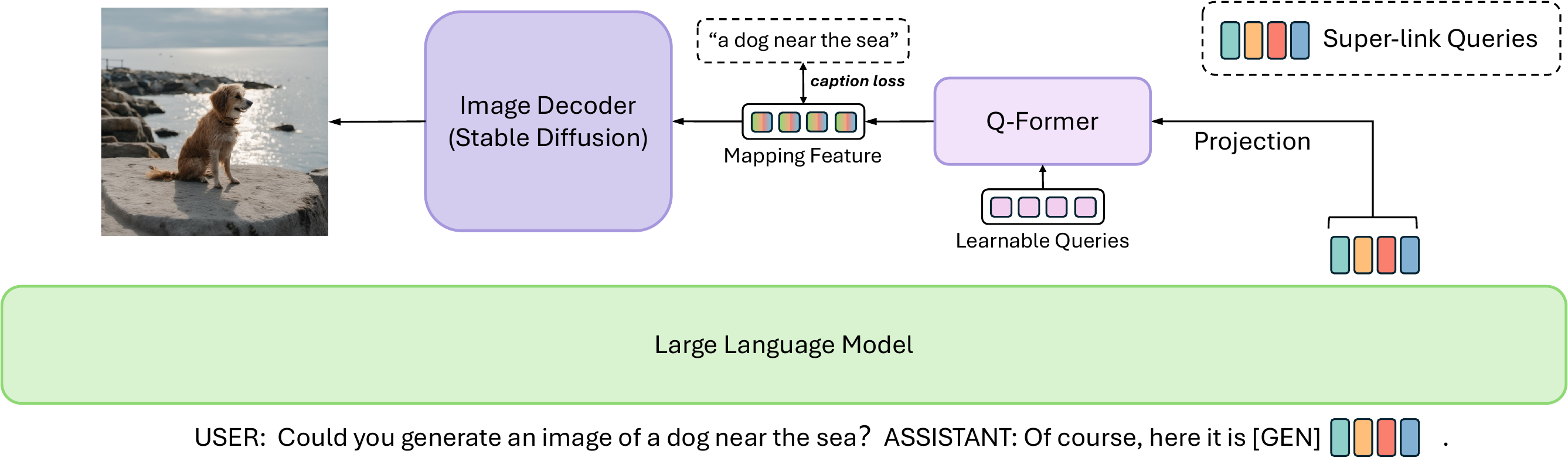}
    \caption{Connecting with image decoder for visual generation.}
    \label{fig:arch_generation}
\end{subfigure}

\caption{\textbf{Architecture details for connecting LLM with task-specific decoder via super-link queries.}
(a) Connecting with object decoder. We first extract the per-category features by performing projection and pooling on the hidden states of corresponding super-link queries. Then these features are sent into the object decoder as text features. (b) Connecting with image decoder. We add a Q-Former for projecting the features of super-link queries to the feature space of Stable Diffusion.
}
\label{fig:detail_arch}
\end{figure}

In this subsection, we provide more explanations about how to connect LLM with task-specific decoders via super-link queries, which enables the end-to-end optimization of the entire network. 

\noindent
\textbf{Connecting with Object Decoder.} For visual perception tasks like object detection, we employ Grounding DINO~\cite{liu2023groundingdino} as the object decoder to localize objects as well as classify their categories. To achieve this, LLM would output each category name in the response, followed by a special token \texttt{[DET]} and super-link queries. We then obtain the per-category features by extracting the hidden states of LLM for corresponding super-link queries and pooling them into one embedding. Grounding DINO receives both the image and the obtained per-category features as inputs and predicts the detection results. The process is illustrated in Figure~\ref{fig:arch_perception}. It is noted that we discard the text encoder in the original Ground DINO and use the obtained per-category features as text features to perform the vision-language alignment for classification. During training, the total loss includes the cross-entropy loss of LLM and detection loss of the object decoder. Similarly, the keypoint decoder is also integrated into the LLM in the same way and performs pose estimation.

\noindent
\textbf{Connecting with Image Decoder.} We utilize Stable Diffusion~\cite{rombach2022sd} as the image decoder and take the example of text-to-image generation for clarification, as depicted in Figure~\ref{fig:arch_generation}. The super-link queries are appended after the special token \texttt{[GEN]} in the LLM's response. After passing through the LLM, an MLP layer and a lightweight Q-Former~\cite{li2023blip2, koh2024gill} module are added to project the features of the super-link queries into the representation space of Stable Diffusion, i.e., mapping features. We bypass the text encoder in Stable Diffusion and directly use the mapping features as the text embedding condition. During training, in addition to the next token loss in the LLM, we employ two MSE losses for supervision: one between the encoded text features by CLIP~\cite{radford2021clip} and the mapping features, and the other between the ground-truth images/noise and predicted images/noise.

\clearpage

\section{More Dataset Details}
\label{sec:detail_dataset}

\begin{table}[t]\scriptsize
\renewcommand{\arraystretch}{1.2}

\begin{subtable}{0.95\textwidth}
    \centering
    \setlength{\tabcolsep}{3.0mm} 
    \begin{tabular}{l|c|p{9cm}}
         task & \#sample & dataset \\
         \hline
        Conversation   & 1.59M & ShareGPT4V~\cite{chen2023sharegpt4v}, 
                              Laion-GPT4V~\cite{laion_gpt4v},
                              ALLaVA~\cite{chen2024allava} \\ 
        \rowcolor{gray!15}
        Image Captioning & 0.59M & COCO~\cite{chen2015cococaption}, 
                              TextCaps~\cite{sidorov2020textcaps} \\ 
                              
         \multirow{2}{*}{Image VQA} & \multirow{2}{*}{5.19M} & ShareGPT4V~\cite{chen2023sharegpt4v}, GRIT~\cite{peng2023kosmos2}, VQAv2~\cite{goyal2017vqav2}, OK-VQA~\cite{marino2019okvqa},
                              A-OKVQA~\cite{marino2019okvqa}, GQA~\cite{hudson2019gqa},
                              AI2D~\cite{kembhavi2016ai2d}, ScienceQA~\cite{saikh2022scienceqa} \\
 
        \rowcolor{gray!15}
        & & OCR-VQA~\cite{mishra2019ocrvqa},                
                              ChartQA~\cite{masry2022chartqa},
                              DocQA~\cite{clark2017docqa}, STVQA~\cite{biten2019stvqa}, 
                              DVQA~\cite{kafle2018dvqa}, InfoVQA~\cite{mathew2022infographicvqa}\\
        \rowcolor{gray!15}
        \multirow{-2}{*}{OCR}& \multirow{-2}{*}{0.58M} & LLaVAR~\cite{zhang2023llavar},        
                              GeoQA+~\cite{cao2022geoqa+},
                              SynthDoG~\cite{kim2022synthdog} \\
        Region Captioning & 2.66M & Visual Genome~\cite{krishna2017visual},                                                RefCOCO/+/g~\cite{yu2016refcoco, mao2016refcocog},
                              Flickr30K~\cite{plummer2015flickr30k}, All-Seeing~\cite{wang2023allseeing}\\
        \rowcolor{gray!15}
        Region VQA       & 1.80M  & VCR~\cite{zellers2019vcr}, Osprey~\cite{yuan2023osprey}, All-Seeing~\cite{wang2023allseeing}  \\  
        Region Recognition & 0.40M & V3Det~\cite{wang2023v3det},  
                               COCO~\cite{lin2014coco}, LVIS~\cite{gupta2019lvis} \\
                            
    \end{tabular}
    \caption{Datasets used in stage-1. 
}
\label{subtab:stage1}
\end{subtable}


\begin{subtable}{0.95\textwidth}
    \centering
    \setlength{\tabcolsep}{3.0mm} 
    \begin{tabular}{l|c|l}
         task & \#sample & dataset \\
         \hline
         \multirow{3}{*}{\begin{tabular}[l]{@{}l@{}}Object Detection \&\\ Instance Segmentation\end{tabular}}
                              & \multirow{3}{*}{1.18M} & COCO~\cite{lin2014coco}, LVIS~\cite{gupta2019lvis}, 
                                    Objects365~\cite{shao2019objects365}, OpenImages~\cite{kuznetsova2020openimage},
                                    CrowdHuman~\cite{shao2018crowdhuman}, \\
                               & &  NC4K~\cite{lv2021nc4k}, COD10K~\cite{fan2021cod10k},
                                   CAMO~\cite{le2019camo}, CPD1K~\cite{zheng2018cpd1k},
                                   DUTS~\cite{wang2017duts}, \\
                          & &      MSRA10K~\cite{cheng2014msra10k}, DOTA~\cite{xia2018dota},
                                   SARDet-100K~\cite{li2024sardet},         
                                   DeepPCB~\cite{tang2019deeppcb} \\
                                   
         \rowcolor{gray!15}
         Grounded Caption & 0.18M &   Flickr30K~\cite{plummer2015flickr30k},
                                  Groma-Instruct~\cite{ma2024groma}    \\  
        Semantic Segmentation & 0.13M & ADE20K~\cite{zhou2017ade20k}, 
                                  CityScapes~\cite{cordts2016cityscapes},
                                  Mapillary~\cite{neuhold2017mapillary},
                                  LoveDA~\cite{wang2021loveda},
                                  Medical MRI~\cite{ye2023samed2d20m} \\
        \rowcolor{gray!15}
        Interactive Segmentation & 0.34M & COCO~\cite{lin2014coco}, SA-1B~\cite{kirillov2023sam} \\
         Visual Grounding & 0.13M  &   RefCOCO/+/g~\cite{yu2016refcoco, mao2016refcocog},
                                  ReasonSeg~\cite{lai2023lisa} \\
        \rowcolor{gray!15}
                         &   &    COCO~\cite{lin2014coco}, CrowdPose~\cite{li2019crowdpose},
                                  Human-Art~\cite{ju2023humanart}, AP-10K~\cite{yu2021ap10k},
                                  APT-36K~\cite{yang2022apt36k}, \\ 
        \rowcolor{gray!15}
                       & &        MacaquePose~\cite{labuguen2021macaquepose}, 
                                  300W-Face~\cite{sagonas2013face}, 
                                  Animal Kingdom~\cite{ng2022animal},
                                  AnimalWeb~\cite{khan2020animalweb}, \\
        \rowcolor{gray!15}
        \multirow{-3}{*}{Pose Estimation} & \multirow{-3}{*}{0.25M} &
                                  Vinegar Fly~\cite{pereira2019fly},
                                  Desert Locust~\cite{graving2019deepposekit} \\
                        &  &      COCO~\cite{lin2014coco}, LVIS~\cite{gupta2019lvis},        
                                  Objects365~\cite{shao2019objects365},
                                  OpenImages~\cite{kuznetsova2020openimage},
                                  CrowdHuman~\cite{shao2018crowdhuman} \\
        \multirow{-2}{*}{Object Counting}    & \multirow{-2}{*}{0.61M} &  CA44~\cite{jiang2023t-rex}, 
                                HierText~\cite{long2022hiertext} \\
        \rowcolor{gray!15}
        Image Generation \& Edit  & 5.90M &  JourneyDB~\cite{pan2023journeydb},
        LAION-Aesthetics~\cite{laion_aes}, InstructPix2Pix~\cite{brooks2023instructpix2pix}\\
        Multimodal In-Context  & 0.89M &  MMIC (ours) \\
        
    \end{tabular}
    \caption{Datasets used in stage-3. 
}
\label{subtab:stage4}
\end{subtable}

\caption{\textbf{Summary of datasets used in each training stage.} The datasets used in stage-2 is the combination of stage-1 and stage-3 datasets, which enables the model to learn multiple capacities without comprising its conversation ability. For some large-scale datasets such as SA-1B~\cite{kirillov2023sam}, we randomly sample a subset from them for training.
}
\label{tab:data}
\end{table}

\begin{table}[t]
\scriptsize
\renewcommand{\arraystretch}{1.2}
\centering
\begin{tabular}{l|p{3cm}|l|p{7cm}}
prompt pattern & Task & \#sample &dataset \\
\hline
\multirow{2}{*}{$IMT \rightarrow T$} & \multirow{2}{*}{VQA with visual marks} & \multirow{2}{*}{147K} &COCO~\cite{lin2014coco}, TextOCR~\cite{Singh2021TextOCRTL}, AP10K~\cite{yu2021ap10k} ICDAR2019~\cite{Sun2019ICDAR2C}, AP10K~\cite{yu2021ap10k}\\
\rowcolor{gray!15}
\multirow{4}{*}{$[IT]I \rightarrow T$}& \multirow{4}{*}{\makecell[l]{In-context VQA,\\In-context captioning,\\In-context visual recognition}} & \multirow{4}{*}{465K}  &Food-101~\cite{bossard2014food}, Oxford Flower~\cite{nilsback2008automated}, CUB-200-2011~\cite{WahCUB_200_2011}, Stanford Dogs~\cite{KhoslaYaoJayadevaprakashFeiFei_FGVC2011}, Oxford-IIIT Pet~\cite{parkhi2012cats}, Stanford Cars~\cite{krause20133d}, Birdsnap~\cite{berg2014birdsnap}, VegFru~\cite{hou2017vegfru}, iNaturalist 2021~\cite{van2021benchmarking}, UECFOOD-256
~\cite{kawano2015automatic}, CNFOOD-241~\cite{CNFOOD241}, ALLaVA~\cite{chen2024allava}, COCO~\cite{lin2014coco}\\
\multirow{2}{*}{$[IMT]IM \rightarrow T$}& In-context visual recognition with visual marks & \multirow{2}{*}{40K} &\multirow{2}{*}{COCO~\cite{lin2014coco}, TextOCR~\cite{Singh2021TextOCRTL}, ICDAR2019~\cite{Sun2019ICDAR2C}, AP10K~\cite{yu2021ap10k}}\\
\rowcolor{gray!15}
$[IM]I \rightarrow M$ & \makecell[l]{In-context object detection\\In-context segmentation\\In-context OCR}& 240K &COCO~\cite{lin2014coco}, TextOCR~\cite{Singh2021TextOCRTL}, ICDAR2019~\cite{Sun2019ICDAR2C}\\
Total&& 892K & \\
\end{tabular}
\vspace{2mm}
\caption{\textbf{Datasets used for visual prompting tasks and in-context visual tasks.} In the table, \textit{I} denotes Image, \textit{M} denotes Mask, such as segmentation mask or visual prompts, and \textit{T} denotes Text. In addition, we use ``[*]'' to represent that the item within ``[]'' repeats one or more times.}
\label{tab:ic_datasets}
\end{table}

To support the training for enhancing our model with various capacities, we meticulously collect and re-organize the datasets from a broad range of tasks. These data are publicly available, and we comprehensively list all the data we used in Table~\ref{tab:data}. In addition to the commonly used dataset for the standard vision and vision-language tasks,
we find that many works explore visual prompting strategies and in-context learning. However, there is still a lack of public datasets focusing on addressing these tasks currently. 
To this end, we organize a series of datasets into a new one coined as a multimodal in-context (\textbf{MMIC}) dataset to facilitate the model with in-context learning abilities, applicable to both visual and textual prompts.
As shown in Table~\ref{tab:ic_datasets}, built upon several datasets, we support lots of visual prompting and in-context tasks for fine-grained visual recognition, including categories such as cats, dogs, fruits, vegetables, food, cars, birds, etc. Additionally, we also make efforts on in-context object detection, in-context object segmentation, in-context captioning, in-context OCR, and in-context VAQ. 

\vspace{-1mm}
\subsection{MMIC Dataset Construction}
For tasks that require visual or textual in-context examples, we randomly select \( N \) samples, where \( N \in [2, 6] \), without replacement from the dataset. The first \( N-1 \) samples are presented as in-context examples of the model. These examples serve to provide a reference or a guide for the type of output expected. The model is then tasked with solving or completing the task based on the last sample in the sequence. This paradigm allows the model to learn from examples and apply that knowledge to new, unseen data.

Inspired by \cite{shtedritski2023does}, visual marks can also serve as the input for multimodal LLMs. As a result, we design five types of visual marks: circle, hand-drawn circle, arrow, box, and mask. Each visual mark can be either solid or hollow. We primarily construct this dataset based on COCO~\cite{lin2014coco}, AP10K~\cite{yu2021ap10k} and some OCR datasets~\cite{Sun2019ICDAR2C,Singh2021TextOCRTL}, where we randomly sample \( M \)($\in$ [1, 5]) instances per image. The same type of visual mark is used to highlight the selected instance within one image, ensuring consistency and clarity for the model's learning process.

The examples of constructed instructions can refer to Figure~\ref{fig:ic_fg},~\ref{fig:ic_captioning},~\ref{fig:ic_det_seg} and ~\ref{fig:ic_imt}. The entire dataset has constructed a multimodal corpus with $\sim$862K question\&answer pairs. We expect that this dataset can further advance the development of this field.

\section{Training Details}
\label{sec:detail_train}

\begin{figure}[t]
    \centering
    \includegraphics[width=0.98\textwidth]{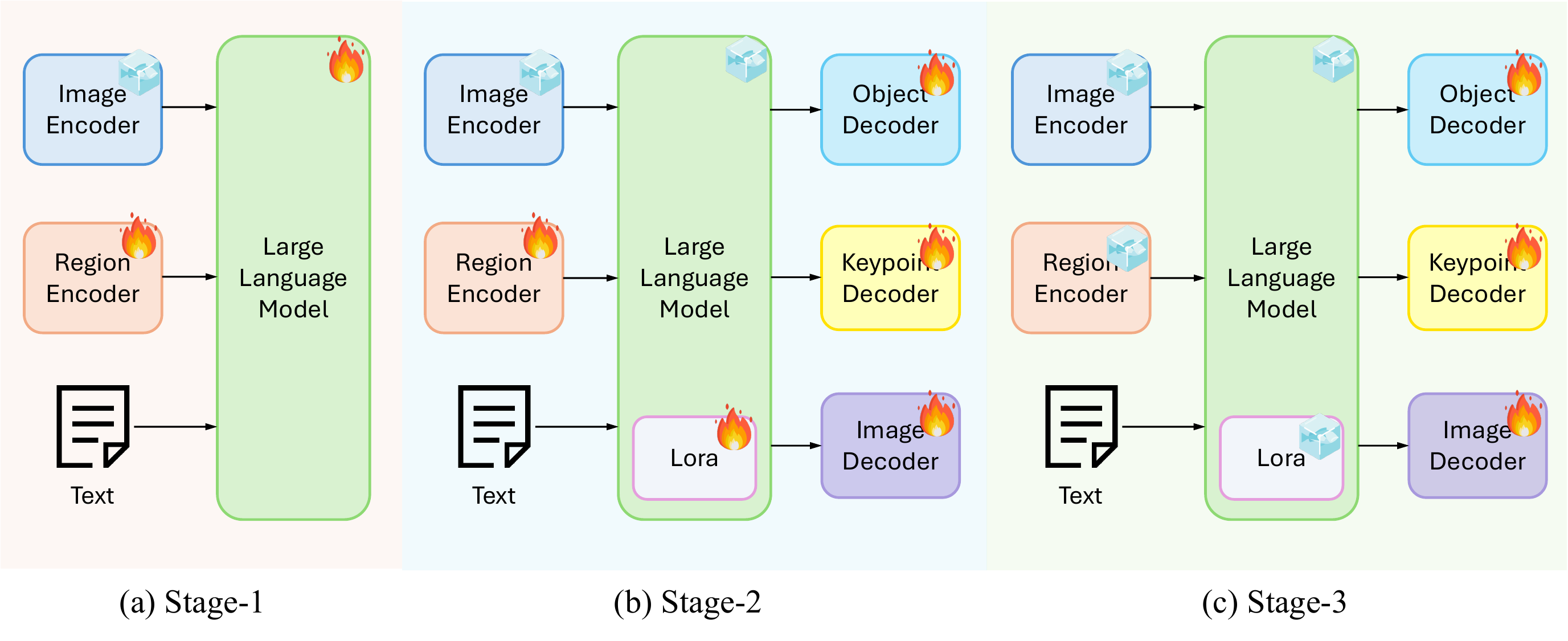}
    \caption{\textbf{The training strategy of \modelname.} It consists of three consecutive stages: (1) multimodal training; (2) multi-capacity fine-tuning; (3) decoder-only fine-tuning. Leveraging this training strategy, \modelname progressively learns the global knowledge and enhances its capacities from a broad range of data sources.
    } 
    \label{fig:train_strategy}
    \vspace{-5pt}
\end{figure}

Figure~\ref{fig:train_strategy} depicts the three-stage training process. Table~\ref{tab:train_config} lists the detailed training configurations of \modelname in different training stages. In each stage, the model inherits the weights from the previous stage and continues training. The image encoder keeps frozen in all stages following previous works~\cite{liu2023llava, liu2023llava1.5}.

\noindent
\textbf{Settings of Stage-1.} Stage-1 consists of pretraining and instruction tuning phases as ~\cite{liu2023llava, liu2023llava1.5}. As shown in Table~\ref{tab:train_config}, in the pretraining phase, We freeze the LLM. And only the region encoder and projections for image embedding and region embedding are trained for efficiency. We adopt the AdamW optimizer~\cite{loshchilov2017adamw} with the peak learning rate of 1e-3 and weight decay of 0. The training involves a total batch size of 2048 across 64 A100 GPUs. In the instruction tuning phase, LLM is unfrozen for full-parameter training. The peak learning rate is decreased to 2.5e-5 for training stabilization. The model is trained on 64 A100 GPUs with a total batch size of 1024. And we begin adopting the dynamic resolution approach~\cite{liu2024llava-next, chen2024internvl_1_5} in this phase. The maximal number of local patches, \textit{i.e.}, max tile, is set as 4.

\noindent
\textbf{Settings of Stage-2.} In stage-2, we add the task-specific decoders and perform the multi-capacity fine-tuning. 
LLM and region encoder are trained with the peak learning rate of 1e-5, while the decoders are trained with the peak learning rate of 1e-4. The model is trained on 128 A100 GPUs with a batch size of 2 per GPU.

\noindent
\textbf{Settings of Stage-3.} In stage-3, we freeze all the components except for the task-specific decoders to maintain the conversational ability. The model undergoes 12 training epochs on 128 A100 GPUs with a peak learning rate of 1e-4 and a total batch size of 256.

These three stages take around 5 / 3 / 10 days to finish the training, respectively.

\noindent
\textbf{Training Losses.} During training, we use the standard cross-entropy loss in stage-1. In stage-2 and stage-3, when integrating the task-specific decoders, we simply sum the losses from the LLM and decoders directly, without reweighting each component. \ie,

\vspace{-1mm}
\begin{equation}
    L_{\text{total}} = L_{\text{llm}} + L_{\text{gdino}} + L_{\text{unipose}} + L_{\text{sd}} + L_{\text{ip2p}}
    \label{eq:train_loss}
\end{equation}
\vspace{-1mm}

\begin{table*}[t!]
\scriptsize
\centering
\renewcommand{\arraystretch}{1.0}
\setlength{\tabcolsep}{2.2mm}

\begin{tabular}{l|cccc}
         config & stage1 pretrain. & stage1 tune. & stage2 & stage3 \\
         \hline
         image enc. peak learning rate & frozen & frozen & frozen & frozen \\
         region enc. peak learning rate & 1e-3 & 2.5e-5 & 1e-5 & frozen \\
         LLM peak learning rate & frozen & 2.5e-5 & 1e-5 & frozen \\
         dec. peak learning  rate & - & - & 1e-4 & 1e-4 \\
         learning rate schedule & cosine decay & cosine decay & cosine decay & cosine decay \\
         optimizer & AdamW~\cite{loshchilov2017adamw} & AdamW~\cite{loshchilov2017adamw} & AdamW~\cite{loshchilov2017adamw} & AdamW~\cite{loshchilov2017adamw} \\
         weight decay & 0. & 0. & 0. & 0. \\
         input resolution & 336$^2$ & 336$^2$ & 336$^2$ & 336$^2$ \\
         dynamic resolution & \no & \yes & \yes & \yes \\
         max tile & - & 4 & 4 & 4 \\
         LLM LoRA rank & - & - & 32 & 32 \\
         LLM LoRA alpha & - & - & 64 & 64 \\
         warmup ratio & 0.03 & 0.03 & 0.03 & 0.03 \\
         total batch size & 2048 & 1024 & 256 & 256 \\
         epoch & 1 & 1 & 1 & 12 \\
         numerical precision & DeepSpeed bf16~\cite{rasley2020deepspeed} & DeepSpeed bf16~\cite{rasley2020deepspeed} & DeepSpeed bf16~\cite{rasley2020deepspeed} & DeepSpeed bf16~\cite{rasley2020deepspeed} \\
         GPUs for training & 64 $\times$ A100 (80G) & 64 $\times$ A100 (80G) & 128 $\times$ A100 (80G) & 128 $\times$ A100 (80G) \\
     
    \end{tabular}

\caption{\textbf{Training settings of \modelname in different stages.} 
Max tile means the maximal number of local patches when adopting the dynamic resolution approach~\cite{liu2024llava-next, chen2024internvl_1_5} for the images. 
}
\label{tab:train_config}
\end{table*}

\section{Instruction Templates}
\label{sec:detail_template}

To support the proper invocation of task-specific decoders, we construct a series of instruction templates for different tasks using ChatGPT~\cite{openai2023gpt4} and use them as instruction tuning data for LLM. We comprehensively list all the instruction templates below, from Table~\ref{tab:instructions-single-region-detail-caption} to Table~\ref{tab:instructions-image-generation}. 

\clearpage

\begin{table*}[t]
\centering
\begin{tcolorbox}[colback=white!100]
\centering
\footnotesize
\begin{tabular}{p{0.97\columnwidth} c}
1. Can you provide a detailed description of <regions> in the image?\\
2. From what you see in the image, could you paint a vivid picture of what <regions> looks like?\\
3. What stands out to you the most about <regions> depicted in the image? Could you describe it in detail?\\
4. I'm interested in this image, especially in the <regions>. Can you provide a comprehensive description of it?\\
5. I'd like to learn about the detailed information of <regions> in this image. Can you describe its characteristics in depth?\\
6. The <regions> in this image seems fascinating. Can you delve into its description, highlighting its notable aspects?\\
7. Can you paint a vivid picture of the scenery within <regions> captured in the image?\\
8. Could you provide a detailed account of the environmental characteristics within <regions> in the image?\\
9. I'm seeking more information about <regions> in the image. Could you provide a comprehensive overview?\\
10. Please help me write a detailed description for <regions> in the image.\\
\end{tabular}
\end{tcolorbox}
\caption{\textbf{A list of instructions for single-region detailed caption.}}
\label{tab:instructions-single-region-detail-caption}
\end{table*}

\begin{table*}[t]
\centering
\begin{tcolorbox}[colback=white!100]
\centering
\footnotesize
\begin{tabular}{p{0.97\columnwidth} c}
1. Can you provide me with a brief description of <regions> in the picture?\\
2. I'm curious about the region represented by <regions> in the picture. Could you describe it in short?\\
3. What can you tell me about <regions> in the image?\\
4. I'd like to know more about the area in the photo labeled <regions>. Can you give me a brief description?\\
5. Could you describe <regions> in the picture in short?\\
6. What content can you give me about <regions> in the photo?\\
7. Please provide me with a short description of <regions> in the image.\\
8. Can you give me a brief account of the region labeled as <regions> in the picture?\\
9. I'm interested in learning more about <regions> in the photo. Can you describe it in short?\\
10. What is the region outlined by <regions> in the picture like? Could you give me a brief description?\\
11. Can you provide me with a brief description of <regions> in the picture, please?\\
12. I'm curious about the region represented by <regions> in the picture. Could you describe it in short, please?\\
13. What can you tell me about <regions> in the image, exactly?\\
14. I'd like to know more about <regions>. Can you give me a brief description?\\
15. Could you describe the region shown as <regions> in the picture in short, please?\\
16. What content can you give me about <regions> in the photo, please?\\
17. Please provide me with a short description of <regions> in the image, please.\\
18. Can you give me a brief account of the region labeled as <regions> in the picture, please?\\
19. I'm interested in learning more about <regions> in the photo. Can you describe it in short, please?\\
20. What is <regions> in the picture like, please? Could you give me a brief description?\\
\end{tabular}
\end{tcolorbox}
\caption{\textbf{A list of instructions for single-region brief caption.}}
\label{tab:instructions-single-region-brief-caption}
\end{table*}

\begin{table*}[t]
\centering
\begin{tcolorbox}[colback=white!100]
\centering
\footnotesize
\begin{tabular}{p{0.97\columnwidth} c}
1. Could you please give me a brief description of <regions>?\\
2. Can you provide a short description of <regions> in this image?\\
3. Please describe in short the contents of the boxed areas <regions>.\\
4. Could you give a brief explanation of what can be found within <regions> in the picture?\\
5. Could you give me a brief explanation of <regions> in this picture?\\
6. Can you provide a short description of <regions> in this photo?\\
7. Help me understand the specific locations labeled <regions> in this picture in short, please.\\
8. What is the brief information about the areas marked by <regions> in this image?\\
9. Could you provide me with a brief analysis of the regions designated <regions> in this photo?\\
10. What are the specific features of the areas marked <regions> in this picture that you can describe in short?\\
11. Could you elaborate on the regions identified by <regions> in this image?\\
12. What can you tell me about the areas labeled <regions> in this picture?\\
13. Can you provide a brief analysis of <regions> in this photo?\\
14. I am interested in learning more about <regions> in this image. Can you provide me with more information?\\
15. Could you please provide a brief description of <regions> in this photo?\\
16. What is the significance of the regions labeled <regions> in this picture?\\
17. I would like to know more about <regions> in this image. Can you provide me with more information?\\
18. Can you provide a brief breakdown of <regions> in this photo?\\
19. What specific features can you tell me about the areas identified by <regions> in this picture?\\
20. Could you please provide a short explanation of the locations labeled <regions> in this image?\\
21. Can you provide a brief account of the regions designated <regions> in this photo?\\
22. I am curious about the areas marked <regions> in this picture. Can you provide me with a brief analysis?\\
23. What important content can you tell me about the specific locations identified by <regions> in this image?\\
24. Could you please provide a brief description of <regions> in this photo?\\
25. What can you tell me about the features of the areas designated <regions> in this picture?\\
26. Can you provide a comprehensive overview of the regions marked <regions> in this image?\\
27. I would like to know more about the specific locations identified by <regions> in this photo. Can you provide me with more information?\\
28. What is the detailed information you have on <regions> in this picture?\\
29. Could you provide me with a brief analysis of <regions> in this image?\\
30. Can you provide a brief explanation of the specific locations marked by <regions> in this photo?\\
\end{tabular}
\end{tcolorbox}
\caption{\textbf{A list of instructions for multi-region caption.}}
\label{tab:instructions-multi-region-caption}
\end{table*}

\begin{table*}[t]
\centering
\begin{tcolorbox}[colback=white!100]
\centering
\footnotesize
\begin{tabular}{p{0.97\columnwidth} c}
1. Whis is the object category of <regions>? Answer the question with a single word or phrase.\\
2. Could you tell me what is the object in <regions>? Answer the question with a single word or phrase.\\
3. What category best describes the area represented by <regions>? Answer the question with a single word or phrase.\\
4. Can you specify the type of object inside the region labeled by <regions>? Answer the question with a single word or phrase.\\
5. How would you label the area indicated by <regions> in the image? Answer the question with a single word or phrase.\\
6. Give a category label to the region outlined by <regions>. Answer the question with a single word or phrase.\\
7. Please identify the category of the object inside the <regions>. Answer the question with a single word or phrase.\\
8. Examine and determine the primary subject located within <regions>. Answer the question with a single word or phrase.\\
9. I need your help to assign an object category to the <regions>, please. Answer the question with a single word or phrase.\\
10. Evaluate the content of the region shown as <regions> and provide its category. Answer the question with a single word or phrase.\\
\end{tabular}
\end{tcolorbox}
\caption{\textbf{A list of instructions for region recognition.}}
\label{tab:instructions-region-recognition}
\end{table*}

\begin{table*}[t]
\centering
\begin{tcolorbox}[colback=white!100]
\centering
\footnotesize
\begin{tabular}{p{0.97\columnwidth} c}
1. Can you analyze the image and identify the <class> present?\\
2. In this image, could you detect all instances of <class>?\\
3. Are you capable of identifying <class> within this image?\\
4. Could you please detect the objects you find that belong to the <class> category in the image?\\
5. Can you perform object detection on the image and tell me the <class> you find?\\
6. I'm trying to detect <class> in the image. Can you help me?\\
7. Can you carry out object detection on this image and identify the <class> it contains?\\
8. In the context of the image, I'd like to know which objects fall under the category of <class>. Is that something you can do?\\
9. I have an image that needs examination for objects related to <class>. Can you perform that?\\
10. Can you determine if there are any <class> present in the image using object detection?\\
11. Could you please carry out object detection on this image and list any <class> that you discover?\\
12. Could you help me identify the objects corresponding to <class> in the provided image?\\
13. Are you capable of detecting and labeling <class> objects within the image?\\
14. I'm curious about the objects in the image that correspond to the <class> category. Could you assist in finding them?\\
15. Can you detect <class> within the image and provide information about its presence?\\
16. Please examine the image and let me know which objects fall under the <class> category.\\
17. Please perform object detection on this image to identify <class>.\\
18. I need your expertise to locate <class> in this image.\\
19. Please let me know the objects falling into the <class> category in the image.\\
20. Please help me identify objects falling under the <class> category in this image.\\
21. Please assist me in identifying the <class> objects within the image.\\
22. Please provide a breakdown of all the <class> objects visible in the image.\\
23. Please analyze the image and let me know if you can find any objects categorized as <class>.\\
24. I'm seeking your help in identifying <class> within the contents of the image.\\
25. Please conduct object detection on the image to locate any <class> that may be present.\\
26. Please execute object detection on this image and provide details about any <class> you detect.\\
27. Please identify and list any <class> in the given image using object detection.\\
28. Please analyze the image and let me know if there are any recognizable <class> objects.\\
29. Detect any <class> in the given image, if possible.\\
30. I need assistance in recognizing the <class> shown in the image.\\
\end{tabular}
\end{tcolorbox}
\caption{\textbf{A list of instructions for object detection.}}
\label{tab:instructions-object-detection}
\end{table*}

\begin{table*}[t]
\centering
\begin{tcolorbox}[colback=white!100]
\centering
\footnotesize
\begin{tabular}{p{0.97\columnwidth} c}
1. Where can we locate the <expression> in the image?\\
2. Do you know where the <expression> is within the image?\\
3. Have you seen the <expression> in this image? Where is it?\\
4. Could you tell me where the <expression> is in the image?\\
5. Whereabouts in the image can we find the <expression>?\\
6. Do you have any idea where the <expression> might be in this image?\\
7. Are you aware of the <expression>'s position within the image?\\
8. Where in the image should we be looking for the <expression>?\\
9. Is it possible to identify the <expression>'s location in this image?\\
10. Have you figured out where the <expression> is in this image?\\
11. Could you provide guidance on finding the <expression> in the image?\\
12. Do you know where I can locate the <expression> in the picture?\\
13. Can you tell me the precise location of the <expression> in the image?\\
14. Would you be able to point out the <expression> within the image?\\
15. Are you able to discern the <expression> in the image?\\
16. Please help me locate the <expression> in the image.\\
17. Please find the object indicated by the expression <expression> in the image.\\
18. Please assist in identifying the <expression> within the image.\\
19. Please determine the exact position of the <expression> in the image.\\
20. Please ascertain the whereabouts of the <expression> in this image.\\
21. Please assist me in locating the <expression> within the image.\\
22. Please take a moment to find the object denoted by the expression <expression> in the image.\\
23. Please help us identify the precise location of the <expression> in this image.\\
24. Please provide your guidance in finding and marking the <expression> within the image.\\
25. Please make it a priority to discover and highlight the <expression> within the image.\\
26. Let's determine the specific area where the <expression> is situated in the image.\\
27. We're aiming to establish the spatial coordinates of the <expression> in this image.\\
28. We need to establish the exact whereabouts of the <expression> within the image.\\
29. We are actively engaged in the process of locating the <expression> in the image.\\
30. Let's find the <expression> within the image.\\
\end{tabular}
\end{tcolorbox}
\caption{\textbf{A list of instructions for visual grounding.}}
\label{tab:instructions-visual-grounding}
\end{table*}

\begin{table*}[t]
\centering
\begin{tcolorbox}[colback=white!100]
\centering
\footnotesize
\begin{tabular}{p{0.97\columnwidth} c}
1. Could you aid me in generating unique masks for every category present in <class> in this image?\\
2. Can you help me generate distinct masks for each category that belongs to <class> in this image?\\
3. Is it possible for you to help me create distinct masks for the different <class> categories in this image?\\
4. Could you assist me in generating masks that correspond to each individual <class> category in this image?\\
5. Would you mind helping me generate separate masks for each <class> category detected in this image?\\
6. Can you guide me in generating unique masks for all the categories falling under <class> in this image?\\
7. Can you provide me with the necessary support to generate masks specific to each <class> category in this image?\\
8. Could you please guide me in creating separate masks for each <class> category detected in this image?\\
9. Can you support me in generating masks for all the categories encompassed by <class> in this image?\\
10. Examine the image and generate masks that correspond to each individual <class> category present.\\
11. Is it possible for you to help me generate separate masks for each detected category falling under <class> in this image?\\
12. Can you assist me in generating masks that isolate each category belonging to <class> in this image?\\
13. Can you provide me with assistance in generating individual masks for every <class> category identified in this image?\\
14. Can you help with the process of generating masks that are specific to each <class> category detected in this image?\\
15. Generate masks that accurately depict each category belonging to <class> in this image.\\
16. I require assistance in producing separate masks for all the <class> categories in this image.\\
17. I need your support to generate masks that are specific to each <class> category in this image.\\
18. Your task is to produce masks that differentiate each category falling under the <class> category in this image.\\
19. Please create masks that are distinct for each category belonging to <class> in this image.\\
20. I'm seeking your help to generate masks that isolate every category within the <class> category in this image.\\
21. Please segment the different categories falling under <class> in this image and generating masks for each.\\
22. Please accurately segment and generate masks for all the categories falling under <class> in this image.\\
23. I need your support to create masks that are specific to each <class> category identified in this image.\\
24. I'm requesting your aid in generating masks that distinguish each category belonging to <class> in this image.\\
25. Please lend me your expertise in creating masks that are unique for each detected <class> category in this image.\\
26. Your help is required to generate distinct masks for each category of <class> in this image.\\
27. It would be appreciated if you could assist in creating separate masks for each <class> category in this image.\\
28. Let's collaborate on segmenting all categories falling under the <class> category in this image and generating masks.\\
29. Assisting me in generating distinct masks for each class categorized as <class> would be greatly appreciated.\\
30. Providing assistance in generating masks that accurately identify the categories falling under <class> in this image would be greatly helpful.\\
\end{tabular}
\end{tcolorbox}
\caption{\textbf{A list of instructions for semantic segmentation.}}
\label{tab:instructions-semantic-segmentation}
\end{table*}

\begin{table*}[t]
\centering
\begin{tcolorbox}[colback=white!100]
\centering
\footnotesize
\begin{tabular}{p{0.97\columnwidth} c}
1. Can you examine the image and pinpoint the keypoint locations of the <class>?\\
2. Could you analyze the picture and determine the keypoint placement of the <class>?\\
3. Please inspect the image and locate the keypoints for <class>.\\
4. Can you evaluate the photo and identify where the keypoints of <class> are situated?\\
5. Look at the image and detect the keypoint positions of the <class>.\\
6. Please analyze this image and find the keypoints of <class>.\\
7. Can you check the image and show me where the keypoints of <class> are located?\\
8. Please find the exact keypoint position of the <class>.\\
9. Please observe the photo and identify the keypoint locations of the <class>.\\
10. Can you review the image and point out the keypoints of <class>?\\
\end{tabular}
\end{tcolorbox}
\caption{\textbf{A list of instructions for pose estimation.}}
\label{tab:instructions-pose-estimation}
\end{table*}

\begin{table*}[t]
\centering
\begin{tcolorbox}[colback=white!100]
\centering
\footnotesize
\begin{tabular}{p{0.97\columnwidth} c}
1. Give me a concise description of the image. Answer the question and localize each object.\\
2. Please briefly summarize the content of this image. Answer the question and localize each object.\\
3. What does this picture show? Please summarize briefly. Answer the question and localize each object.\\
4. Can you give me a quick overview of what's depicted in this image? Answer the question and localize each object.\\
5. Could you describe the key elements in this photograph? Answer the question and localize each object.\\
6. Offer a brief explanation of what this image represents. Answer the question and localize each object.\\
7. Sum up the contents of this picture in one or two sentences. Answer the question and localize each object.\\
8. What is the main content in this image? Answer the question and localize each object.\\
9. Provide a brief caption for the picture. Answer the question and localize each object.\\
10. Could you give me a short description of the image? Answer the question and localize each object.\\
\end{tabular}
\end{tcolorbox}
\caption{\textbf{A list of instructions for grounded caption.}}
\label{tab:instructions-grounded-caption}
\end{table*}

\begin{table*}[t]
\centering
\begin{tcolorbox}[colback=white!100]
\centering
\footnotesize
\begin{tabular}{p{0.97\columnwidth} c}
1. Can you examine the image and segment the corresponding objects denoted as <regions>?\\
2. Where are the objects marked by <regions> in the image? Could you help me segment these objects?\\
3. Could you please segment all the corresponding objects according to the visual prompt as <regions>?\\
4. Can you help me draw the instance segmentation masks of <regions> in the picture?\\
5. Please help me find all the objects shown as <regions> and segment them.\\
6. I'd like to know the objects outlined by <regions>. Please help me draw their masks.\\
7. Given the <regions>, I need your help to segment the corresponding object masks.\\
8. Examine the image and identify all the objects that belong to the provided <regions>.\\
9. I'm interested in the objects labeled as <regions>. Could you please draw their instance masks?\\
10. There are some regions represented by <regions>. I need your assistance to find their corresponding objects.\\
\end{tabular}
\end{tcolorbox}
\caption{\textbf{A list of instructions for interactive segmentation.}}
\label{tab:instructions-interactive-segmentation}
\end{table*}

\begin{table*}[t]
\centering
\begin{tcolorbox}[colback=white!100]
\centering
\footnotesize
\begin{tabular}{p{0.97\columnwidth} c}

1. Generate image with caption: <caption>.\\
2. Can you give me the image with caption: <caption>.\\
3. Help me to generate this image: <caption>.\\
4. Generate the image according to the caption: <caption>.\\
5. According to the caption, generate the image: <caption>.\\
6. An image with caption: <caption>.\\
7. Can you visualize this caption: <caption>.\\
8. Create an image based on this caption: <caption>.\\
9. Generate a visual representation for this caption: <caption>.\\
10. Provide me with an image corresponding to this caption: <caption>.\\
11. Craft an image with the following caption: <caption>.\\
12. Generate an image accompanied by this caption: <caption>.\\
13. Turn this caption into an image: <caption>.\\
14. Generate an image reflecting this caption: <caption>.\\
15. Translate this caption into a visual representation: <caption>.\\
16. Produce an image that matches this caption: <caption>.\\
17. Create an image in line with this caption: <caption>.\\
18. Generate an image to illustrate this caption: <caption>.\\
19. Construct an image based on the given caption: <caption>.\\
20. Give me an image associated with this caption: <caption>.\\
\end{tabular}
\end{tcolorbox}
\caption{\textbf{A list of instructions for image generation.}}
\label{tab:instructions-image-generation}
\end{table*}



\end{document}